\documentclass[journal]{IEEEtran}

\usepackage{graphicx}
\usepackage{subfigure}
\usepackage{lipsum}
\usepackage{cite}
\usepackage{verbatim}
\usepackage{amsmath}
\usepackage{setspace}   
\usepackage{fancyhdr}
\usepackage{color}
\usepackage{amsfonts}
\usepackage{tikz}
\usepackage{amsfonts}
\usepackage{url}
\usepackage{bm}
\usepackage[normalem]{ulem} 
\usepackage{threeparttable}
\usepackage[ruled]{algorithm2e} 

\SetAlFnt{\small}
\SetAlCapFnt{\small}
\SetAlCapNameFnt{\small}
\SetAlCapHSkip{0pt}
\usepackage{subcaption}
\usepackage{stfloats}      
\usepackage{placeins}      
\usepackage{booktabs} 
\usepackage{textcomp}
\usepackage{wrapfig}
\usepackage{physics}
\usepackage{multirow}
\usepackage{xcolor}
\graphicspath{{./fig/}}
\DeclareGraphicsExtensions{.jpg}

\newcommand{\JENew}[1]{{\textcolor[rgb]{0,0,0}{#1}}}
\newcommand{\finalrev}[1]{{\textcolor[rgb]{0,0,0}{#1}}}

\newcommand{\rev}[2]{#2}

\begin{document}
\title{Correspondence-Free, Function-Based Sim-to-Real Learning for Deformable Surface Control}

\author{
Yingjun Tian,
Guoxin Fang,~\IEEEmembership{Member,~IEEE}, Renbo Su, Aoran Lyu, Neelotpal Dutta, Weiming Wang,\\
Simeon Gill, Andrew Weightman, and Charlie C.L. Wang,~\IEEEmembership{Senior Member,~IEEE}       

\thanks{
Y. Tian, R. Su, A. Lyu, N. Dutta, W. Wang, A. Weightman and C.C.L. Wang are all with the Department of Mechanical and Aerospace Engineering, The University of Manchester, United Kingdom (email: yingjun.tian@manchester.ac.uk;
renbo.su@postgrad.manchester.ac.uk; aoran.lyu@postgrad.manchester.ac.uk; wwmdlut@gmail.com; 
andrew.weightman@manchester.ac.uk;
charlie.wang@manchester.ac.uk).

G. Fang is with the Department of Mechanical and Automation Engineering, The Chinese University of Hong Kong, Shatin, Hong Kong. (email: guoxinfang@mae.cuhk.edu.hk).

S. Gill is with the Department of Materials, The University of Manchester, United Kingdom (email: simeon.gill@manchester.ac.uk).

The project is partially supported by the chair professorship fund of the University of Manchester and the research fund of UK Engineering and Physical Sciences Research Council (EPSRC) (Ref.\#: EP/W024985/1). 

\textit{Corresponding authors}: Charlie C.L. Wang.

}

}

\markboth{IEEE Transactions on Robotics}
{\MakeLowercase{Tian \textit{et al.}}: Correspondence-Free, Function-Based Sim-to-Real Learning for Deformable Surface Control}

\maketitle

\begin{abstract}
This paper presents a correspondence-free, function-based sim-to-real learning method for controlling deformable freeform surfaces. Unlike traditional sim-to-real transfer methods that strongly rely on marker points with full correspondences, our approach simultaneously learns a deformation function space and a confidence map -- both parameterized by a neural network -- to map simulated shapes to their real-world counterparts. As a result, the sim-to-real learning can be conducted by input from either a 3D scanner as point clouds (without correspondences) or a motion capture system as marker points (tolerating missed markers). The resultant sim-to-real transfer can be seamlessly integrated into a neural network-based computational pipeline for inverse kinematics and shape control. We demonstrate the versatility and adaptability of our method on two vision devices and across four pneumatically actuated soft robots: a deformable membrane, a robotic mannequin, and two soft manipulators.
\end{abstract}

\begin{IEEEkeywords}
Sim-to-Real learning, correspondence-free, free-form surface, deformation control, soft robotics.
\end{IEEEkeywords}

\IEEEpeerreviewmaketitle

\section{Introduction}\label{secIntro}
Soft robotic systems have demonstrated the ability to dynamically adapt their shapes in response to programmed actuation or environmental stimuli. Existing prototypes were developed for haptic interfaces (ref.~\cite{inForm2013UIST,Stanley2015CtrlSurf,Bryan2019SoRoInterface,koehler2020model,Je2021ElevateAW})\JENew{~and soft manipulators (ref.~\cite{della2020model,Thomas2019TRO, Renda2014TRO}). Many of these soft robotic systems have free-form surfaces with large deformation to be controlled -- e.g., a soft robotic mannequin} developed in~\cite{Tian2022SoRoMannequin}, which is driven by pneumatic actuators to mimic the front shapes of different human bodies. 
\JENew{
Shape control of such freeform surfaces is challenging due to the infinite \textit{degrees-of-freeform} (DoFs) deformation embedded in the elastic bodies of soft robots.
}

\subsection{Kinematics for Shape Control}\label{subsecIntroShapeCtrl}
An essential shape control problem of soft robotics is to compute the actuation (denoted by a vector $\mathbf{a}$) that can drive the free-form surface $\mathcal{S}$ to approach a target shape $\mathcal{S}^t$. It presents much higher DoFs compared to the articulated robots, which leads to high complexity in both the forward shape estimation (i.e., predicting the resultant shape $\mathcal{S}$) and the \textit{inverse kinematics} (IK) computation to determine the values of actuation parameters $\mathbf{a}$. Earlier works have been developed to conduct an iteration-based IK solver with Jacobian physically evaluated on hardware setup (e.g.,~\cite{Tian2022SoRoMannequin, Yip2014TROHardwardGradient}). However, this approach is time-consuming in both computation and data acquisition. Additionally, its performance is unstable as the Jacobian evaluation is highly dependent on the accuracy of shapes captured on the physical setup, which is hard to be guaranteed in practice. 

\begin{figure}[!t] 
\centering
\includegraphics[width=1.0\linewidth]{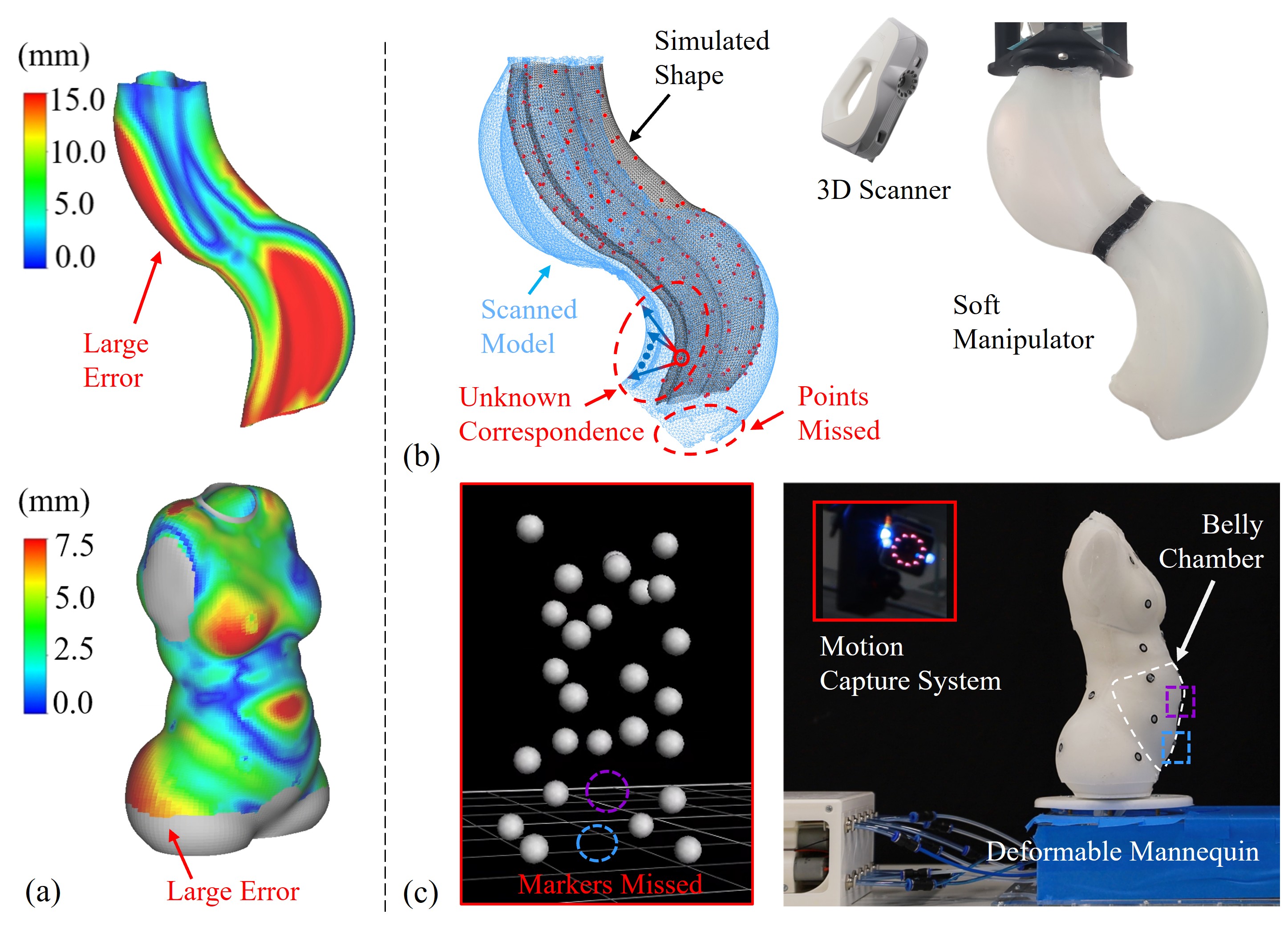}\\
\caption{Motivation and challenges. 
\JENew{(a) Significant geometric discrepancies visualized by a color map are observed between simulation and reality on a soft manipulator composed of six chambers and a soft robotic mannequin with nice pneumatic actuation DoFs. 
(b) When using a 3D scanner to capture dense point clouds of the real shape, conventional sim-to-real approaches fail due to the absence of known correspondences between scanned points and the regions with partially missed points. 
(c) When using a motion capture (MoCap) system to collect marker positions for sim-to-real correction, markers can be missed due to occlusion caused by large local deformation (e.g., the two missed on the belly). As a result, this frame cannot be used in conventional correspondence-based sim-to-real learning.
}
}\label{fig:motivation}
\end{figure}

In the existing research works of soft robotics, high DoFs in the configuration space are typically computed by either reduced analytical models or discrete numerical models \JENew{(see the comprehensive review in~\cite{Renda2023TROModeling,arriola2020modeling})}. Reduced analytical models are often adopted for soft manipulators with simple bending deformation (e.g., ref. \cite{marchese2016design,Thomas2019TRO}), where the surface deformation is trivial in general. \JENew{Differently, numerical simulation is usually employed to establish the mapping $\mathcal{S}(\mathbf{a})$ between the actuation parameters $\mathbf{a}$ and the deformed freeform shapes $\mathcal{S}$ (ref.~\cite{fangICRA,Goury2018TRO,du2021_diffpd,hu2019chainqueen}). The differentiation of this forward kinematic mapping can then be used to solve the IK problem for determining the actuation parameters $\mathbf{a}$ according to the target shape $\mathcal{S}^t$. This was formulated as an optimization problem to iteratively run a simulator to determine the optimal actuation $\mathbf{a}_{opt}$ that minimizes the difference between $\mathcal{S}(\mathbf{a})$ and $\mathcal{S}^t$ \cite{Guoxin2020TRO}.} However, numerical simulation in general cannot be computed efficiently to realize a fast IK solver. A widely adopted strategy to mitigate the efficiency issue is to train a \textit{neural network} (NN) \JENew{by the simulation dataset as a surrogate model} for the forward shape estimation (e.g., ref.~\cite{Thomas2018SoRoCtrlReview, bern2020soft, Fang2022Sim2Real}). \JENew{This paper also adopts this strategy for IK computation.}

\subsection{Challenges of Sim-to-Real Transfer}\label{subsecIntroSim2Real}
\JENew{
Freeform surfaces predicted by numerical simulation often diverge from physically deformed shapes due to hardware uncertainties and simplifications in modeling. These discrepancies become significant during large deformations of a soft robot’s surface (see Fig.~\ref{fig:motivation}(a) for an example). To address this issue, sim-to-real transfer networks have been trained using marker-based MoCap systems in prior works \cite{Zhang2022Sim2RealSoRoFish,Dubied2022Sim2RealFEM,Fang2022Sim2Real}, where marker positions collected from physical experiments are used to align simulation outputs with real-world shapes. However, marker-based methods face three major limitations.} 
\begin{itemize}
    \item \JENew{First, they offer limited spatial coverage, as only a sparse set of markers can be tracked. For example, only the position of the end effector is corrected in \cite{Fang2022Sim2Real}, while the shape of the entire manipulator remains unaccounted errors.}

    \item \JENew{Second, these methods require full correspondence between captured and simulated points, making them incompatible with 3D scanners that output unstructured point clouds (see Fig.\ref{fig:motivation}(b)).}

    \item \JENew{Lastly, marker-based capture is not robust to occlusion. In cases of significant deformation, some markers may be hidden from views -- such as the missing belly markers in Fig.\ref{fig:motivation}(c), making the affected frames unusable.}
\end{itemize}   
Consequently, \finalrev{a correspondence-free method for sim-to-real learning is needed to correct} shape errors under large deformations.

\subsection{Our Method}
\finalrev{This paper focuses on a learning-based sim-to-real transfer method that corrects shape errors in NN-based forward prediction trained on simulation data. The method establishes a function-based mapping between simulated and real deformation responses, enabling efficient and accurate NN-based IK computation for soft robot control. It builds on a learning-based theoretical foundation rather than a new control strategy.}

\begin{figure}
\centering
\includegraphics[width=1.0\linewidth]{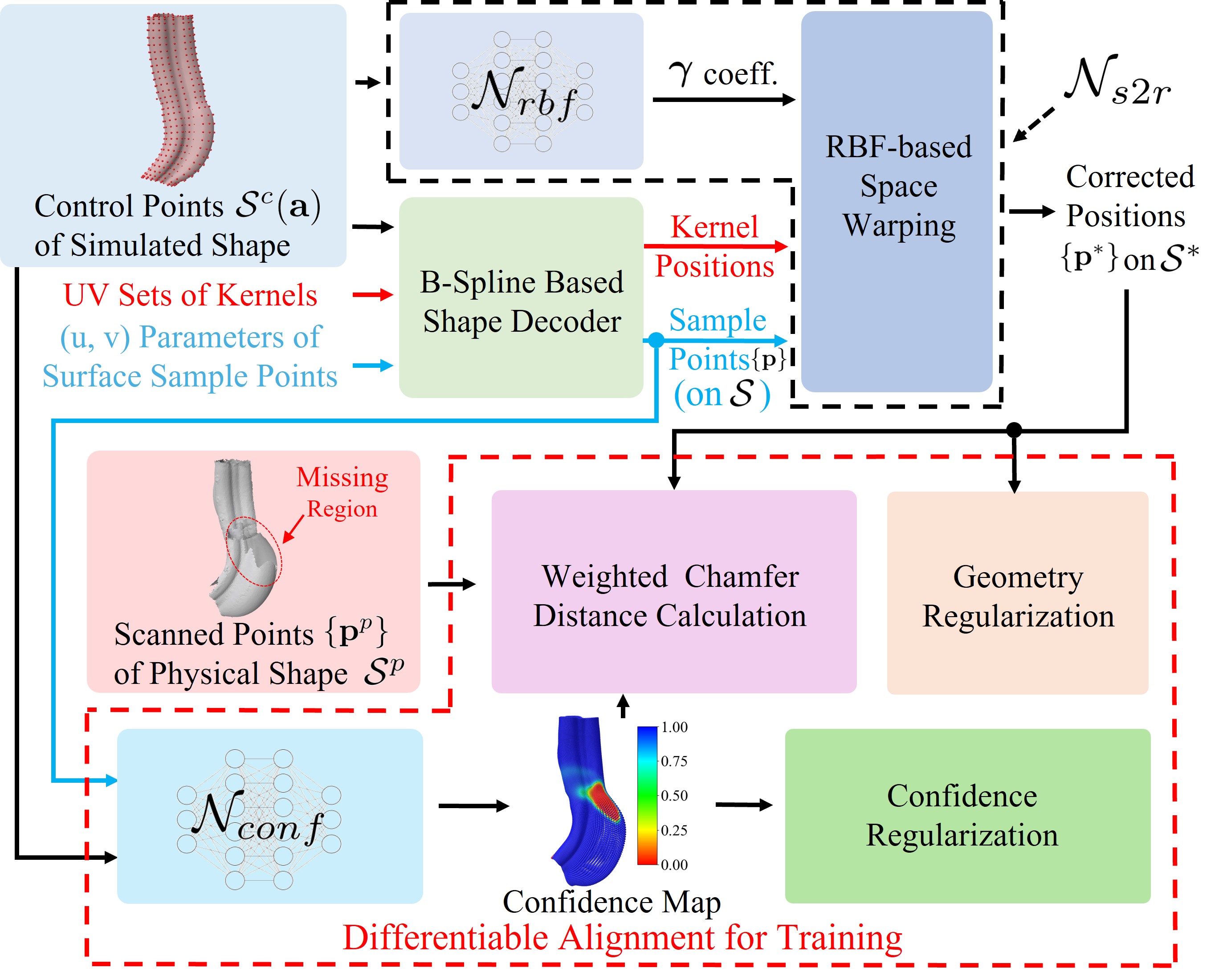}\\
\caption{\JENew{Overview of our correspondence-free sim-to-real learning pipeline: the function based space warping learning and the differentiable alignment based joint training.}
}\label{fig:sim2RealPipelineOverview}
\end{figure}

\JENew{Using a compact B-spline based representation of freeform surfaces, we propose a function-based pipeline that enables robust sim-to-real learning even when the input lacks of point correspondences (e.g., 3D scanned point clouds) or contains missing markers (e.g., MoCap data) / regions (e.g., 3D scan). 
First of all, it is based on} 
a new architecture that learns the function space of shape deformation represented by \textit{radial basis functions} (RBF). Specifically, each simulated free-form surface is approximated by a B-spline surface, the control points of which are employed as the shape descriptor in a lower dimension. The input of our sim-to-real network is the control points of a simulated body shape, and the output is the coefficients of RBFs that define a spatial warping function $\bm{\Phi}(\mathbf{p})$ by using the positions of kernel centers (Sec.~\ref{secFuncSpace}). 
\JENew{The sim-to-real network $\mathcal{N}{s2r}$ consists of two components enclosed by the black dashed lines as shown in Fig.~\ref{fig:sim2RealPipelineOverview}. Rather than relying on explicit point correspondences, it is trained jointly with a network $\mathcal{N}{conf}$ that predicts a confidence map. This map serves as a weighting function for evaluating the Chamfer distance between simulated and captured shapes. The training objective of sim-to-real transfer minimizes this weighted distance across numerous pairs of simulated and captured shapes, as illustrated in the differentiable alignment training module highlighted by the red dashed lines in Fig.~\ref{fig:sim2RealPipelineOverview}. Details of this training process will be presented in Sec.~\ref{secDiffRegTraining}.}

\JENew{After learning, the} space warping function $\bm{\Phi}(\mathbf{p})$ is obtained in a closed-form and gives a continuous mapping from the shape of a simulated model to the shape of a corrected model that matches the reality. Working together with the forward kinematics prediction that is trained by simulation, this forms an end-to-end NN-based pipeline that can accurately predict the physical shapes of robotic deformable surfaces\JENew{, which is crucial for efficient IK computation}. This computation pipeline based on NN is analytically differentiable so that the gradient-based iteration for IK can be computed efficiently \finalrev{(details given in Sec.~\ref{subsecNNBasedFK}).} The quality of IK solutions will be verified by scanning the physical shape $\mathcal{S}^p$ to compare with the input target shape $\mathcal{S}^t$. All notations used in this paper are summarized in Table \ref{tab:Symbols}.

\begin{table*}[t]
\footnotesize
\centering
\caption{List of Notations}\label{tab:Symbols}
\begin{tabular}{cl|cl}
\hline \hline
\specialrule{0em}{2pt}{1pt}
Symbol & Description  & Symbol & Description  \\ 
\specialrule{0em}{1pt}{1pt} \hline \specialrule{0em}{1pt}{2pt}

$ \mathbf{a} $  & Actuation parameters (i.e., chamber pressures)
&  $\mathcal{S}^t$  & The target body shape to be achieved by IK \\

$\mathcal{S}$ & The simulated shape of the deformable surface & 
 $\mathcal{S}^*$  & The corrected shape obtained from our \textit{sim-to-real} method \\

$\mathcal{S}^c$ & The compact shape descriptor (i.e., control points) of $\mathcal{S}$ &
 $\mathcal{S}^p$    & The 3D shape acquired on the physical setup \\

 $\mathcal{N}_{fk}$ & Forward kinematics network that predicts $\mathcal{S}^c$ from $ \mathbf{a} $ & $\mathcal{N}_{rbf}$ & Function prediction network that predicts $\bm{\gamma}$ by $\mathcal{S}^c$ \\

%
$\bm{\gamma} = [\bm{\alpha}_{0}, \ldots \bm{\beta}_i]$ & Variables to determine the function space & $\mathcal{N}_{s2r}$& Combination of $\mathcal{N}_{rbf}$ and $\bm{\Phi}(\cdot)$ 

\\

$\mathbf{p}$, $\mathbf{p}^*$, $ \mathbf{p}^{p} \in \mathbb{R}^{3}$ & The surface point on $\mathcal{S}$, $\mathcal{S}^*$ and $\mathcal{S}^p$ respectively &
$\{ \mathbf{q}_i \}$ & Centers of Gaussian kernels in $\bm{\Phi}(\mathbf{p})$
\\ 

$\bm{\Phi}(\cdot)$ & The space warping function used to deform $\mathcal{S}$ to $\mathcal{S}^*$ & 
$\mathbf{c}^*$  & Closet point of $\mathbf{p}^*$ on $\mathcal{S}^t$ \\

$(\mathbf{R},\mathbf{t})$ & Rotation matrix and translation vector to re-pose $\mathcal{S}^t$&
$\mathbf{B}(\cdot)$ & B-Spline function as decoder mapping $(u_p,v_p)$ to $\mathbf{p}$ \\


$(u_p,v_p)$ & A point in the parametric domain of B-spline surface &
\finalrev{$M_u,M_v$}  & \finalrev{\# of control points along $u$ and $v$ directions respectively}\\

\JENew{$w_c(\cdot)$} & \JENew{Real-value confidence map on the simulated shape} &
\JENew{$\mathcal{N}_{conf}$} & \JENew{Confidence estimation network that predicts $w_c(\cdot)$} \\

\specialrule{0em}{1pt}{1pt} \hline\hline
\end{tabular}\label{table_variable}
\end{table*}

The technique contributions of our work are summarized as follows: 
\begin{itemize}
\item \JENew{A resilient sim-to-real learning pipeline for deformable free-form surfaces that bridges the reality gap through a space of RBF-based spatial warping functions;}

\item \JENew{A differentiable alignment based joint training method that eliminates the need for correspondences between simulated and captured freeform surfaces;}

\item \JENew{A neural network-based method that can efficiently solve the inverse kinematics problem for shape control in soft deformable robots.}
\end{itemize}
\JENew{The effectiveness of our sim-to-real learning and the resultant fast IK solver has been verified on four pneumatically actuated soft robots, including a deformable membrane, a deformable robotic mannequin, and two distinct soft manipulators.} 

\JENew{This article significantly extends our preliminary work presented in~\cite{Tian-RSS-24}. First, we newly introduce an advanced differentiable alignment-based training method (Sec.~\ref{secDiffRegTraining}) that enables correspondence-free sim-to-real learning using unorganized point clouds with partial occlusions instead of relying on fully corresponding markers captured by a MoCap system. In addition, we have conducted extensive validation and experiments on three new types of soft robots, including a deformable membrane and two distinct soft manipulators to demonstrate the generality and robustness of our approach (Sec.~\ref{secResult}).}

\section{Deformation Function Space for Sim-to-Real}\label{secFuncSpace}
This section introduces our function based sim-to-real learning for shape prediction. Given a set of points $\{ \mathbf{p} \in \mathbb{R}^3 \}$ on the free-form surface of a simulated model $\mathcal{S}$, we propose a method to learn a network $\mathcal{N}_{s2r}$ that can predict a deformation function for mapping these points onto a corrected model $\mathcal{S}^*$ (i.e., the corrected shape that has eliminated the sim-to-real gap). The shape difference between $\mathcal{S}^*$ and the physical shape $\mathcal{S}^p$ should be minimized by the sim-to-real transfer. $\mathcal{N}_{s2r}$ is expected to be a general network that can handle different shapes on the same hardware\footnote{Different networks still need be trained for different soft robotic setups.}.

\begin{figure}
\centering
\includegraphics[width=1.0\linewidth]{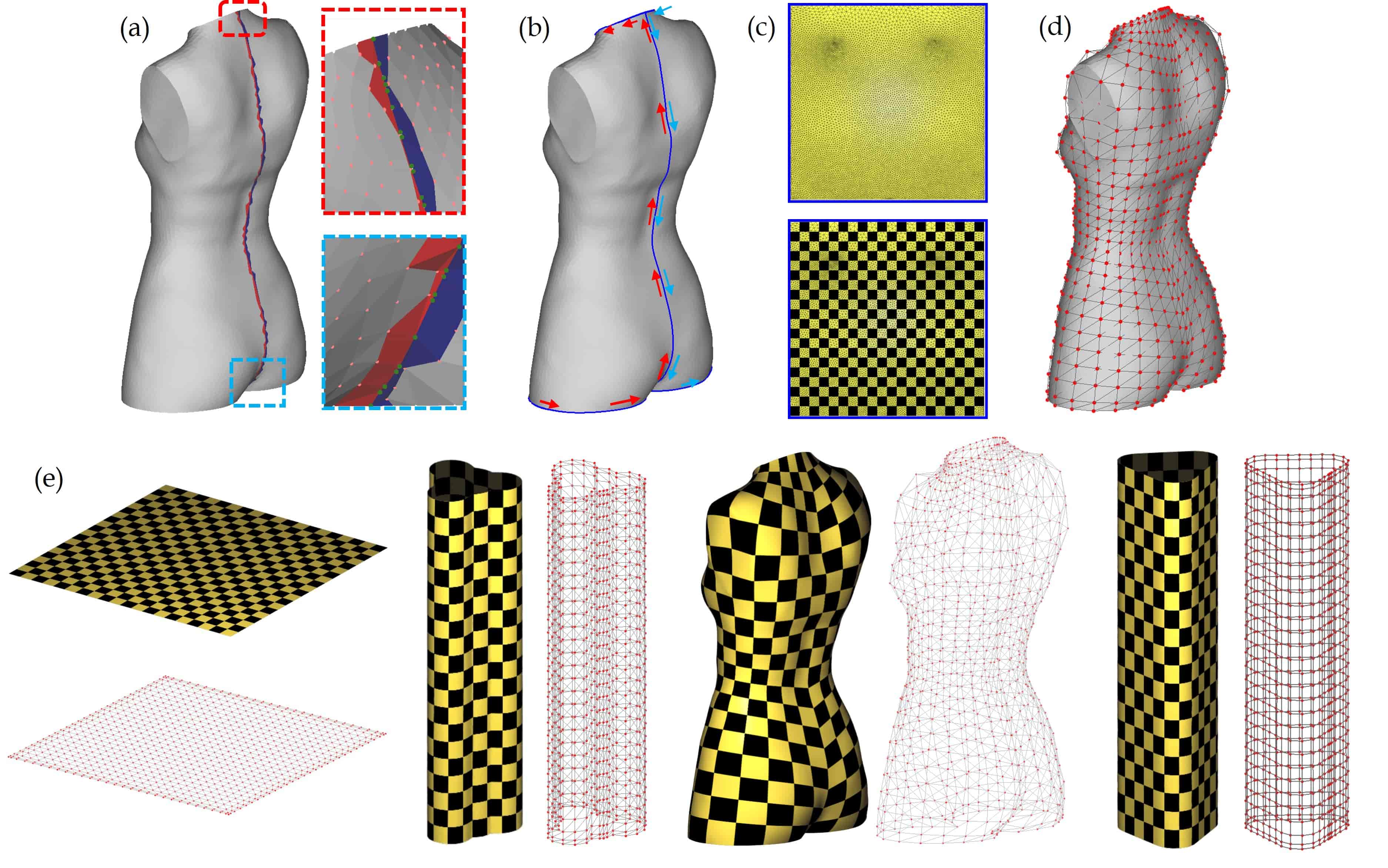}
\caption{\JENew{The parameterization and B-spline fitting steps of a free-form surface include (a) cutting, (b) tracing the boundary, (c) flattening the surface mesh (ref.~\cite{floater2003mean}), and (d) fitting to determine the control points. The B-Spline surfaces and their control points of four different soft robots tested in this paper are shown in (e).}
}\label{fig:BSplineFitting}
\end{figure}

\subsection{Compact Representation by B-Spline Surface}\label{subsecBSplineBasedRep}
Using all sample points on the surface of $\mathcal{S}$ may introduce redundant information and therefore requires a complex network for prediction. \JENew{To overcome this issue, we first convert the freeform surface $\mathcal{S}$ of a soft deformable robot into a B-spline surface representation through 1) the parametrization taken at its rest shape (see Fig.\ref{fig:BSplineFitting}(a-c) for an example) and 2) the surface fitting conducted at every deformed shapes to determine control points of B-spline surface (denoted by $\mathcal{S}^c$ and see Fig.\ref{fig:BSplineFitting}(d)). The control points and B-Spline surfaces for four different soft robots are given in Fig.\ref{fig:BSplineFitting}(e).} 
We then employ the control points of B-spline surface $\mathcal{S}^c$ as the shape descriptor of $\mathcal{S}$, with which any point in the parametric domain with the parameter $(u_p,v_p)$ can be mapped to a point $\mathbf{p} \in \mathbb{R}^3$ as $\mathbf{p}=\mathbf{B}(u_p,v_p,\mathcal{S}^c)$. \finalrev{According to the definition of a B-spline surface~\cite{mortenson1997geometric}, we have
\begin{flalign}\label{eqBSplineFunc}
\mathbf{p}=\mathbf{B}(u,v,\mathcal{S}^c) = \sum_{i=1}^{M_u}\sum_{j=1}^{M_v} N_{i,k}(u) N_{j,l}(v) \mathcal{S}^c_{ij}.
\end{flalign}
The sample point $\mathbf{p}$ on the B-spline surface is the sum of the linear combination of the B-spline basis function (i.e., $N_{ik}(u)$ and $N_{jl}(v)$) and the control point $\mathcal{S}^c_{ij}$. $k$ and $l$ denote the order of B-spline basis function in $u$ and $v$ direction, where cubic B-spline is chosen in the paper (i.e., $k=l=4$). $M_u$ and $M_v$ are the number of control points along $u$ and $v$ direction respectively.}
This function $\mathbf{B}(\cdot)$ is named as the B-spline based shape decoder in our computational pipeline (illustrated as the light green block at the top of Fig.\ref{fig:sim2RealPipelineOverview}). 

The dimension of $\mathcal{S}^c$ using B-spline representation \finalrev{as $M_u \times M_v$} is much lower than mesh representation. For instance, we use $30 \times 30$ control points for \JENew{all robots} as shown in Fig.\ref{fig:BSplineFitting}(e). \JENew{Note that, since the surfaces obtained via simulation retain an unchanged mesh topology (i.e., preserving correspondence), the computation of B-spline surface fitting is based on the parameterization determined by the rest shape of $\mathcal{S}$.}

\subsection{RBF-based Spatial Warping}\label{subsecRBFSpaceWarping}
Considering the gap between the simulated shape $\mathcal{S}$ and the physical shape $\mathcal{S}^p$, we propose to model it as continuous spatial warping by the \textit{radial basis functions} (RBF) \finalrev{for any 3D point $\mathbf{p} \in \mathbb{R}^3$} as follows. 
\begin{equation}\label{eqRBFSpaceWarping}
 \mathbf{p}^* = \bm{\Phi}(\mathbf{p}) = \bm{\alpha}_0 + A \mathbf{p} + \sum_{i=1}^N \bm{\beta}_i e^{-c\| \mathbf{p} - \mathbf{q}_i \|^2} 
\end{equation}
with $N$ being the number of kernels employed for the space warping and $A = [\bm{\alpha}_1, \bm{\alpha}_2, \bm{\alpha}_3]$.
\finalrev{The centers of Gaussian kernels, $\{ \mathbf{q}_i =\mathbf{B}(u_i,v_i,\mathcal{S}^c)\}$, are chosen by sparsely sampling across the UV domain to achieve comprehensive coverage of the simulated model's surface with the control points $\mathcal{S}^c$. The number of kernels for different robots will be provided in Table~\ref{tab:LearningData}.}
The value of coefficient $c$ controlling the width of the Gaussian kernel is chosen as $3.0 \times 10^{-5}$ by experiments. We remark that $\bm{\gamma} = [\bm{\alpha}_0, \bm{\alpha}_1, ...  \bm{\beta}_i]$ \finalrev{with $\bm{\alpha}_j,\bm{\beta}_i \in \mathbb{R}^3$} are variables of the function space \finalrev{to be determined via the learning process}. \JENew{Note that the same kernel on different simulated shapes will have the same UV parameters but different center-positions in $\mathbb{R}^3$.} 
\begin{figure}[t] 
\centering
\includegraphics[width=1.0\linewidth]{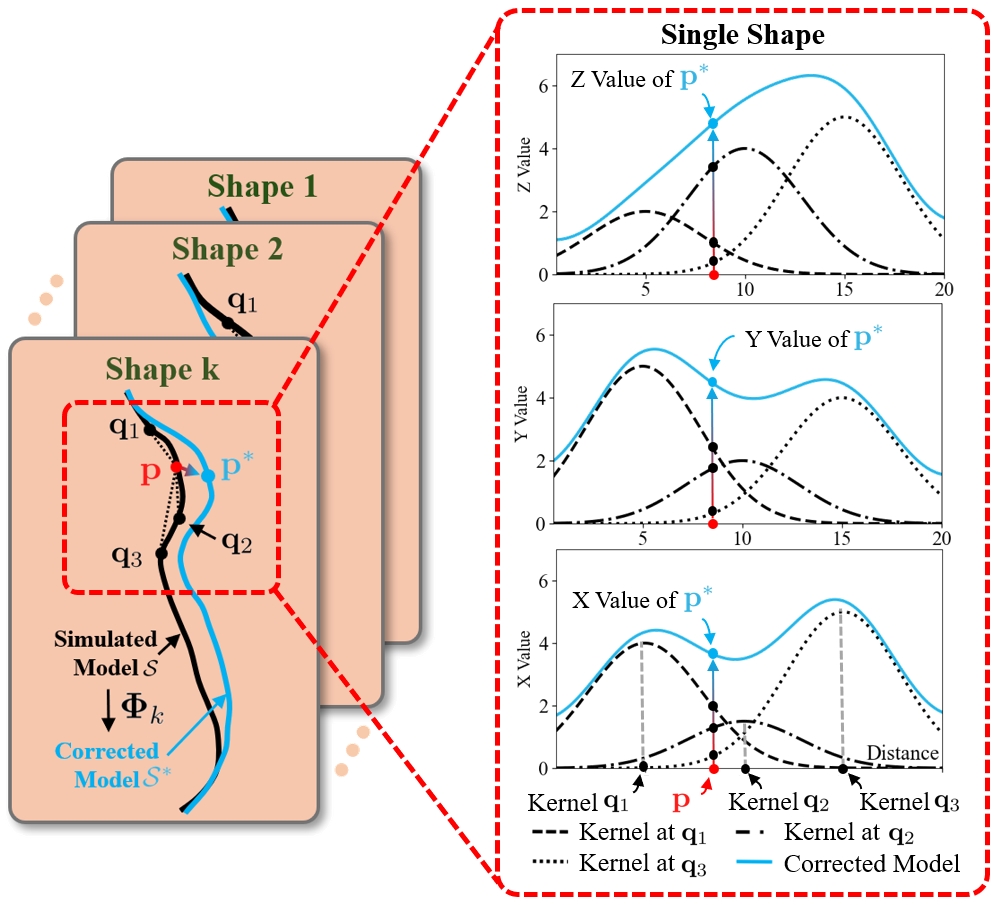}\\
\vspace{-2pt}
\caption{ 
The space warping $\mathbf{\Phi}(\cdot)$ is built on the \textit{radial basis functions} (RBF) with kernel centers located on the free-form surface of a simulated model. Different warping functions \JENew{with kernels having the same set of $(u,v)$-parameters} are used to correct different simulated shapes. 
}\label{fig:spaceWarping}
\end{figure}

With a determined warping function $\bm{\Phi}(\mathbf{p})$, we are able to map any point \finalrev{on the simulated surface $\mathcal{S}$ denoted as} $\mathbf{p} \in \mathcal{S}$ to its \finalrev{corrected position as} $\mathbf{p}^* \in \mathcal{S}^*$ -- see Fig.\ref{fig:spaceWarping} for the illustration. However, different warping functions \finalrev{-- by changing the values of $\gamma$ --} need to be employed for different simulated shapes. 
%
\JENew{Specifically, we introduce a network $\mathcal{N}_{rbf}$ to predict the function coefficients $\bm{\gamma}$ according to the simulated shape $\mathcal{S}$ as illustrated in Fig.\ref{fig:sim2RealPipelineOverview}.}

\subsection{NN-based Deformation Function Space}\label{subsecNNBasedDefField}
Our function-prediction based sim-to-real pipeline is formed by three building blocks: 1) the network $\mathcal{N}_{rbf}$, 2) the B-spline based shape decoder $\mathbf{B}(\cdot)$ and 3) the RBF-based space warping $\Phi(\cdot)$ (i.e., Eq.(\ref{eqRBFSpaceWarping})). The diagram has been shown in the top of Fig.\ref{fig:sim2RealPipelineOverview}. \finalrev{The purpose of network $\mathcal{N}_{rbf}$ is to generate shape-dependent RBF functions by determining their weights as
\begin{equation}\label{eqRBFSpaceNN}
    \bm{\gamma} = \mathcal{N}_{rbf}(\mathcal{S}^c)
\end{equation}
with its input as $M_u \times M_v$ control points of a simulated shape (denoted as $\mathcal{S}^c$) and the output $\bm{\gamma}=[\bm{\alpha}_0, \bm{\alpha}_1, ...  \bm{\beta}_i]$ as the collection of $N+4$ coefficient vectors in $\mathbb{R}^3$. 
}

When a deformable robot is actuated into various shapes $\{ \mathcal{S}_j \}$ by corresponding actuation $\{ \mathbf{a}_j \}$, the function-prediction network $\mathcal{N}_{rbf}$ \finalrev{can be} trained on pairs of shape differences between the shape $\mathcal{S}_j^*$ predicted by the sim-to-real network $\mathcal{N}_{s2r}$ and the physically captured shape $\mathcal{S}_j^p$. More specifically, for the $j$-th \finalrev{deformation}, the surface of the simulated shape $\mathcal{S}_j$ is sampled as a set of points $\{(u_i, v_i)\}$ in its parametric domain. After applying sim-to-real correction, the resultant surface $\mathcal{S}_j^*$ is represented as a point set $\mathcal{S}_j^* = \{\mathbf{p}^*_i\}$, where each point \finalrev{can be computed by}
\begin{equation} \label{eqSim2RealPrediction}
\finalrev{
\mathbf{p}_i^*=\Phi_{\bm{\gamma}}(\mathbf{p}) = \Phi_{\mathcal{N}_{rbf}(\mathcal{S}_j^c)}\left( \mathbf{B}(u_i,v_i,\mathcal{S}_j^c) \right).
}
\end{equation}
The warping function $\bm{\Phi}_j$ is parameterized by $\bm{\gamma}$, which is predicted by the network as $\bm{\gamma} = \mathcal{N}_{rbf}(\mathcal{S}^c_j)$. The corresponding physically acquired shape $\mathcal{S}_j^p$ is also represented as a point set $\{\mathbf{p}^{p}_k\}$. \finalrev{For training the network $\mathcal{N}_{rbf}$, the shape differences between all pairs of point clouds as $S_j^* =\{\mathbf{p}_k^*\}$ and $S_j^p=\{\mathbf{p}^{p}_k \}$ are employed to define loss functions to be minimized. Note that our loss functions are defined in a way that does not require pre-defined correspondences between two point clouds and also tolerate missed regions. Details will be presented in Sec.~\ref{secDiffRegTraining}.}


Given a well trained network $\mathcal{N}_{rbf}(\cdot)$, the sim-to-real transfer can be evaluated by the pipeline at the top of Fig.\ref{fig:sim2RealPipelineOverview} in the inference phase. \finalrev{
For any point $(u_p,v_p)$ sampled in the UV domain on a simulated shape compactly encoded as a set of control points $\mathcal{S}^c$, we can obtain its predicted position by Eq.\eqref{eqSim2RealPrediction}. By modeling $\mathcal{S}^c$ as a neural network–based function of the actuation vector $\mathbf{a}$, the above sim-to-real transfer process can be interpreted} 
as a network $\mathcal{N}_{s2r}(\cdot)$ that gives
\begin{equation}\label{eqSim2RealPrediction2}
\mathbf{p}^*=\mathcal{N}_{s2r}(\mathbf{B}(u_p,v_p,\mathcal{S}^c(\mathbf{a})),\mathcal{S}^c(\mathbf{a}),\{ \mathbf{B}(u_q,v_q,\mathcal{S}^c(\mathbf{a})\}).
\end{equation}
All operators in this sim-to-real pipeline is differentiable w.r.t. the actuation parameter $\mathbf{a}$. \JENew{Detail analysis of differentiability can be found in Appendix \ref{AppendixDifferentiation}}. In Sec.~\ref{secNNShapeCtrl}, we will present the detail of NN-based forward kinematics and the IK computing for shape control, which is based on minimizing the distance between $\mathbf{p}^*$ and its closest point $\mathbf{c}^*$  on the target shape $\mathcal{S}^t$.

\section{Differentiable Alignment for Training}\label{secDiffRegTraining}
To enable correspondence-free training of the \finalrev{function-prediction network $\mathcal{N}_{rbf}$}, we introduce a confidence-map weighted Chamfer distance as a robust metric for evaluating discrepancies between the predicted shape $\mathcal{S}_j^*$ and the captured shape $\mathcal{S}_j^p$. This weighted Chamfer distance serves as the primary loss for aligning $\mathcal{S}_j^*$ with $\mathcal{S}_j^p$ during training. The process jointly optimizes  $\mathcal{N}_{rbf}$ and \finalrev{a newly introduced} confidence-prediction network $\mathcal{N}_{conf}$, aided by additional regularization terms -- namely, confidence regularization, normal compatibility, and geometric regularization. Finally, we describe how this framework can be adapted to scenarios involving partially captured markers with known correspondences.

\subsection{Confidence Weighted Chamfer Distance}\label{subsec:ConfWeightedChamferDist}
\JENew{
Chamfer distance has been widely used in previous works for rigid registration (e.g., \cite{yew2020rpm}). In contrast, our task involves non-rigid registration on models with large deformations, which presents additional challenges. Moreover, the captured shapes of free-form surfaces undergoing large deformations often contain missing regions. To address this, we introduce a real-valued confidence map $w_c(\cdot) \in [0,1]$ on the surface of each simulated shape $\mathcal{S}_j$, representing the likelihood of finding reliable correspondences on the scanned shape $\mathcal{S}_j^p$. 
}

\JENew{
Without loss of generality, this confidence map is parameterized by a neural network $\mathcal{N}_{conf}$, whose parameters are \finalrev{determined jointly} during the sim-to-real training process. The input to $\mathcal{N}_{conf}$ includes a surface point $\mathbf{p} \in \mathcal{S}$ and the set of control points $\mathcal{S}^c(\mathbf{a})$ that encode the simulated shape, allowing the network to adapt to varying shape configurations. The network outputs a scalar in the range $[0,1]$, representing the confidence that the point $\mathbf{p}$ has a reliable correspondence on the scanned shape $\mathcal{S}_j^p$ -- i.e., $w_c(\mathbf{p}) = \mathcal{N}_{conf}(\mathbf{p}, \mathcal{S}^c(\mathbf{a}))$. 
}

\JENew{
The Chamfer distance between the predicted shape $\mathcal{S}_j^*$ and the captured shape $\mathcal{S}_j^p$ can then be weighted by the confidence map $w_c(\cdot)$ as follows:
\begin{align}\label{eqConfWeightedChamferDist}
    \Tilde{D}(\mathcal{S}_j^*,\mathcal{S}_j^p) &=  \frac{1}{|\mathcal{S}_j|}\sum_{\mathbf{p} \in \mathcal{S}_j} w_c(\mathbf{p}) \min_{\mathbf{x} \in \mathcal{S}_j^p}  \| \mathbf{p}^* - \mathbf{x} \|_2^2 \nonumber
\\ & + \frac{1}{|\mathcal{S}_j^p|}\sum_{\mathbf{x} \in \mathcal{S}^p_j} \min_{\mathbf{p}^* \in \mathcal{S}_j^*} \| \mathbf{x} - \mathbf{p}^* \|_2^2,
\end{align}
where $\mathbf{p}^* \in \mathbb{R}^3$ is the sim-to-real transferred surface point corresponding to $\mathbf{p}=\mathbf{B}(u_p,v_p,\mathcal{S}^c(\mathbf{a}))$, as defined in Eq.~\eqref{eqSim2RealPrediction}. Here, $|\cdot|$ denotes the number of sample points on surface. 
}
\finalrev{As only the scanned shape $\mathcal{S}^p_j$ may have missed regions, the confidence map is not applied to the closest point search from the physically sampled geometry $\mathcal{S}^p_j$ to the predicted shape $\mathcal{S}^*_j$ that is assumed to be complete.}

\subsection{Loss Functions for Training}\label{subsec:LossFunc}
\JENew{
The loss function for shape alignment based training is defined by the confidence weighted Chamfer distances as
\begin{equation}\label{eq:chamferDistLoss}
    \mathcal{L}_{cd} = \sum_{j} \Tilde{D}(\mathcal{S}_j^*,\mathcal{S}_j^p).
\end{equation}
Besides of this primary loss, other three loss terms are defined to serve for purpose of  regularization.
}

\begin{figure}
\centering
\vspace{-5pt}
\includegraphics[width=0.97\linewidth]{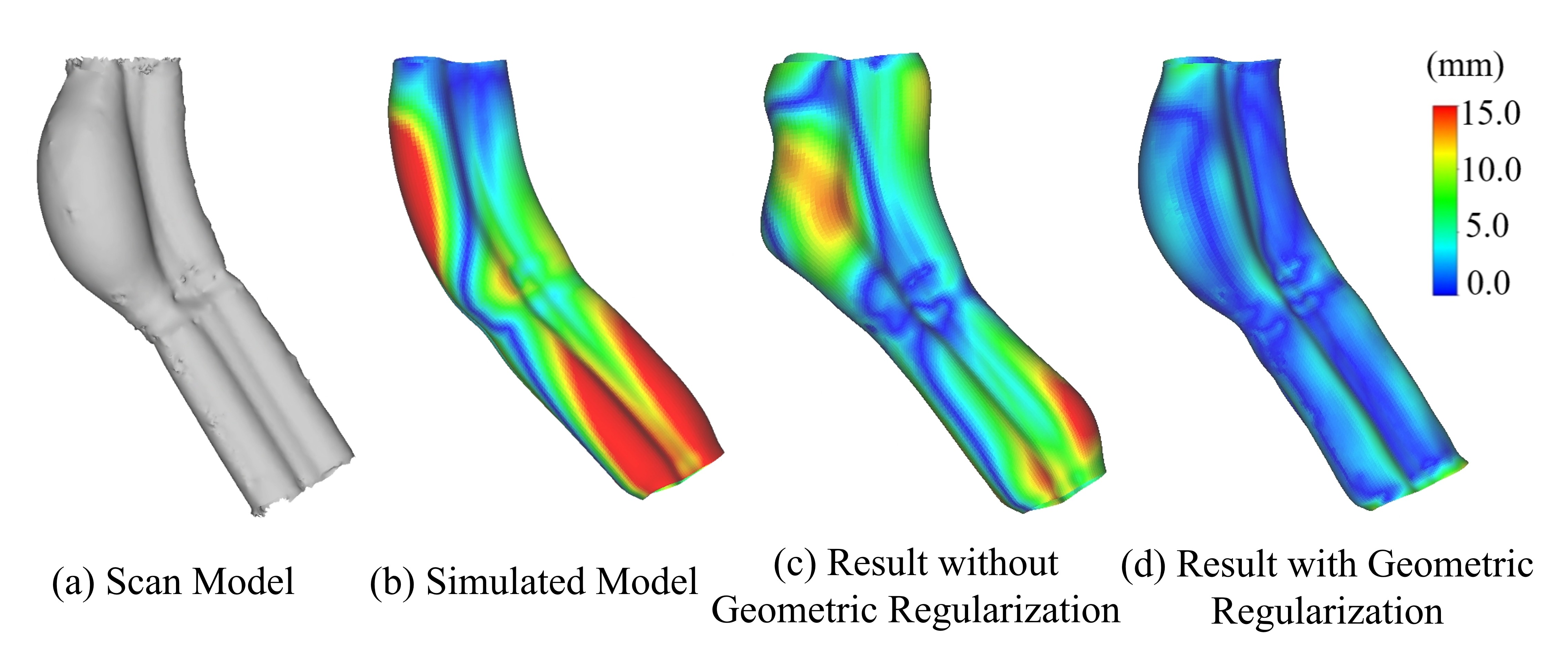}
\caption{
\JENew{When there is a large gap between (a) real and (b) simulated shapes, the RBF based space warping may generate a weird shape as shown in (c). This problem can be effectively solved by introducing the function compatibility requirement as a geometric regularization during training -- see (d) for the sim-to-real result with this additional regularization. The color map illustrates the distribution of geometric errors.}
}\label{fig:geoRegLossAblation}
\end{figure}

\subsubsection{Confidence Regularization}
\JENew{
To prevent trivial solutions where the confidence map $w_c(\cdot)$ is zero across the entire set of surfaces, we introduce a confidence regularization loss: 
\begin{equation}\label{eq:confRegLoss}
    \mathcal{L}_{cr} := \sum_{j} \sum_{\mathbf{p} \in \mathcal{S}_j} (1 - w_c(\mathbf{p}) )^2.
\end{equation}
}

\subsubsection{Normal Compatibility}
\JENew{
To ensure the normals of a predicted surface $\mathcal{S}_j^*$ compatible with those of the captured surface $\mathcal{S}_j^p$, we introduce a normal compatibility loss:
\begin{equation}\label{eq:normCompLoss}
    \mathcal{L}_{nc} := \sum_{j} \sum_{\mathbf{x} \in \mathcal{S}_j^p} (1 - \mathbf{n}(\mathbf{x}) \cdot \mathbf{n}(\mathbf{c}(\mathbf{x})) ),
\end{equation}
where $\mathbf{c}(\mathbf{x})$ denotes $\mathbf{x}$'s closest point on $\mathcal{S}_j^*$, and $\mathbf{n}(\cdot)$ represents the unit surface normal. Notably, we perform a one-way search from the captured shape $\mathcal{S}_j^p$ to the predicted shape $\mathcal{S}_j^*$, which is assumed to be complete. This asymmetry improves robustness in cases where the captured surface contains large holes or missing regions.
}

\subsubsection{Geometric Regularization}
\JENew{
The shape produced by the RBF-based warping function may appear weird when the number of kernels is limited and the discrepancy between the simulated and captured shapes is large. In such cases, the learned high-dimensional function space may fail to generate smooth deformations without appropriate regulation, as illustrated in Fig.\ref{fig:geoRegLossAblation}(a–c). Similar challenges have been observed in previous RBF-based deformation studies (ref.~\cite{Charlie2007RBFVolPara,turk2002modelling}). To address these issues and improve control over the resulting deformed 3D shapes, prior research imposed the following compatibility conditions.
\begin{align}
\label{eqCompatibilityCondition}\sum_{i=1}^N \bm{\beta}_{i} 
=\sum_{i=1}^N \bm{\beta}_{i} \left(\mathbf{q}_{i}\right)_{x}
&=\sum_{i=1}^N \bm{\beta}_{i} \left(\mathbf{q}_{i}\right)_{y}
=\sum_{i=1}^N \bm{\beta}_{i} \left(\mathbf{q}_{i}\right)_{z}
= \mathbf{0}
\end{align}
To impose similar constraint on the deformation function predicted by the network $\mathcal{N}_{rbf}$, we formulate the compatibility conditions of Eq.\eqref{eqCompatibilityCondition} as a geometric regularization loss:
\begin{align}\label{eqCompatibilityLoss}
    \mathcal{L}_{gr} &:=  \sum_j
    \|\sum_{i=1}^N \bm{\beta}_{i}\|^{2} +\sum_j\|\sum_{i=1}^N\bm{\beta}_{i} (\mathbf{q}_{i})_{x}\|^{2} \nonumber
\\ & + \sum_j\|\sum_{i=1}^N\bm{\beta}_{i} (\mathbf{q}_{i})_{y}\|^{2} + \sum_j\|\sum_{i=1}^N\bm{\beta}_{i} (\mathbf{q}_{i})_{z}\|^{2}.
\end{align}
The effectiveness of this geometric regularization term has been demonstrated in Fig.\ref{fig:geoRegLossAblation}(c, d).
}

\JENew{
In summary, the total loss for training is defined as
\begin{equation}\label{eq:totalLoss}
    \mathcal{L}_{total} := \mathcal{L}_{cd} + \omega_{cr} \mathcal{L}_{cr}  + \omega_{nc} \mathcal{L}_{nc}  + \omega_{gr} \mathcal{L}_{gr}
\end{equation}
which combines the aforementioned loss terms with corresponding weights. The function-prediction network $\mathcal{N}_{rbf}$ and the confidence-prediction network $\mathcal{N}_{conf}$ are jointly \finalrev{trained} by minimizing $\mathcal{L}_{total}$ (see also Fig.\ref{fig:sim2RealPipelineOverview} for the illustration of training pipeline). The weight coefficients $\omega_{cr}=12.0$, $\omega_{nc}=0.5$, and $\omega_{gr}=\text{1.5e-4}$ are empirically selected based on experiments and remain fixed across all examples. 
}

\subsection{Training with Known Correspondences}\label{subsec:trainingWithMarkers}
\JENew{
In scenarios where hardware such as a MoCap system is used for sim-to-real learning, the captured surface $\mathcal{S}_j^p$ is represented by a set of markers ${\mathbf{x}_i}$ with known correspondences. This allows the training process to be simplified. Specifically, for each marker, its corresponding surface sample point $(u_i, v_i)$ in the parametric domain can be obtained through registration with the deformable surface in its rest shape. Consequently, the set of simulated surface points becomes $\mathcal{S}_j = \{\mathbf{p}_i = \mathbf{B}(u_i, v_i, \mathcal{S}^c_j)\}$. 
}

\JENew{
Training proceeds similarly to the correspondence-free case, with the following modifications:}
\begin{itemize}

    \item \JENew{Replace the nearest-neighbor search in Eq.\eqref{eqConfWeightedChamferDist} with the known correspondence term $\|\mathbf{p}_i^* - \mathbf{x}_i \|_2^2$}\finalrev{, which gives
\begin{equation}
    \Tilde{D}(\mathcal{S}_j^*,\mathcal{S}_j^p) =  \frac{1}{|\mathcal{S}_j|}\sum_{\mathbf{p}_i \in \mathcal{S}_j} w_c(\mathbf{p}_i)  \| \mathbf{p}_i^* - \mathbf{x}_i \|_2^2;
\end{equation}
}
    \item \JENew{Set $w_c(\mathbf{p}_i) = 1$ if marker $\mathbf{x}_i$ is captured in the $j$-th frame, or $w_c(\mathbf{p}_i) = 0$ when it is missing;}

    \item \JENew{Assign $\omega_{nc} = \omega_{cr} = 0.0$, as normal compatibility and confidence regularization are no longer necessary with known correspondences.}
\end{itemize}
\JENew{
The effectiveness of this simplified sim-to-real training process using a MoCap system will be demonstrated in Sec.\ref{secResult} (see the right of Fig.\ref{fig:SamplePntDensityStudy}(b)).
}
\section{Neural Networks Based Kinematic Computing}\label{secNNShapeCtrl}
This section presents the NN-based computational pipeline for kinematics. \finalrev{With the help of a sim-to-real network trained above, accurate} shape control of deformable free-form surfaces can be realized by our fast IK solution. 
\begin{figure}[t]
\centering
\includegraphics[width=1.0\linewidth]{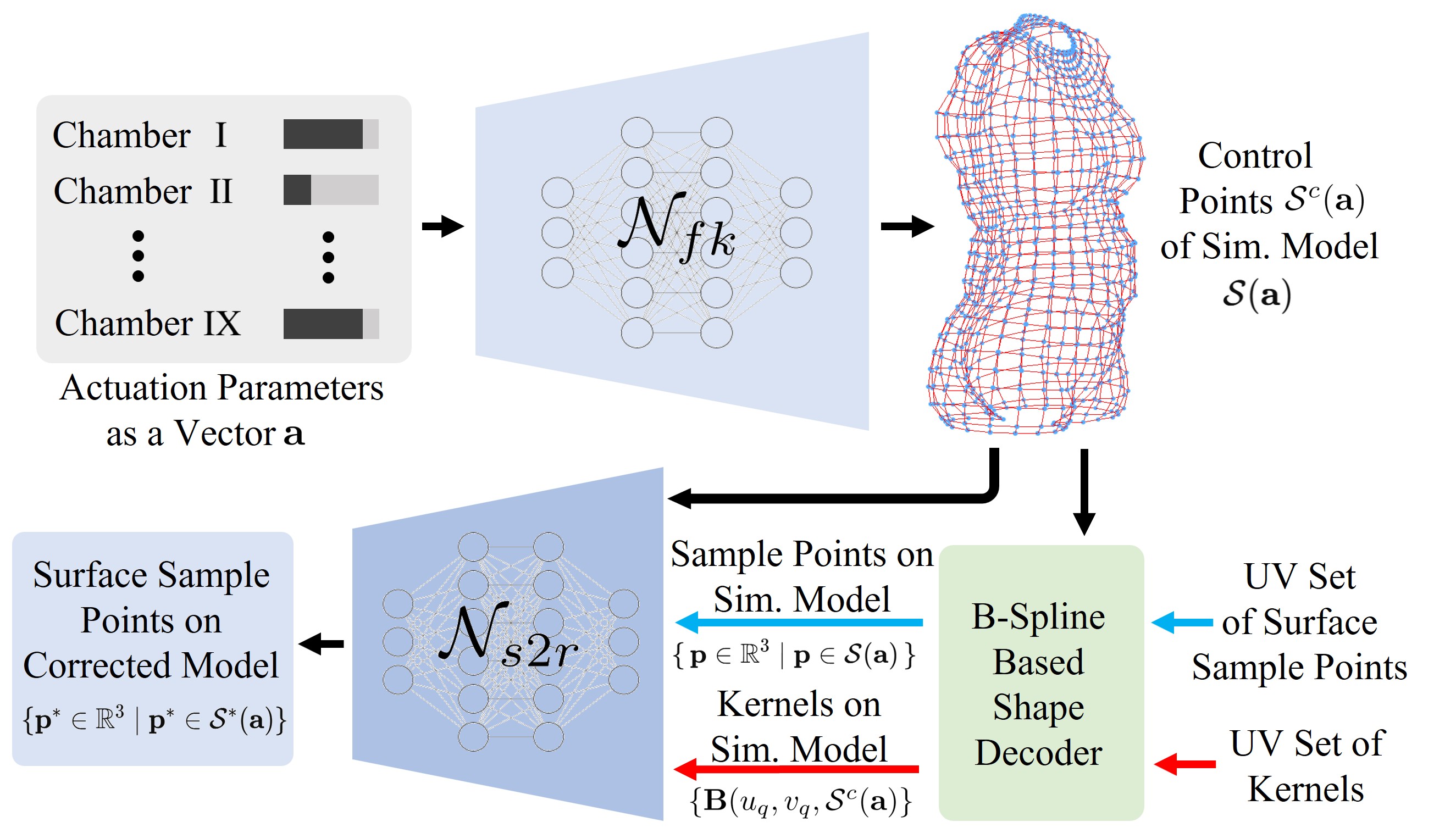}
\caption{NN-based pipeline for forward kinematic computing\finalrev{, by which the accurate shape of a deformable robot represented as a set of 3D point $\{\mathbf{p}^* \in \mathbb{R}^3\}$ can be predicted by the input actuation parameters as a vector $\mathbf{a}$}.
}\label{fig:NNBasedFK}
\end{figure}

\begin{algorithm}[t]
\caption{\finalrev{NN-Based Forward Kinematics}}
\label{alg:NNBasedFK}
\LinesNumbered

\KwIn{\finalrev{The actuation vector $\mathbf{a}$.}}

\KwOut{\finalrev{The deformed shape $\mathcal{S}^*$ as a set of points $\{ \mathbf{p}^*\}$.}}

\finalrev{
Compute the simulated shape as control pnts. $\mathcal{S}^c$ by Eq.\eqref{eqActuateToCtrlPnts};}

\finalrev{
Determine the coefficients of RBF function as $\bm{\gamma}$ by Eq.\eqref{eqRBFSpaceNN};}

\For{every sample point $(u_p,v_p)$}{
    \finalrev{Compute its position as $\mathbf{p}=\mathbf{B}(u_p,v_p,\mathcal{S}^c)$ by Eq.\eqref{eqBSplineFunc};} 
    
    \finalrev{Update its position as $\mathbf{p}^*=\Phi_{\bm{\gamma}}(\mathbf{p})$ by Eq.\eqref{eqSim2RealPrediction};}
}

\KwRet{\finalrev{$\mathcal{S}^* = \{\mathbf{p}^* \}$}};

\end{algorithm}

\subsection{NN-based Forward Kinematics}\label{subsecNNBasedFK}
While the recently developed fast simulator provides an efficient solution to predict the deformed shape of soft robots~\cite{Guoxin2022IROS}, applying it to estimate gradients of the IK \finalrev{computation} by numerical differentiation remains time-consuming~\cite{Guoxin2020TRO}. \finalrev{To enable fast IK computation,} we propose to employ an NN-based computational pipeline for forward kinematics \finalrev{as shown in Fig.\ref{fig:NNBasedFK}}. 
The input of our \finalrev{shape prediction} network \finalrev{$\mathcal{N}_{fk}$} is the actuation parameter $\mathbf{a}$, and the output is the control points $\{ \mathcal{S}^c \}$ as a compact representation of freeform surfaces. \finalrev{This gives
\begin{equation}\label{eqActuateToCtrlPnts}
    \mathcal{S}^c = \mathcal{N}_{fk}(\mathbf{a}).
\end{equation}
The network $\mathcal{N}_{fk}$ actually serves as a surrogate model for the numerical simulation of soft robot's deformation.} 

\finalrev{The network can be trained via a supervised learning process.} Given a set of randomly generated actuation parameters as $\{ \mathbf{a}_k \}$, we can run the numerical simulators to obtain the deformed shapes and generate their corresponding control points as $\{ \mathcal{S}^c_k \}$. A training dataset with $M$ such pairs of results $\{ \mathbf{a}_k : \mathcal{S}^c_k \}_{k=1,\ldots,M}$ can be obtained by a simulator (e.g.,~\cite{Guoxin2022IROS}). It is important to generate \finalrev{example} shapes that span the entire space of shape variation. The dataset is then employed to train the network $\mathcal{N}_{fk}$.

\finalrev{
The computation of forward kinematics is conducted with the help of $\mathcal{N}_{fk}$ and $\mathcal{N}_{s2r}$. Given an actuation $\mathbf{a}$, the control points of its simulated surface $\mathcal{S}^c$ can be obtained by Eq.\eqref{eqActuateToCtrlPnts}.
The corresponding coefficients $\bm{\gamma}$ of the RBF function for sim-to-real transfer can then be obtained by Eq.\eqref{eqRBFSpaceNN}.
}
As a result, for any point \JENew{$(u_p,v_p)$} sampled in the $u,v$-parametric domain, its position on the physical model can be predicted by Eq.\eqref{eqSim2RealPrediction}. This NN-based computational pipeline for forward kinematics was illustrated in Fig.\ref{fig:NNBasedFK} \finalrev{-- see also the pseudo-code given in Algorithm \ref{alg:NNBasedFK}}. The predicted shape $\mathcal{S}^*(\mathbf{a})$ is represented by a set of points $\{ \mathbf{p}^* \}$, which is a differentiable function in terms of the actuation $\mathbf{a}$.

\begin{figure}[t]
\centering
\includegraphics[width=1.0\linewidth]{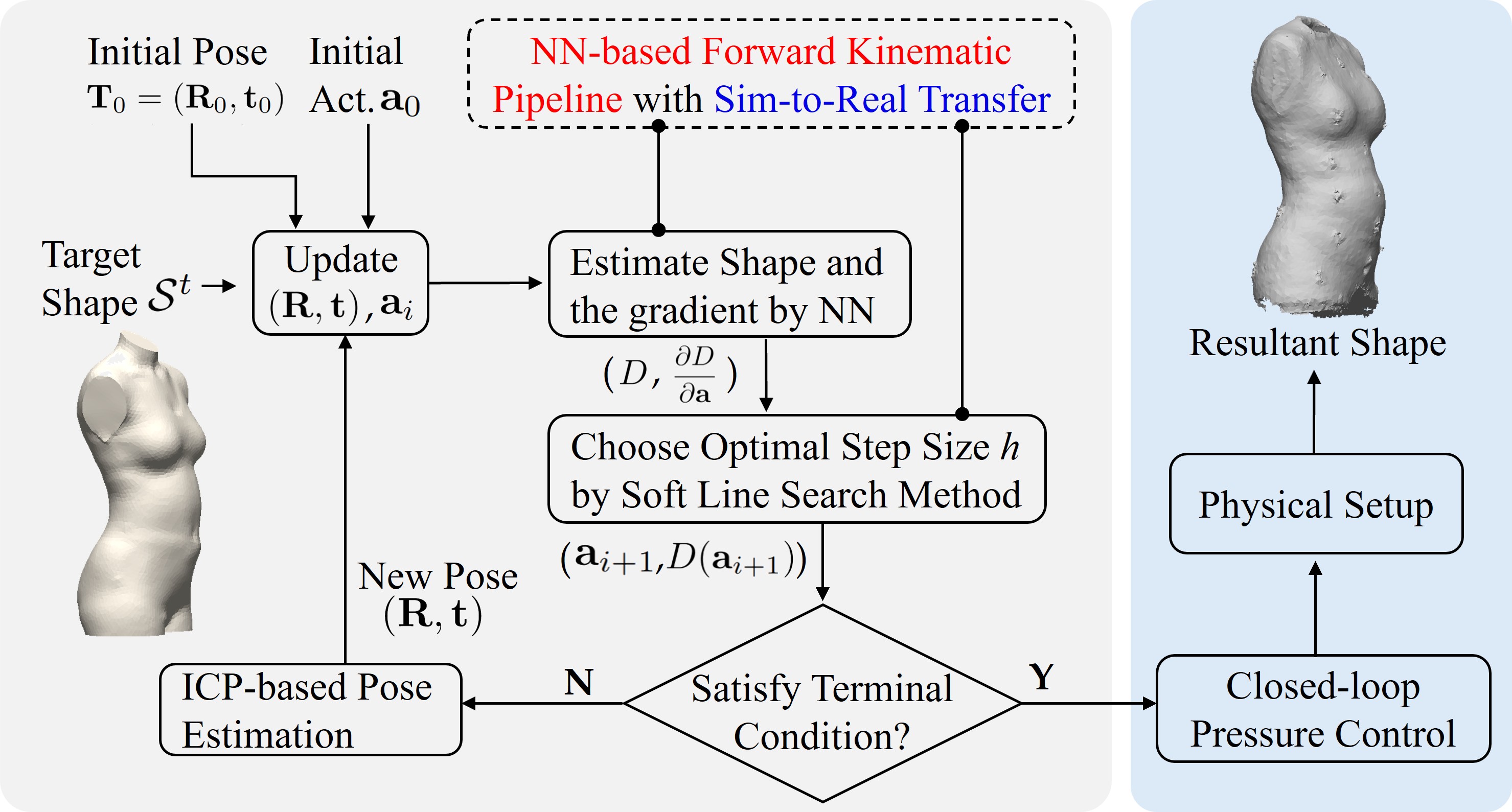}
\caption{The diagram of our fast IK solver, \JENew{where the NN-based forward kinematics pipeline is developed with our sim-to-real method}.
}\label{fig:IK-Pipeline}
\end{figure}

\begin{algorithm}[t]
\caption{\finalrev{Fast Inverse Kinematics Solver}}
\label{alg:FastIK}
\LinesNumbered

\KwIn{The target shape $\mathcal{S}^{t}$, the threshold $\lambda$ and the maximally allowed steps $i_{max}$ for termination.}

\KwOut{The optimized actuation $\mathbf{a}_{opt}$.} 

Set $i = 0$ and $r_0 = 1.0$; 

Set the initial actuation $\mathbf{a_0}$ and the initial transformation matrix $\mathbf{T}_{0} = (\mathbf{R}_0, \mathbf{t}_0)$;

Apply the transformation $\mathbf{T}_{0}$ to the target model $\mathcal{S}^{t}$;

\While{$i<i_{max}$ and $r_{i}>\lambda$}
{
    \tcc{Update transformation matrix}
    Apply the ICP-based pose estimation to update $\mathbf{T}_{i}$ (applied on $\mathcal{S}^{t}$) to better align with the corrected model $\mathcal{S}^{*}$;

    \tcc{Gradient-based iteration with fixed transformation matrix}
    Evaluate the objective function $D_i =D(\mathbf{a}_{i})$;

     Compute the gradient of $D(\mathbf{a}_{i})$ as $\frac{\partial D}{\partial \mathbf{a}} $ by Eq.\eqref{eqDiffD_Diffa};

    Compute the optimal step size $h$ by line search \cite{nocedal1999numerical};

    Set $\mathbf{a}_{i+1} = \mathbf{a}_{i} - h \frac{\partial D}{\partial \mathbf{a}}$ and $i=i+1$;

    Evaluate the objective function $D_i = D(\mathbf{a}_{i})$;

    Compute the relative decreasing percentage as: $r_{i} =\frac{\abs{D_{i}-D_{i-1}}}{\abs{D_{i}}}$;


}
\KwRet{$\mathbf{a}_{opt} = \mathbf{a}_{i}$};

\end{algorithm}

\subsection{Gradient Iteration Based Inverse Kinematics}\label{subsecGradientBasedIK}
\finalrev{Following the strategy proposed in \cite{Guoxin2020TRO}, the} IK computation is formulated as an optimization problem that minimizes the shape difference between the predicted shape $\mathcal{S}^*$ and the target shape $\mathcal{S}^t$ \finalrev{while allowing the pose change}. That gives 
\begin{equation}\label{eqIKObjFunc}
    \arg \min_{\mathbf{a},\mathbf{R},\mathbf{t}} D(\mathcal{S}^*(\mathbf{a}), \mathcal{S}^t) = 
        \sum_{\mathbf{p}^* \in \mathcal{S}^*(\mathbf{a})} \|\mathbf{p}^* -  \left(\mathbf{R} \mathbf{c}^* + \mathbf{t} \right) \|^2,
\end{equation}
where $\mathbf{c}^*$ is the closet point of $\mathbf{p}^*$ on the target shape $\mathcal{S}^t$. The rotation matrix $\mathbf{R}$ and the translation vector $\mathbf{t}$ are applied to the target model to eliminate the influence of pose change. The objective function can be minimized using the gradient descent method with linear search. After each iteration of updating $\mathbf{a}$, we apply an ICP-based rigid registration \cite{ICP1992} to determine a new pair of $(\mathbf{R},\mathbf{t})$. In other words, the optimization process alternates between updating the actuation parameter $\mathbf{a}$ and adjusting the pose of target model $(\mathbf{R},\mathbf{t})$\footnote{\finalrev{For cases where the target shape’s pose is fixed (e.g., the soft manipulator examples), the registration step is omitted by setting $(\mathbf{R} = \mathbf{I}, \mathbf{t} = \mathbf{0})$.}}. The diagram of our fast IK solver is as illustrated in Fig.\ref{fig:IK-Pipeline}, and the pseudo-code of our IK solver can be found in \finalrev{Algorithm \ref{alg:FastIK}}.

Thanks to the formulation of NN-based forward kinematics \finalrev{including sim-to-real}, the gradients of $D(\cdot)$ in Eq.\eqref{eqIKObjFunc} can be computed analytically. That is
\begin{flalign}
\label{eqDiffD_Diffa}
    \frac{\partial D}{\partial \mathbf{a}} &=
    \sum_{\mathbf{p}^* \in \mathcal{S}^*(\mathbf{a})} 2 (\frac{\partial \mathbf{p}^*}{\partial \mathbf{a}})^{T}
   (\mathbf{p}^* -  \left(\mathbf{R} \mathbf{c}^* + \mathbf{t} \right)),
\end{flalign}
where 
\begin{flalign}
\label{eqDiffp_Diffa}
\frac{\partial \mathbf{p}^*}{\partial \mathbf{a}} &= \frac{\partial \mathcal{N}_{s2r}}{\partial \mathbf{B}} 
\frac{\partial \mathbf{B}}{\partial \mathbf{a}} + \frac{\partial \mathcal{N}_{s2r}}{\partial \mathcal{N}_{fk}}
\frac{\partial \mathcal{N}_{fk}}{\partial \mathbf{a}} +  \frac{\partial \mathcal{N}_{s2r}}{\partial \mathbf{Q}} \frac{\partial \mathbf{Q}}{\partial \mathbf{a}}
\nonumber
\\&= \left(\frac{\partial \mathcal{N}_{s2r}}{\partial \mathbf{B}} 
\frac{\partial \mathbf{B}} {\partial \mathcal{N}_{fk}}
+ \frac{\partial \mathcal{N}_{s2r}}{\partial \mathcal{N}_{fk}} + \frac{\partial \mathcal{N}_{s2r}}{\partial \mathbf{Q}}  \frac{\partial \mathbf{Q}} {\partial \mathcal{N}_{fk}}\right)\frac{\partial \mathcal{N}_{fk}}{\partial \mathbf{a}}.
\end{flalign}
$\mathbf{Q}$ denotes those terms contributed by the set of kernels for RBF-base warping while $\mathbf{B}$ is for the surface point $\mathbf{p}$. \JENew{Detail formulas can be found in Appendix \ref{AppendixDifferentiation}.}

\section{Implementation Details and Results}\label{secResult}
We have implemented the proposed approach in C++ and Python. The network training phase is implemented on the PyTouch platform with the learning rate at $0.001$ and the maximum number of epochs as $150$. For the \JENew{inference} phase, we transferred the trained networks to integrate with our IK solver in C++ running on CPU. The analytical gradients of the networks are employed in our IK solver. Our source code will be released on GitHub upon the acceptance of this paper at: \url{https://yinggwan.github.io/CFS2R.github.io}. All the training and computational tests are conducted on a PC with Intel i7-12700H CPU, RTX $3060$ GPU and $32$ GB RAM. 

\begin{figure}[t] 
\centering
\includegraphics[width=1.0\linewidth]{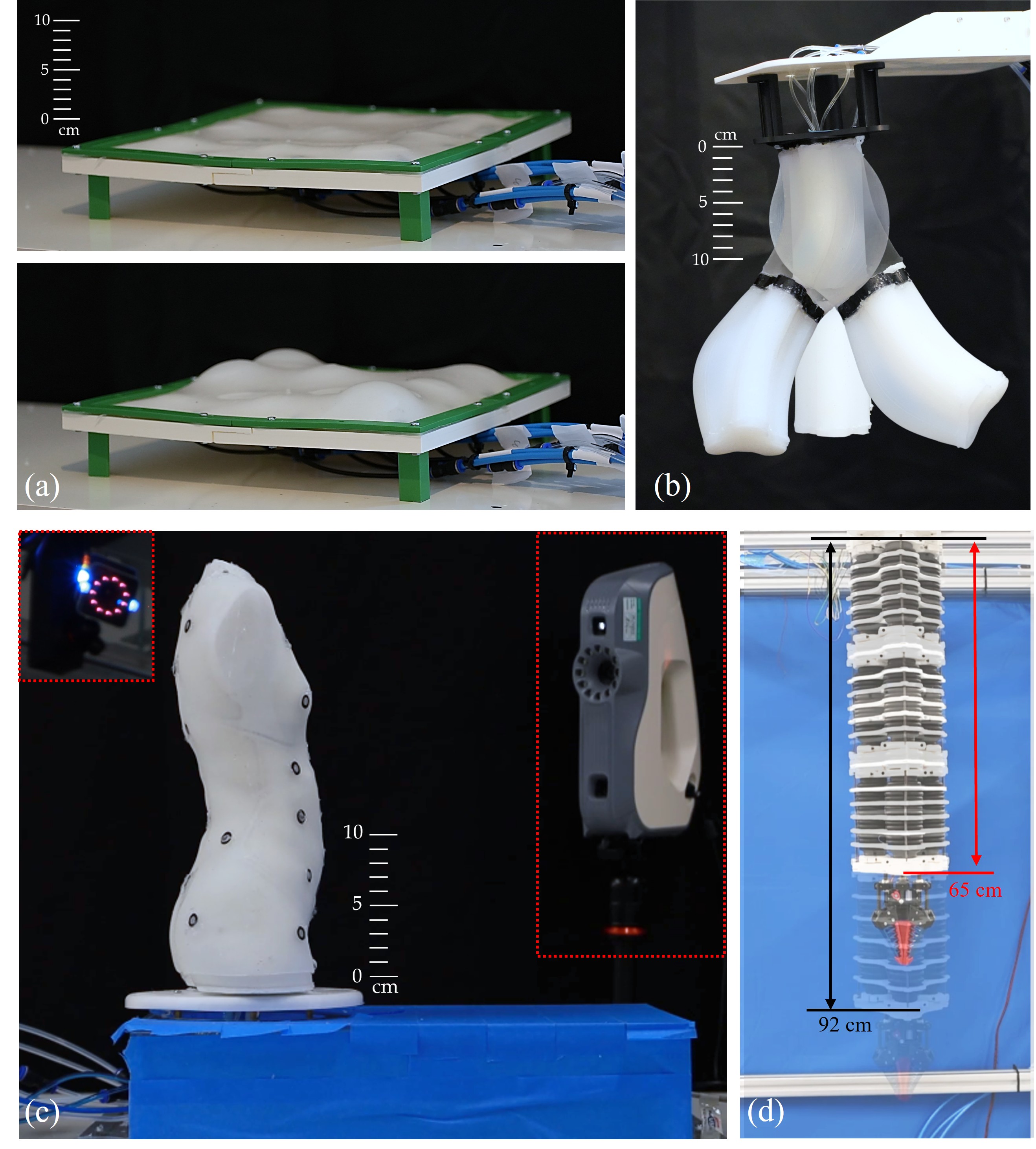}
\caption{
\JENew{Four pneumatic actuated soft robots with large deformation are conducted to validate the effectiveness of our sim-to-real approach, including (a) a deformable membrane with 9 chambers, (b) a soft manipulator with two segments -- 3 inflation chambers per segment, (c) a soft robotic mannequin having 15 chambers with 9 independent DoFs, and (d) an advanced soft manipulator with $3 \times 3$ bellows that can undergo substantial compression, inflation, and rotational deformations. Both the Mocap system and the 3D scanner highlighted by red dash lines were used in our tests.}
}\label{fig:HardwareSystem}
\end{figure}

\subsection{Hardware for Verification}\label{subsecResHardware}
\JENew{To evaluate the generality of our function-based sim-to-real pipeline, we tested it on four different soft robots as shown in Fig.\ref{fig:HardwareSystem}), where an open source pneumatic platform -- OpenPneu~\cite{tian2023openpneu} -- was used to actuate these robots with well-controlled pressures. In addition, two types of vision systems, a Vicon motion capture system~\cite{Vicon} with 8 cameras and an Artec Eva 3D Scanner~\cite{Scanner}, were used for data acquisition. Note that the scanner is also used for capturing ground-truth shapes for validation.}

\subsubsection{Deformable Membrane}\label{secDefMembrane}
\JENew{The first soft robot as shown in Fig.\ref{fig:HardwareSystem}(a) consists of nine chambers that can be inflated independently by pneumatic actuation. Each chamber is covered by a silicone membrane with thickness as 2mm. The dimension of the membrane is 32.0 cm (L) x 32.0 cm (W). By adjusting combination of air pressures across the nine chambers, the deformable membrane can be shaped into a variety of free-form surfaces~\cite{scharff2021sensing}.}

\subsubsection{Soft Manipulator}\label{secSoftManipulator}
\JENew{The second soft robot for this research is a two-segment soft manipulator \cite{marchese2016design}, with each segment containing three chambers. The total length of the manipulator is 26.0 cm. Rotational deformation in its global structure produces significant shape changes across the robot’s surface, as illustrated in Fig.\ref{fig:HardwareSystem}(b). In existing methods (e.g.,~\cite{webster2010design}), the deformed states of this robot are simplified by only modeling the global structure shape, disregarding the substantial deformation of its outer curved surface. This inaccuracy will lead to the difficulty of realizing the planned collision-free motion on a physical setup, which can be effectively solved by using our sim-to-real method.}

\subsubsection{Deformable Mannequin}
\label{secSoftMannequin}
The third robot, shown in Fig. \ref{fig:HardwareSystem}(c), contains a soft robotic mannequin representing both the front and back sides of a human body at a 1:2 scale. \JENew{The total height of the mannequin is 31.5 cm.} It can be deformed by pressurized air and features a design with three layers: an inner soft chamber layer attached to the base solid core, a middle soft chamber layer attached on top of the inner layer, and a top membrane layer (passively morphed) covering the base layers to form the overall smooth shape. \JENew{Detailed design can be found in \cite{Tian2022SoRoMannequin,Tian-RSS-24}.}  By pumping air at varying pressures into the chambers, the soft mannequin can be deformed into different shapes. 

\subsubsection{Compressible Soft Manipulator}
\label{secFlexPAL}
\JENew{
The fourth robot is an advanced soft manipulator composed of three segments, each containing three soft-bellow actuators as shown in Fig.~\ref{fig:HardwareSystem}(d). By adjusting the internal pressure of each actuator, the soft robot can elongate or contract through chamber inflation (positive pressure) or compression (negative pressure) while simultaneously performing deformations in bending. For example, the robot can change its length from a natural state of 92.0 cm down to 65.0 cm. The sim-to-real approach proposed in this paper can accurately model the robot’s entire external surface, which is essential for enabling collision-free motion planning. 
}

The four hardware setups were selected to span representative deformation modes and structural complexities in pneumatically actuated soft robots. The deformable membrane provides inflation-induced deformations with regular chamber geometry, while the robotic mannequin involves inflation through freeform chamber boundaries with overlaps, resulting in more complex and spatially coupled surface deformations. The two soft manipulators highlight structural bending, with one driven purely by bending and the other combining inflation / compression and bending, thereby introducing hybrid deformation behaviors. Together, these systems cover the spectrum from simple to complex chamber inflation and from pure bending to coupled bending, inflation and compression, demonstrating that our sim-to-real learning framework generalizes across diverse actuation principles and deformation characteristics rather than being tailored to a single morphology.

\begin{table*}[t] \footnotesize

\centering
\caption{\JENew{Training Details for Different Hardware Setups}}\label{tab:DetailsofTraining}

\scalebox{0.93}{
\begin{tabular}{c||l|c|c|c|c|c}         
\hline \hline                          
\specialrule{0em}{1pt}{1pt}
\multicolumn{1}{c|}{\multirow{3}{*}{Network}} &
\multicolumn{1}{l|}{\multirow{3}{*}{Detail Parameters}} & \multicolumn{1}{c|}{Deformable Membrane} & \multicolumn{1}{c|}{Soft Manipulator}& 
\multicolumn{2}{c|}{Robotic Mannequin}& 
\multicolumn{1}{c}{Compressible Manipulator}
\\ 
\specialrule{0em}{1pt}{1pt}
\cline{3-7} 
\specialrule{0em}{1pt}{1pt}
\multicolumn{1}{c|}{} &
\multicolumn{1}{c|}{} & \multicolumn{1}{c|}{(3D Scanner)} & \multicolumn{1}{c|}{(3D Scanner)}& \multicolumn{1}{c|}{(3D Scanner)}& \multicolumn{1}{c|}{(MoCap)}
& \multicolumn{1}{c}{(MoCap)}\\ 

\specialrule{0em}{1pt}{2pt}
\hline \hline
\specialrule{0em}{1pt}{2pt}
 \multicolumn{1}{c|}{
 \multirow{6}{*}
 {
 $\mathcal{N}_{fk}$
 }} & 
 \multicolumn{1}{l|}{Actuation DoFs} &  \multicolumn{1}{c|}{9} & \multicolumn{1}{c|}{6}& 
\multicolumn{2}{c|}{9}& 
\multicolumn{1}{c}{9}
\\

\multicolumn{1}{c|}{}&
\multicolumn{1}{l|}{\# of Simulated Shapes for Training} & \multicolumn{1}{c|}{1,000} & \multicolumn{1}{c|}{1,000}& 
\multicolumn{2}{c|}{1,000}& 
\multicolumn{1}{c}{4,000}

\\

\multicolumn{1}{c|}{}&
 \multicolumn{1}{l|}{Avg. Time (sec.) for Simulating Each Pose} & \multicolumn{1}{c|}{110.0} & \multicolumn{1}{c|}{900.0}& 
\multicolumn{2}{c|}{150.0}&
\multicolumn{1}{c}{5.0}
\\

\multicolumn{1}{c|}{}&
 \multicolumn{1}{l|}{\# of Hidden Layers} & \multicolumn{1}{c|}{2} & \multicolumn{1}{c|}{2}&  
\multicolumn{2}{c|}{2}&
\multicolumn{1}{c}{3}
\\

\multicolumn{1}{c|}{}&
 \multicolumn{1}{l|}{\# of Neurons per Hidden Layer} & \multicolumn{1}{c|}{128} & \multicolumn{1}{c|}{128}& 
\multicolumn{2}{c|}{128}
& 
\multicolumn{1}{c}{128}
\\

\multicolumn{1}{c|}{}&
 \multicolumn{1}{l|}{\# of Control Points} & \multicolumn{1}{c|}{30 $\times$ 30} & \multicolumn{1}{c|}{30 $\times$ 30}& 
\multicolumn{2}{c|}{30 $\times$ 30}
& 
\multicolumn{1}{c}{30 $\times$ 30}
\\

\specialrule{0em}{1pt}{2pt}
\hline \hline

\specialrule{0em}{1pt}{2pt}

\multicolumn{1}{c|}{}&
 \multicolumn{1}{l|}{\# of Kernels$^\dagger$} & \multicolumn{1}{c|}{100} & \multicolumn{1}{c|}{100}& 
\multicolumn{1}{c|}{100}& 
\multicolumn{1}{c|}{34}& 
\multicolumn{1}{c}{100}
\\

\multicolumn{1}{c|}{}&
 \multicolumn{1}{l|}{\# of Hidden Layer} & \multicolumn{1}{c|}{2} & \multicolumn{1}{c|}{2}& 
\multicolumn{1}{c|}{2}& 
\multicolumn{1}{c|}{2}& 
\multicolumn{1}{c}{3}
\\

\multicolumn{1}{c|}{}&
 \multicolumn{1}{l|}{\# of Neurons per Hidden Layer} & \multicolumn{1}{c|}{24} & \multicolumn{1}{c|}{24}& 
\multicolumn{1}{c|}{24}& 
\multicolumn{1}{c|}{24}& 
\multicolumn{1}{c}{64}
\\
 \cline{2-7} 
 \multicolumn{1}{c|}{
 \multirow{6}{*}
 {
 $\mathcal{N}_{s2r}$
 }} & 
 \multicolumn{1}{l|}{\# of Frames Captured for Training} & \multicolumn{1}{c|}{40} & \multicolumn{1}{c|}{30}& 
\multicolumn{1}{c|}{40}& 
\multicolumn{1}{c|}{40}& 
\multicolumn{1}{c}{500}
\\
\cline{2-7} 

\multicolumn{1}{c|}{}&
\multicolumn{1}{l|}{\# of Sample Points Used for Each Frame} & \multicolumn{1}{c|}{200} & \multicolumn{1}{c|}{400}& 
\multicolumn{1}{c|}{1000}& 
\multicolumn{1}{c|}{}& 
\multicolumn{1}{c}{}
\\

\multicolumn{1}{c|}{}&
 \multicolumn{1}{l|}{Total \# of Valid Sample Points} & \multicolumn{1}{c|}{7,661} & \multicolumn{1}{c|}{9,891}& 
\multicolumn{1}{c|}{31,588}& 
\multicolumn{1}{c|}{}& 
\multicolumn{1}{c}{}
\\

\multicolumn{1}{c|}{}&
 \multicolumn{1}{l|}{\% of Missing Sample Points} & \multicolumn{1}{c|}{4.2\%} & \multicolumn{1}{c|}{17.6\%}& 
\multicolumn{1}{c|}{21.0\%}& 
\multicolumn{1}{c|}{}& 
\multicolumn{1}{c}{}
\\

\multicolumn{1}{c|}{}&
 \multicolumn{1}{l|}{\# of Frames with Missing Sample Points} & \multicolumn{1}{c|}{5} & \multicolumn{1}{c|}{16}& 
\multicolumn{1}{c|}{25}& 
\multicolumn{1}{c|}{}& 
\multicolumn{1}{c}{}
\\
\cline{2-7} 
\multicolumn{1}{c|}{}&
\multicolumn{1}{l|}{\# of Makers Used for Each Frame} & \multicolumn{1}{c|}{} & \multicolumn{1}{c|}{}& 
\multicolumn{1}{c|}{}& 
\multicolumn{1}{c|}{34}& 
\multicolumn{1}{c}{400}
\\

\multicolumn{1}{c|}{}&
 \multicolumn{1}{l|}{Total \# of Valid Markers} & \multicolumn{1}{c|}{} & \multicolumn{1}{c|}{}& 
\multicolumn{1}{c|}{}& 
\multicolumn{1}{c|}{1,330}& 
\multicolumn{1}{c}{180,362}
\\

\multicolumn{1}{c|}{}&
 \multicolumn{1}{l|}{\% of Missing Markers} & \multicolumn{1}{c|}{} & \multicolumn{1}{c|}{}& 
\multicolumn{1}{c|}{}& 
\multicolumn{1}{c|}{2.2\%}& 
\multicolumn{1}{c}{9.8\%}
\\

\multicolumn{1}{c|}{}&
 \multicolumn{1}{l|}{\# of Frames with Missing Markers} & \multicolumn{1}{c|}{} & \multicolumn{1}{c|}{}& 
\multicolumn{1}{c|}{}& 
\multicolumn{1}{c|}{18}& 
\multicolumn{1}{c}{161}
\\

\specialrule{0em}{1pt}{2pt}
\hline\hline

\specialrule{0em}{1pt}{2pt}

\multicolumn{1}{c|}{\multirow{2}{*}
 {
 $\mathcal{N}_{conf}$
 }}&
 \multicolumn{1}{l|}{\# of Hidden Layer} & \multicolumn{1}{c|}{3} & \multicolumn{1}{c|}{3}& 
\multicolumn{1}{c|}{3}& 
\multicolumn{1}{c|}{3}& 
\multicolumn{1}{c}{3}
\\

\multicolumn{1}{c|}{}&
 \multicolumn{1}{l|}{\# of Neurons per Hidden Layer} & \multicolumn{1}{c|}{128} & \multicolumn{1}{c|}{128}& 
\multicolumn{1}{c|}{128}& 
\multicolumn{1}{c|}{128}& 
\multicolumn{1}{c}{128}
\\
\specialrule{0em}{1pt}{2pt}

\hline\hline

\end{tabular}
}
\begin{flushleft}\footnotesize
$^\dagger$~\finalrev{When Mocap is employed for the soft mannequin example, the kernels are located to be consistent with 34 real markers; for other examples,} kernels are sampled in a decoupled way in the $u,v$-domain.
\end{flushleft}

\label{tab:LearningData}
\vspace{-10pt}
\end{table*}

\subsection{Details of Networks and Training}\label{subsecImplementationDetails}
\subsubsection{Forward kinematics network $\mathcal{N}_{fk}$}
For training $\mathcal{N}_{fk}$, a dataset with \JENew{$m \geq 1000$} different pairs of actuation and shapes is collected from the simulation. This dataset contains shapes with the minimal and the maximal pressures applied to every chamber -- in total $2^k$ shapes were generated with $k$ being the DoFs of actuation (see Table \ref{tab:LearningData} shown parameters for different robots). We also randomly sampled other \JENew{$(m-2^k)$} actuation parameters using the Halton sequence \cite{halton1960efficiency}. All these $m$ samples of actuation are applied in the simulation system to obtain their simulated shapes to explore the whole deformation space. Considering the large non-linearity between actuation and shape, we non-uniformly resample the actuation space by the amount of chamber inflation to enhance the training accuracy. The average time used for completing each simulation has been reported in Table \ref{tab:LearningData}. This dataset based on simulation is separated in the ratio of $7:3$ for training and testing.

\begin{figure}[t] 
\centering
\includegraphics[width=0.95\linewidth]{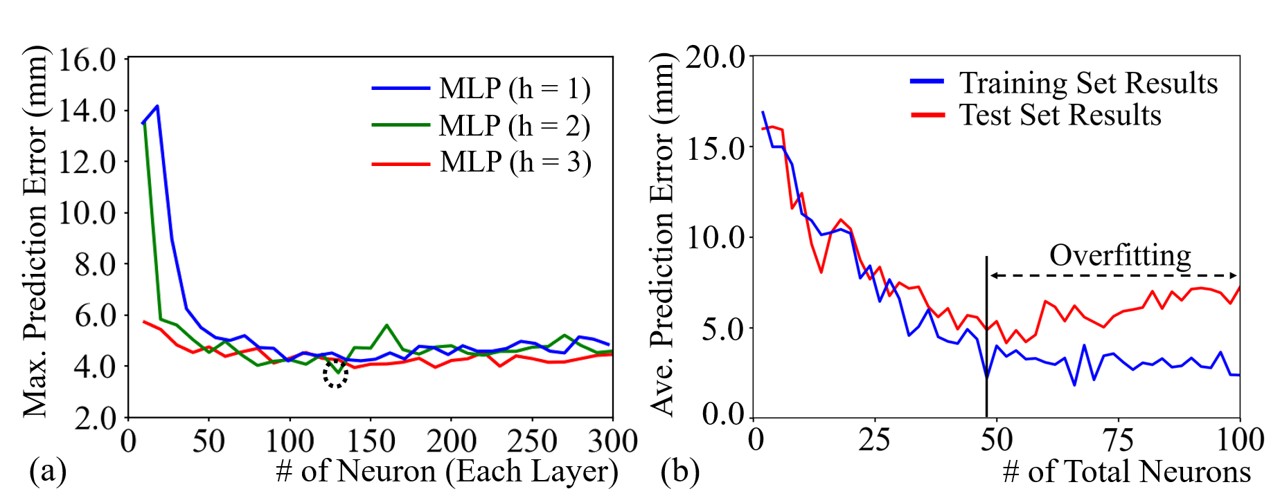}
\caption{\rev{}{Study of network parameters \JENew{taken on the soft deformable mannequin} -- (a) the maximal shape prediction error w.r.t. different numbers of layers ($h$) and numbers of neurons per layer ($l$) are evaluated to choose the `best' values of $h$ and $l$ for $\mathcal{N}_{fk}$ (circled by black dash lines), and (b) the average shape prediction error w.r.t. the number of neurons in $\mathcal{N}_{rbf}$ is studied to avoid overfitting while training $\mathcal{N}_{s2r}$ using a limited number of samples. Note that $\mathcal{N}_{s2r}$ contains $\mathcal{N}_{rbf}$ plus a space warping module as illustrated in Fig.\ref{fig:sim2RealPipelineOverview}.}}\label{fig:NfkTraining}
\end{figure}

\begin{figure*}[!t] 
\centering
\includegraphics[width=1.0\linewidth]{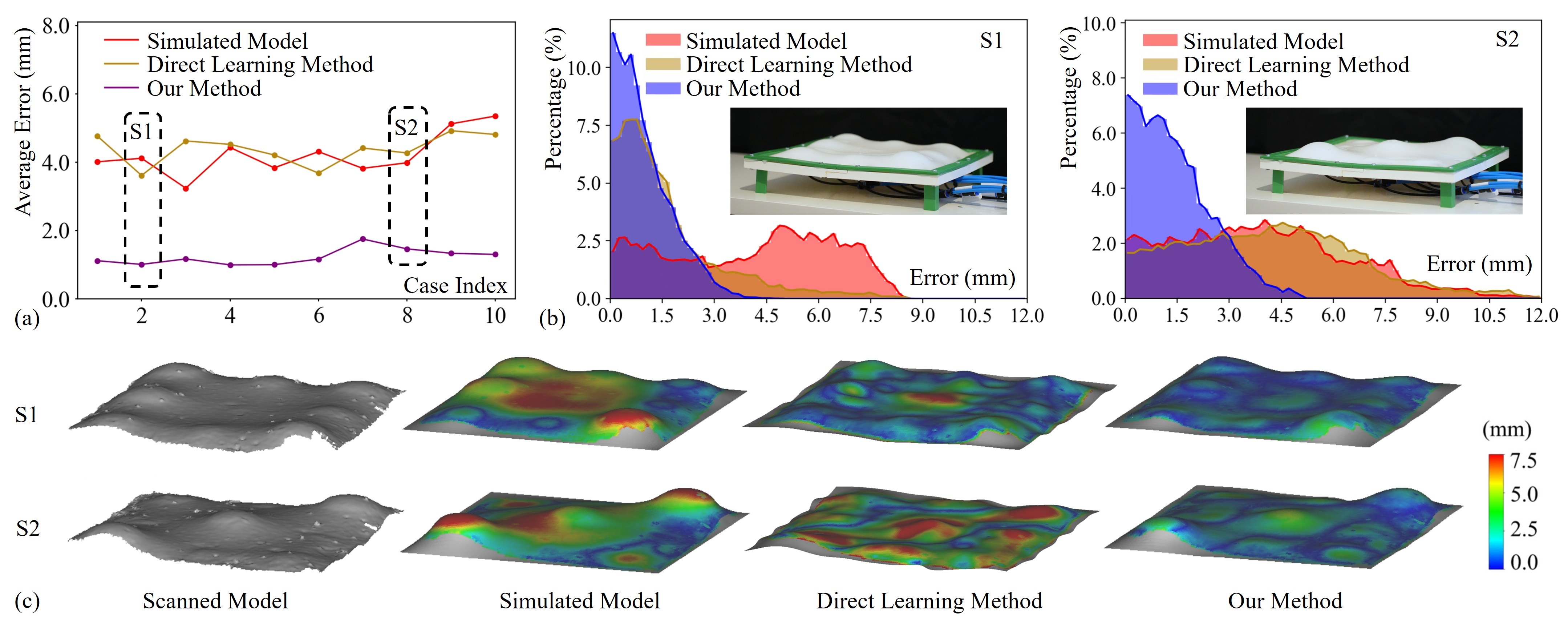}
\caption{
\JENew{Shapes predicted by using 1) simulation only, 2) direct learning and 3) our sim-to-real method are compared on the deformable membrane: (a) the average shape errors measured on 10 different randomly selected configurations of actuation, (b) the histograms of shape error distributions on the configuration S1 and S2, and (c) the predicted shapes of the deformed membrane and their geometry errors displayed in colormaps.
} 
}\label{fig:sim2realOnMembrane}
\end{figure*}

We carefully select the network structure, which includes the number of hidden layers (\textit{h}), and the number of neurons in each hidden layer (\textit{l}). The experimental results of the maximal shape prediction error according to different network structures for $\mathcal{N}_{fk}$ are studied and illustrated in Fig.~\ref{fig:NfkTraining}(a), and we choose two hidden layers with each layer contains 128 neurons as the final network structure to balance the quality of training result and the computational cost. In our implementation, ReLU is selected as the activation function and batch normalization is applied to improve stability in training. Note that our experiment finds that prediction with better quality can be achieved when learning the translation vectors applied to the positions of control points on an average model of all shapes.

\subsubsection{Function prediction network $\mathcal{N}_{rbf}$} 
For learning the sim-to-real transfer, we employ a network architecture with 2 or 3 hidden layers as $\mathcal{N}_{rbf}$, where each layer has a certain number of neurons using the ReLU activation function. The input layer contains the positions of $30 \times 30$ control points, and the output layer is the coefficients for the RBF-based warping function as $\bm{\gamma} \in \mathbb{R}^{3(N+4)}$ with $N$ being the number of \JENew{kernels}. The training of $\mathcal{N}_{rbf}$ is taken on \JENew{shapes captured by} either 3D scanner or MoCap while randomly varying the pneumatic actuation within the working range of the chambers. Among these captured shapes, a certain number of frames have missing \JENew{regions (or markers)}. However, all shapes are employed to train $\mathcal{N}_{rbf}$. The total number of valid \JENew{samples (or markers)} for different cases have been reported in Table \ref{tab:LearningData}. Using the dataset of deformable mannequin generated from MoCap \JENew{as an eaxmple}, it can be observed from the study as shown in Fig.~\ref{fig:NfkTraining}(b) that the prediction error starts to increase when the total number of neurons exceeds 50 -- i.e., overfitting happens. Based on this analysis, we select 24 neurons for each layer in $\mathcal{N}_{rbf}$.

\subsubsection{Confidence map network $\mathcal{N}_{conf}$} 
\JENew{
The input of the confidence map network $\mathcal{N}_{conf}$ is a vector comprising the simulated position $\mathbf{p} \in \mathbb{R}^3$ of a point on the surface, along with the control points $\mathcal{S}^c$ that define the global shape features. This combined input is processed through three fully connected hidden layers, each with 128 neurons and ReLU activation, followed by a fully connected layer with a single neuron and sigmoid activation. The output is a score in the range of $[0,1]$.}

In our implementation, we first train $\mathcal{N}_{fk}$, and then \JENew{jointly train $\mathcal{N}_{rbf}$ and $\mathcal{N}_{conf}$ while keeping $\mathcal{N}_{fk}$ unchanged}. The RBF-based space warping function $\mathbf{\Phi}$ and the B-spline shape decoder $\mathbf{B}$ are both implemented as differentiable layers, enabling seamless integration into the training process of $\mathcal{N}_{rbf}$ \finalrev{by backpropagation}. 

\begin{figure}[!t] 
\centering
\includegraphics[width=1.0\linewidth]{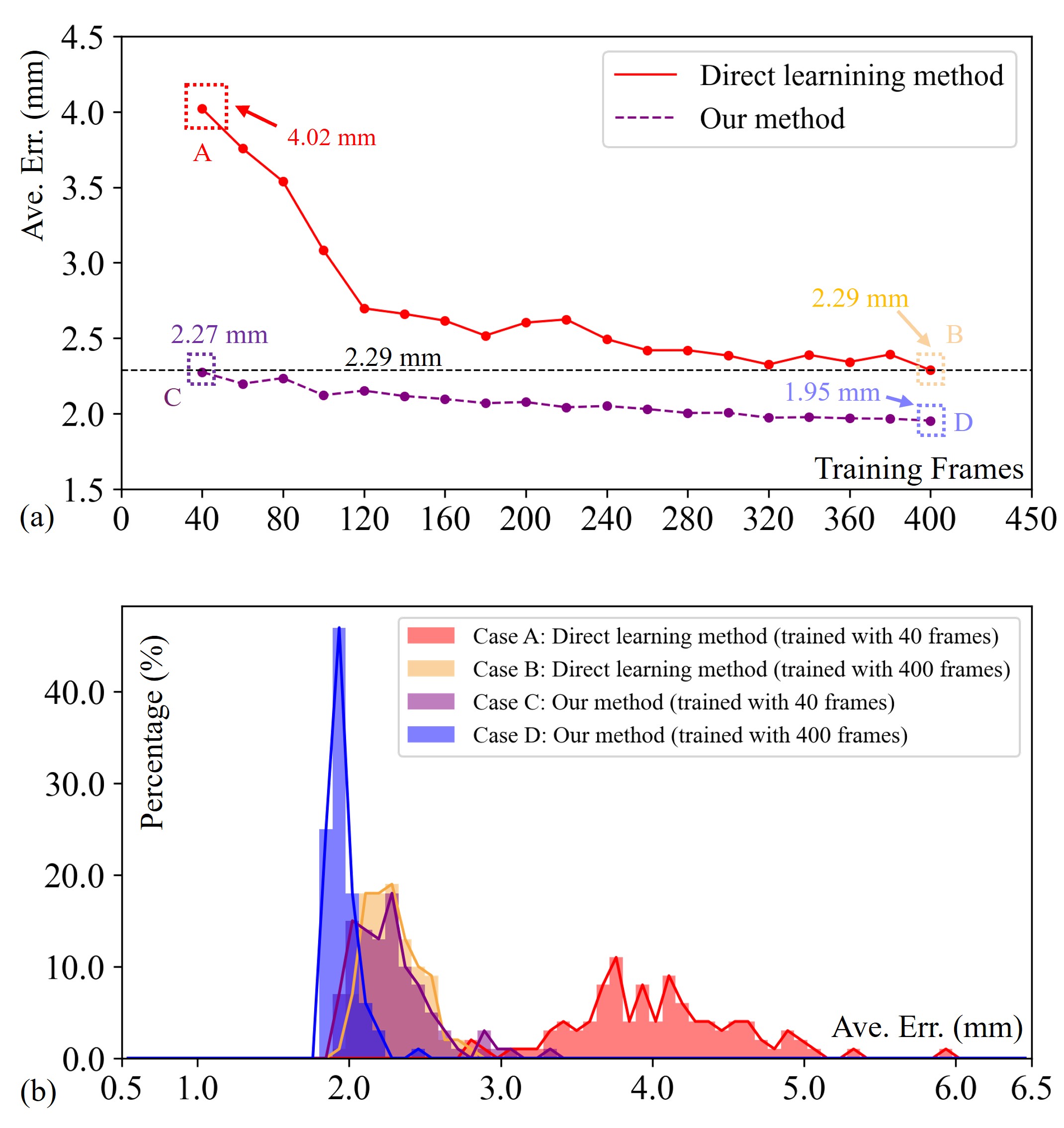}
\caption{\JENew{The comparison of the direct learning result and our sim-to-real transfer learning by using different frames of scanned 3D shapes: (a) the average shape approximation error and (b) the histogram of error distributions when 40 and 400 frames of shapes are employed.}
}\label{fig:sim2realOnMembraneRevision}
\end{figure}

\begin{figure*}[t] 
\centering
\includegraphics[width=1.0\linewidth]{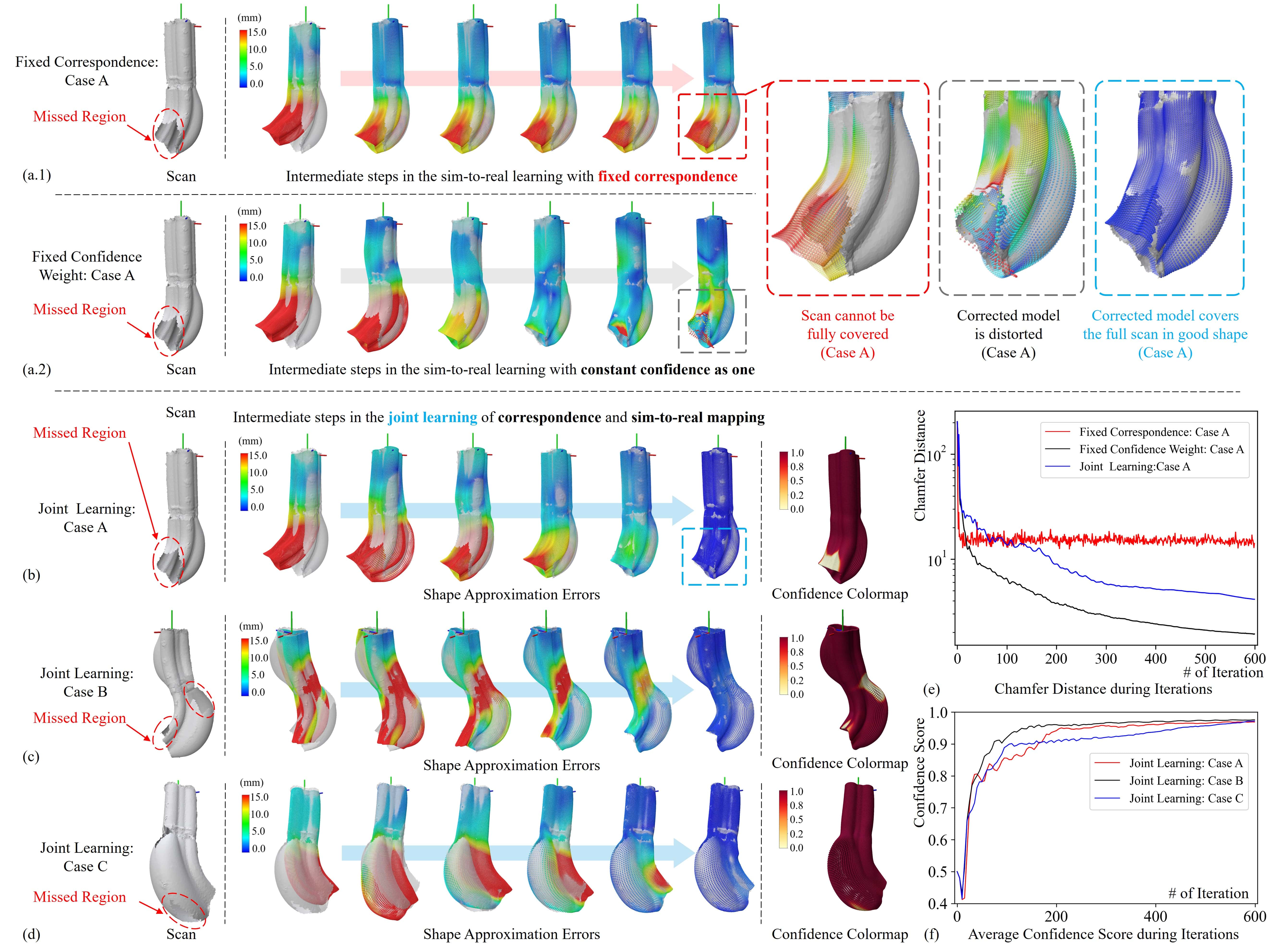}
\caption{\JENew{Compared to (a.1) the sim-to-real learning with pre-determined correspondence and (a.2) the joint sim-to-real learning with fixed confidence map (i.e., keeping $w_c(\mathbf{p}) = 1.0$), our joint learning pipeline can generate the predicted shape with much smaller geometric approximation errors with the help of the differentiable alignment module -- see (b)-(d) for three different cases. All cases have large missed regions on the scanned point clouds.} 
}\label{fig:coLearnDefCorrespondence}
\end{figure*}

\subsection{Results of Sim-to-Real Learning by 3D Scanner}
\JENew{We now study the performance of our sim-to-real learning approach by a structure-light-based 3D scanner, obtaining 3D shapes as point clouds without correspondences. The tests have been conducted on the deformable membrane (Fig.\ref{fig:HardwareSystem}(a)) and the soft manipulator with two segments (Fig.\ref{fig:HardwareSystem}(b)).}

\subsubsection{Sim-to-real vs. direct learning}
\JENew{
In the first study, we apply 10 random unseen actuation parameters to the deformable membrane, where the  corresponding shapes are predicted by our $\mathcal{N}_{fk}$ and $\mathcal{N}_{s2r}$ networks and compared with the scanned shapes. The results of the comparison are shown in Fig.\ref{fig:sim2realOnMembrane}, where the geometric errors are visualized on surfaces by colormaps. The statistical error distributions are given as histograms. For the selected two configurations -- S1 and S2, the average errors are reduced by 75.5\% and 63.3\% when using our sim-to-real approach, and the maximal errors are reduced by 41.8\% and 58.6\% accordingly.
}

\begin{figure}[t] 
\centering
\vspace{-10pt}
\includegraphics[width=\linewidth]{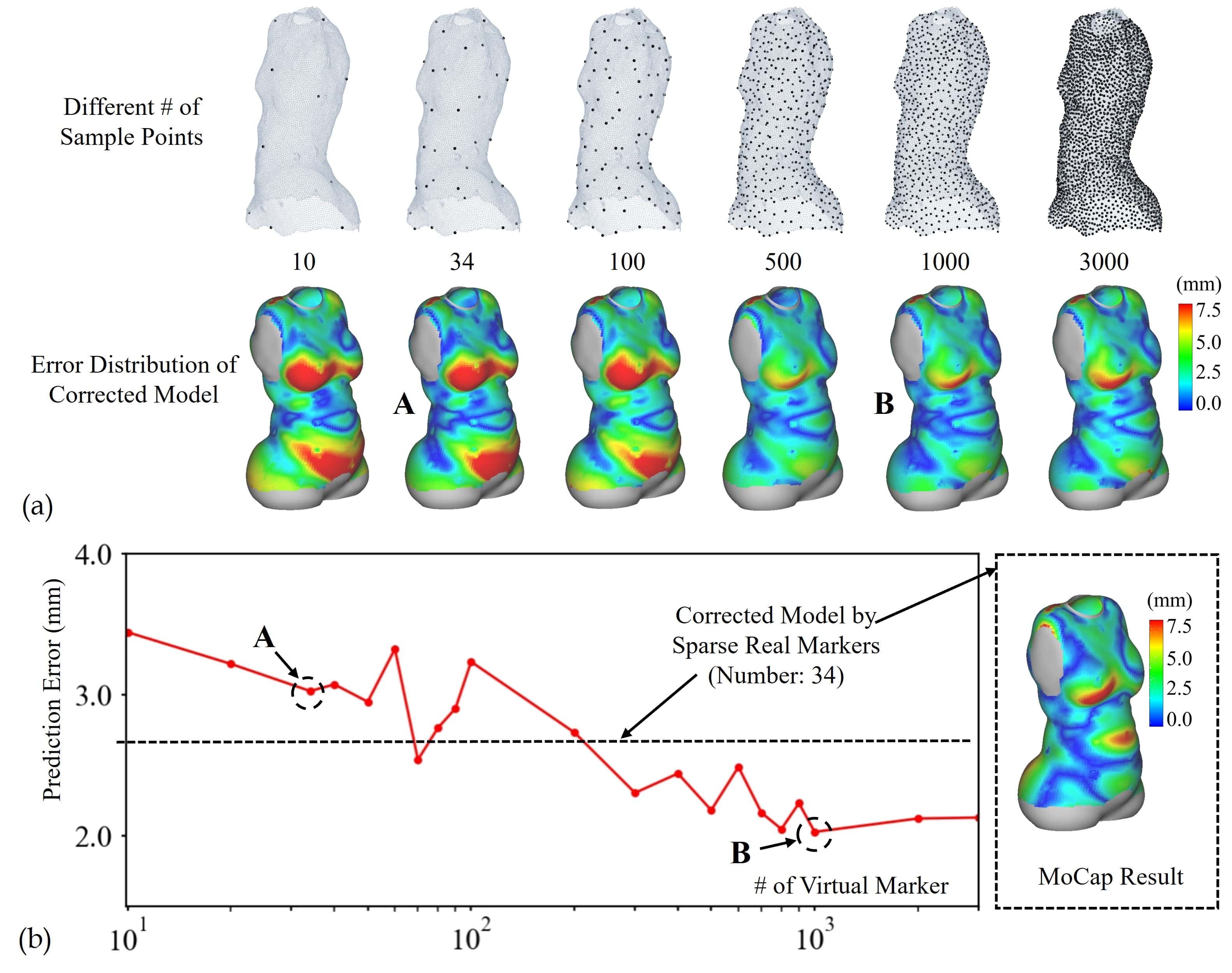}
\caption{\JENew{Experimental tests to study the accuracy of shape prediction (a) when using different number of sample points in $\mathcal{S}^p$ for the sim-to-real learning. (b) The results in terms of average geometric errors are compared with the sim-to-real result by using 34 real markers with correspondences.}
}\label{fig:SamplePntDensityStudy}
\end{figure}
\begin{figure}[t] 
\vspace{-5pt}
\centering
\includegraphics[width=1.0\linewidth]{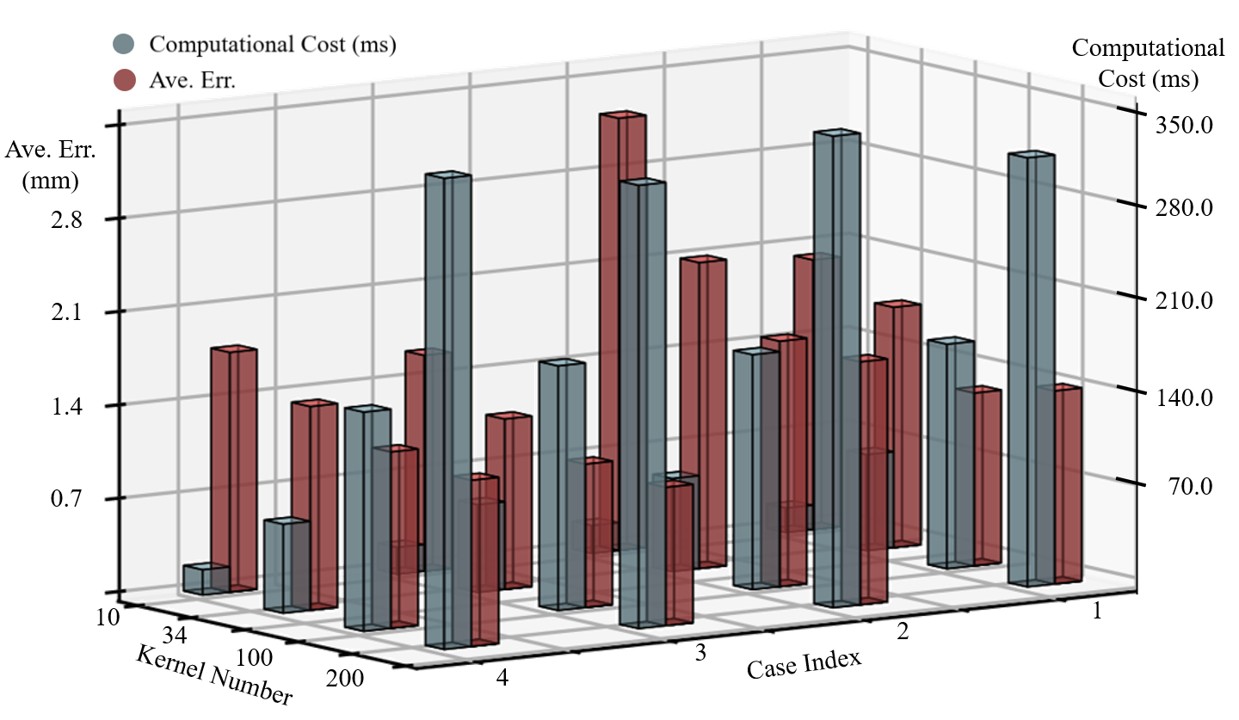}
\caption{\JENew{The experiment demonstrates how the number of RBF kernels affects the accuracy of prediction (measured as the average of shape approximation error) and the computational cost (measured as the required time to evaluate $10^5$ warping operations. Four different shapes (as A1-A4 shown in Fig.\ref{fig:Sim2RealEffectiveStudy}) are evaluated in these tests.}
}\label{fig:DiffKernelNum}
\end{figure}

An experiment is taken now to study if we can directly learn to predict the deformed shape from the actuation parameters using neural networks. We applied 40 random sets of  actuation parameters to deform the membrane and then scanned the corresponding physically deformed shapes as the training dataset. The direct learning network consists of two hidden MLP layers with 128 neurons in each layer. The input of the network is actuation parameters (i.e., 9 values for each shape) and the output is 200 points sampled on the deformed surface. 
\finalrev{To reduce the geometry discrepancy between the predicted surfaces $\mathcal{S}_j^*$ and the scanned surfaces $\mathcal{S}_j^p$, the Hausdorff distance \cite{HausdorffLossTorch} as
\begin{align}\label{eq:HausdorffDist}
    D_{\text{Haus}}(\mathcal{S}_j^*, \mathcal{S}_j^p) 
&= \max \Big(
\max_{\mathbf{p}^* \in \mathcal{S}_j^*} 
\min_{\mathbf{x} \in \mathcal{S}_j^p} \| \mathbf{p}^* - \mathbf{x} \|_2, \nonumber\\
&\quad\;\;
\max_{\mathbf{x} \in \mathcal{S}_j^p} 
\min_{\mathbf{p}^* \in \mathcal{S}_j^*} \| \mathbf{p}^* - \mathbf{x} \|_2
\Big), 
\end{align}
and the chamfer distance \cite{ChamferLossTorch} as
\begin{align}\label{eq:ChamferDist}
D_{\text{Cham}}(\mathcal{S}_j^*, \mathcal{S}_j^p) 
&= \frac{1}{|\mathcal{S}_j^*|} 
\sum_{\mathbf{p}^* \in \mathcal{S}_j^*} 
\min_{\mathbf{x} \in \mathcal{S}_j^p} \| \mathbf{p}^* - \mathbf{x} \|_2^2 \nonumber\\
&\quad + \frac{1}{|\mathcal{S}_j^p|} 
\sum_{\mathbf{x} \in \mathcal{S}_j^p} 
\min_{\mathbf{p}^* \in \mathcal{S}_j^*} \| \mathbf{p}^* - \mathbf{x} \|_2^2
\end{align}
are jointly used to penalize the maximum and the average deviations between two point clouds by a loss as}
\begin{equation}\label{eq:totalLossDirectLearn}
    \finalrev{\mathcal{L}_{direct} = D_{\text{Haus}}  + 5.0 D_{\text{Cham}}.}
\end{equation}
The resultant surfaces of prediction are then generated by RBF-based surface deformation using these 200 predicted sample points. The result comparison between this direct learning method and our method has been given in Fig.\ref{fig:sim2realOnMembrane}. It can be observed that direct learning leads to results with significantly larger errors.

\JENew{
We further conducted this comparison by using more sample shapes for both the direct learning and our sim-to-real based learning -- 500 shapes are captured by using 80\% for training and 20\% for testing. As shown in Fig.\ref{fig:sim2realOnMembraneRevision}, our method also delivers substantially better performance. When using 40 frames in our sim-to-real based training, the error can already be reduced to a very low level similar to the direct training using 400 frames.
}

\subsubsection{Effectiveness of joint learning}
\JENew{
We now conduct a study to demonstrate the effectiveness of jointly learning the sim-to-real transfer network and the confidence map network with the help of the differentiable alignment module. The experiments are performed on the soft manipulator with two segments. Due to large deformations, the scanned 3D shapes often contain missing regions (see Fig.\ref{fig:coLearnDefCorrespondence}). For comparison, we also generate sim-to-real training results using unchanged correspondences that are pre-determined by a state-of-the-art non-rigid registration method \cite{yao2020quasi}. This actually treats the sample points as `virtual' markers by using $L^2$-norm distances. As shown in Fig.\ref{fig:coLearnDefCorrespondence}(a.1), using static correspondences can lead the training to become stuck in poor local minima. Furthermore, we also conduct a test by using a fixed confidence weight $w_c(\mathbf{p})=1.0$ but using Chamfer distance. The result is as shown in Fig.\ref{fig:coLearnDefCorrespondence}(a.2), where the surface of transferred model shows unwanted distortion caused by the points falling in the missed regions. In contrast, our joint learning method consistently produces results with significantly smaller geometric errors. Examples of our method applied to different shapes are shown in Fig.~\ref{fig:coLearnDefCorrespondence}(b)–(d).
}

\subsubsection{Number of samples and kernels}\label{subsub:NumSamplesKernels}
\JENew{
When performing sim-to-real learning with dense point cloud input, we usually downsample the scanned 3D point clouds to a fixed number of points to define $\mathcal{S}^p$. To determine an appropriate sampling density, we created datasets with varying numbers of points and trained the sim-to-real network $\mathcal{N}_{s2r}$ on each dataset. After training, these networks were evaluated on unseen cases to assess shape prediction accuracy. As shown in Fig.\ref{fig:SamplePntDensityStudy}(a), the approximation error decreases as the number of sample points increases, and begins to plateau around 1,000 points. Furthermore, when comparing networks trained on point clouds without correspondence, those incorporating MoCap-provided correspondences consistently achieve higher accuracy under comparable sampling densities (see Fig.\ref{fig:SamplePntDensityStudy}(b)). This is likely because motion capture datasets provide precise correspondences, whereas 3D scan datasets are subject to alignment noise and inaccuracies.
}

\JENew{The number of Gaussian kernels, denoted as $N$ in Eq.\eqref{eqRBFSpaceWarping}, is also a factor influencing the accuracy and efficiency of the sim-to-real method. With too few kernels, the space warping function lacks the capacity to capture complex variations, leading to notable errors. Conversely, an excessive number of kernels increases computational demands, heavily impacting the IK computation as it must repeatedly apply the RBF-based space warping. To determine the number of kernels, we maintain 1000 sample points and train different sim-to-real networks on the same dataset with 40 shapes, varying only the number of kernels. Shape approximation errors were then evaluated on four unseen cases, with results shown in Fig.\ref{fig:DiffKernelNum}. As a result, more kernels generally lead to higher prediction accuracy but come with the cost of longer computation times. To balance between the accuracy and the efficiency, we usually select 100 kernels for our implementation and tests.}

\begin{figure*}[t] 
\centering
\includegraphics[width=1.0\linewidth]{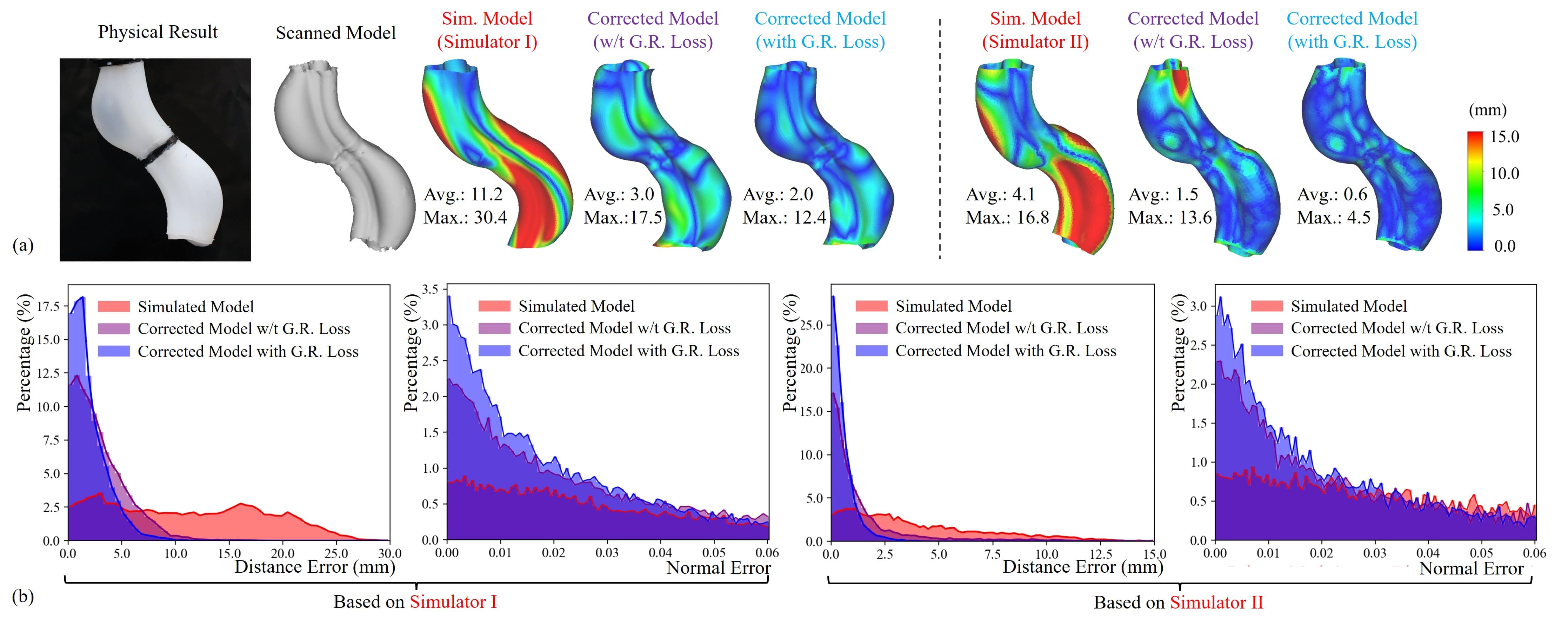}
\caption{ 
\JENew{An ablation study was conducted to test the effectiveness of the geometric regularization loss in sim-to-real learning based on two different simulators -- (I) the one based on FEA~\cite{xavier2021finite} and (II) the geometry-based simulator (ref.~\cite{Guoxin2020TRO,Guoxin2022IROS}), where both simulators are struggling to predict the deformed surfaces accurately as shown in (a). It can be observed from the colormaps as the distance errors in (a) and also the histograms of distance errors and normal errors in (b) that the corrected model incorporating the geometric regularization loss (labeled as `with G.R. Loss') produces a significantly improved shape comparing to the corrected model without compatibility loss (labeled as `w/t G.R. Loss'). The generality of our approach has been proved by fixing the sim-to-real gap on the results generated by two different simulators.}
}\label{fig:CompatibleCompare}
\end{figure*}

\subsubsection{Ablation study -- geometric regularization}
\JENew{
An ablation study was conducted on the soft manipulator with two segments to illustrate the importance of geometric regularization loss in achieving accurate shapes and reducing the gap between simulation and reality. This setup is selected due to its complex deformation, where both the global structure shape and the local surface shape are significantly changed. To further validate the effectiveness and the generality of our sim-to-real pipeline with compatibility loss, we have tested it by using two different simulators -- one is based on FEA~\cite{xavier2021finite} and the other is based on the geometry-based simulator (ref.~\cite{Guoxin2020TRO,Guoxin2022IROS}). The results can be found in Fig.\ref{fig:CompatibleCompare}. The normal error at any point $\mathbf{p}_j$ on a simulated (or corrected) surface is defined as:
\begin{equation}\label{eqNormalVarHistogram}
    \eta(\mathbf{p}_j) = 1-\mathbf{n}(\mathbf{p}_j) \cdot \mathbf{n}(\mathbf{c}^{p}(\mathbf{p}_j))
\end{equation}
with $\mathbf{c}^{p}(\cdot)$ giving the closest point of $\mathbf{p}_j$ on the scanned point cloud $\mathcal{S}^p$. The distribution of the distance errors and the normal difference errors are plotted as the histograms shown in Fig.\ref{fig:CompatibleCompare}(b). Note that the histograms of normal errors are capped at 0.06 -- i.e., an angular difference of less than $20^\circ$ between the two normals. This allows to focus on a range of interest for better comparison.}

\JENew{The experimental tests shown in Fig.\ref{fig:CompatibleCompare} demonstrate that the geometric regularization loss, derived from the compatibility condition of the RBF warping function (i.e., Eq.\eqref{eqCompatibilityCondition}), is essential for achieving better shape accuracy on the corrected models. These tests also confirm the generality of our network when working on different types of numerical simulators.}

\begin{figure}[!t] 
\centering
\hspace{-10pt}
\includegraphics[width=1.0\linewidth]{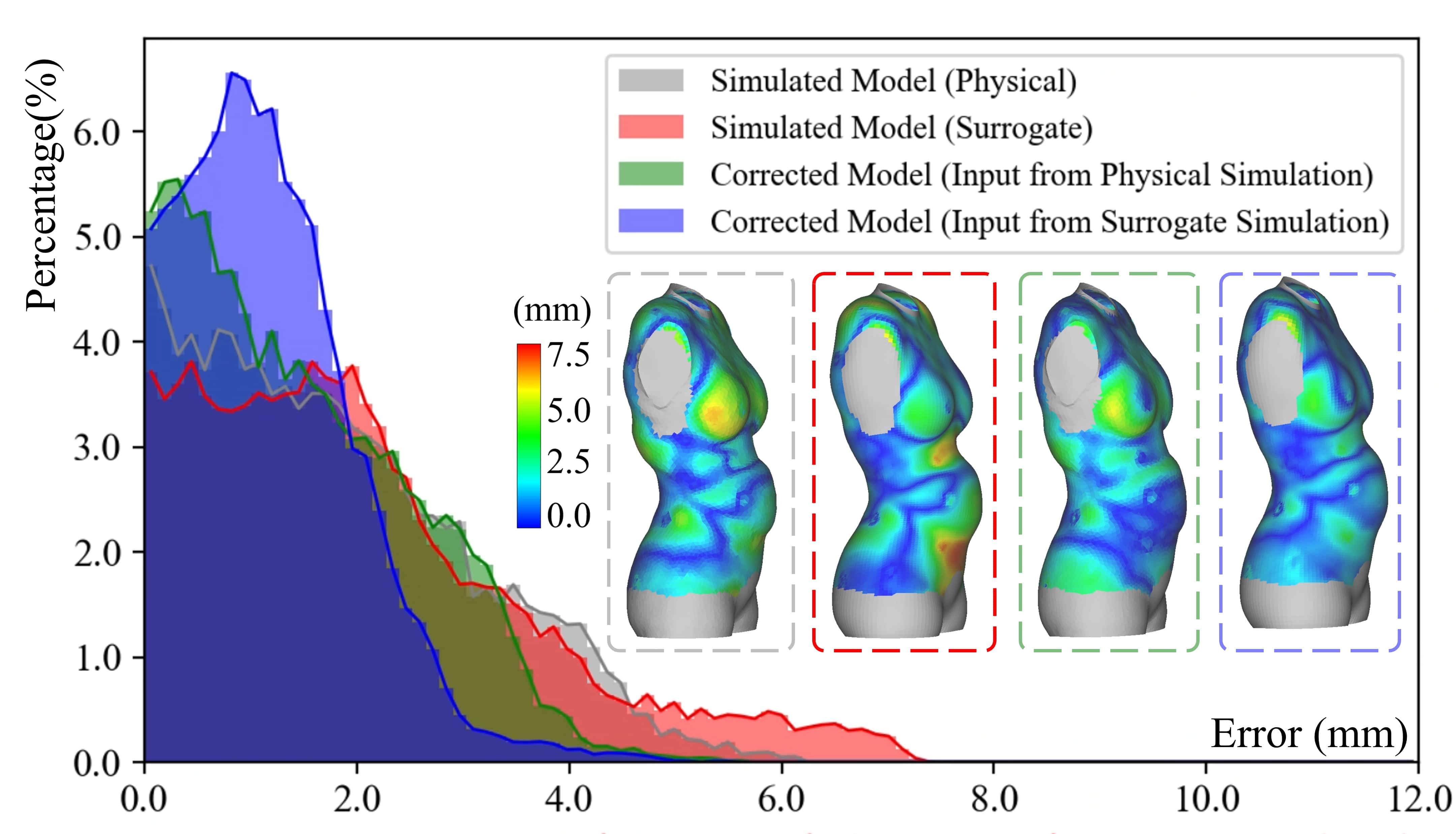}
\caption{\JENew{We apply the sim-to-real network $\mathcal{N}_{s2r}$ trained from the surrogate simulation on an unseen actuation given on the simulated models from both the direct physical simulation and the surrogate based simulation. The results are compared with the 3D scanned real shape, and the distance errors are visualized by colormaps and histograms.}
}\label{fig:simToRealonPhySim}
\end{figure}

\subsubsection{Ablation study -- surrogate vs. direct simulation}
\JENew{
To further evaluate the generalizability of our sim-to-real approach, we tested the performance of the sim-to-real network $\mathcal{N}_{s2r}$ -- trained using a surrogate simulator -- on input shapes generated by direct physical simulation. The comparison results are shown in Fig.~\ref{fig:simToRealonPhySim}. We observe that the sim-to-real network can still effectively reduce the shape approximation errors. However, the errors are slightly higher than those using input shapes predicted by the surrogate simulator $\mathcal{N}_{fk}$. This is likely because that the sim-to-real network is trained on the data from  the surrogate simulation, which introduces a small domain gap when being applied to the physically simulated data.}

\begin{figure*}[t] 
\centering
\includegraphics[width=1.0\linewidth]{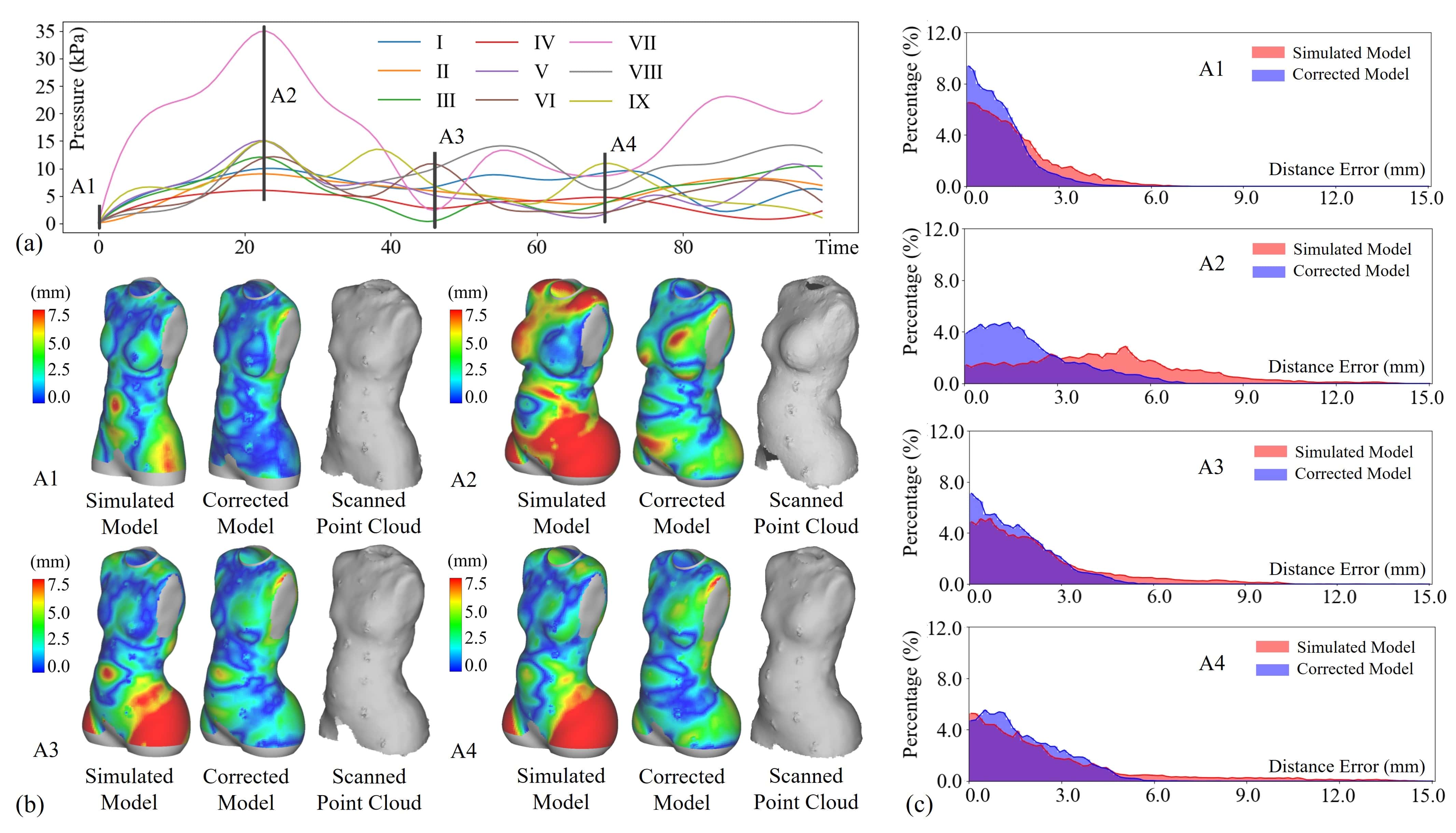}
\caption{Result of different free-form surfaces predicted by our forward kinematics pipeline (i.e., Fig.\ref{fig:NNBasedFK} and Eq.(\ref{eqSim2RealPrediction})) while changing the actuation parameters (a). Four instants (A1-A4) are selected to apply onto the physical setup with the resultant shapes scanned and compared with the predicted shapes, where the shape approximation errors are visualized as colors.
The results with and without sim-to-real transfer (denoted by corrected and simulated models respectively) are compared in both the color maps (b) and the error histograms (c).
\JENew{The average errors are reduced by $30.4\%$, $53.5\%$, $32.1\%$ and $33.5\%$, and the maximal errors are reduced by $6.7\%$, $44.0\%$, $50.6\%$ and $60.6\%$ on these different shapes. }
}\label{fig:Sim2RealEffectiveStudy}
\end{figure*}

\begin{figure}[!t] 
\vspace{-5pt}
\centering
\includegraphics[width=1.0\linewidth]{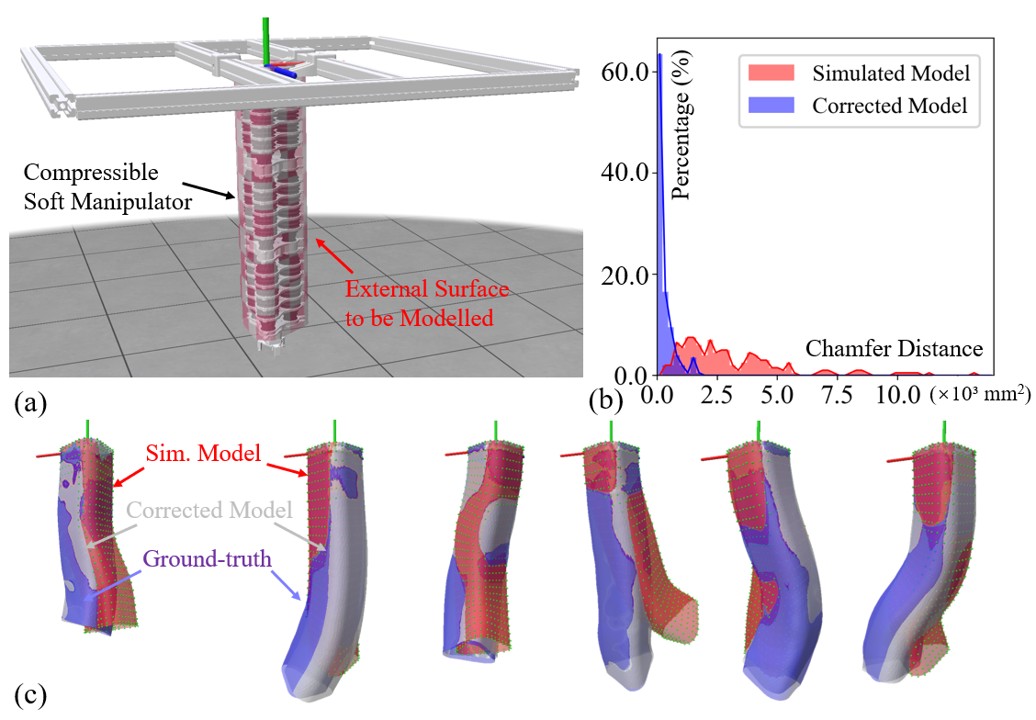}
\caption{\JENew{
Our sim-to-real approach is evaluated on an advanced soft manipulator undergoing compressive deformations, with simulations conducted in MuJoCo (a). The Chamfer distance statistics on the simulated model and the corrected model over 200 test shapes are presented in (b). Representative test frames illustrating different deformations are shown in (c).}
}\label{fig:Compression}
\end{figure}

\subsection{Results of Sim-to-Real Learning by MoCap}
\JENew{The performance of our approach has also been tested by the data captured on a Mocap system.}
\subsubsection{Deformable mannequin}
Experimental tests have been conducted to verify the performance of our sim-to-real method \JENew{with sparsely and partially acquired marker sets -- obtained on the deformable mannequin setup}. First of all, we randomly change all the nine actuation parameters and use our forward kinematic network to predict free-form surface shapes. Four instances are selected to apply the actuation onto the physical setup. The resultant shapes on the soft robotic mannequin are scanned and compared with the simulated models (i.e., without sim-to-real transfer) and the corrected models (i.e., by applying the sim-to-real transfer). The shape approximate errors are evaluated by first correcting the pose of scanned model using the ICP-based registration and then compute the distances between every surface sample points to their closest points on the scanned model. As can be found from the results shown in Fig.\ref{fig:Sim2RealEffectiveStudy}, the shape approximation errors on the corrected models were significantly reduced.

\subsubsection{Compressible manipulator}
\JENew{
To validate the effectiveness of our method in handling deformation under compression, we tested our sim-to-real transfer approach on the advanced soft manipulator setup with a $3 \times 3$ chamber configuration (Fig.\ref{fig:HardwareSystem}(d)). The simulation dataset was generated using MuJoCo \cite{todorov2012mujoco} (see Fig.\ref{fig:Compression}(a)), while the corresponding physical dataset was collected using a MoCap system. A total of 700 sample shapes were recorded during the experiments, with 500 samples used for training and 200 for testing. As shown in Fig.\ref{fig:Compression}(b), our proposed sim-to-real method achieves significantly improved accuracy on the test set. Additionally, Fig.~\ref{fig:Compression}(c) presents 6 representative frames from the test set -- capturing different states of deformations -- where the simulated model, corrected model, and ground truth are visualized together to illustrate the differences.
}

\begin{figure*}[!t] 
\centering
\includegraphics[width=1.0\linewidth]{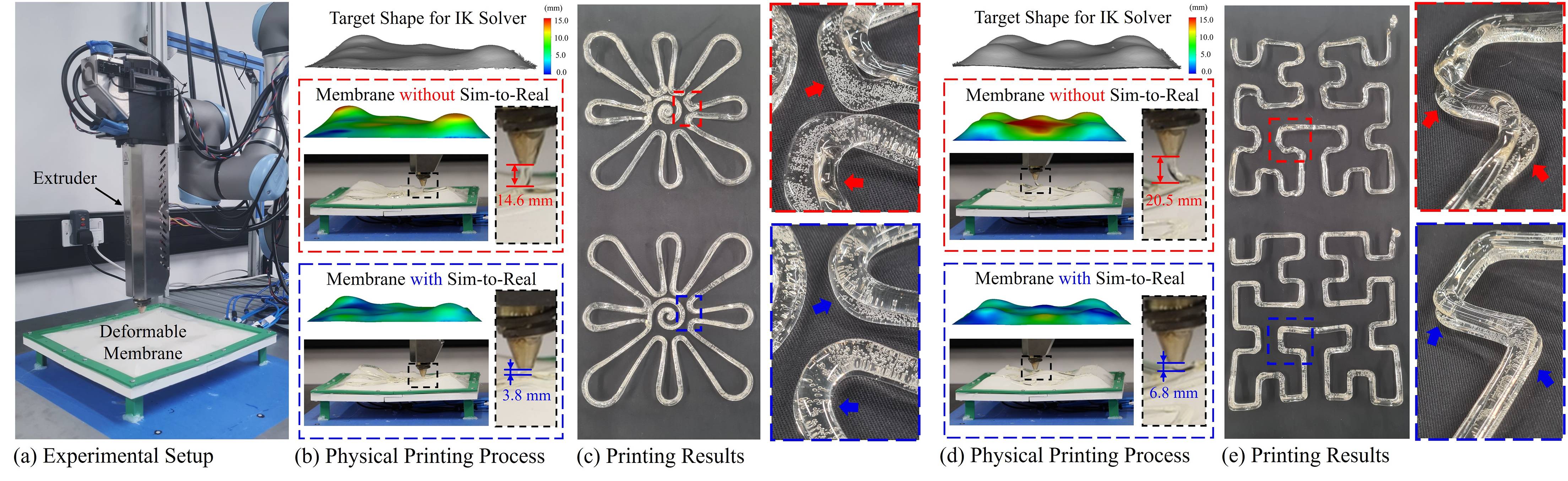}
\caption{\JENew{To verify the effectiveness of our sim-to-real approach for improving the accuracy of IK computation, experiments were conducted to deform the membrane into target shapes using the actuation parameters determined by IK with vs. without sim-to-real. The deformable membrane is employed as a reconfigurable mold for curved 3D printing in this example -- see (a) for the hardware setup. The tests are conducted for printing models on two different target shapes as shown in (b) and (d). The resultant shapes on the deformable membrane by IK with (bottom row) and without (top row) sim-to-real are given in (b) and (d) together with the color maps illustrating the shape approximation errors. In the top row of (c) and (e), we can find poor printing results caused by large gaps. The gap can be significantly reduced after applying sim-to-real transfer.}
}\label{fig:IKResultMembrane}
\end{figure*}

\begin{figure*}[!t] 
\centering
\includegraphics[width=1.0\linewidth]{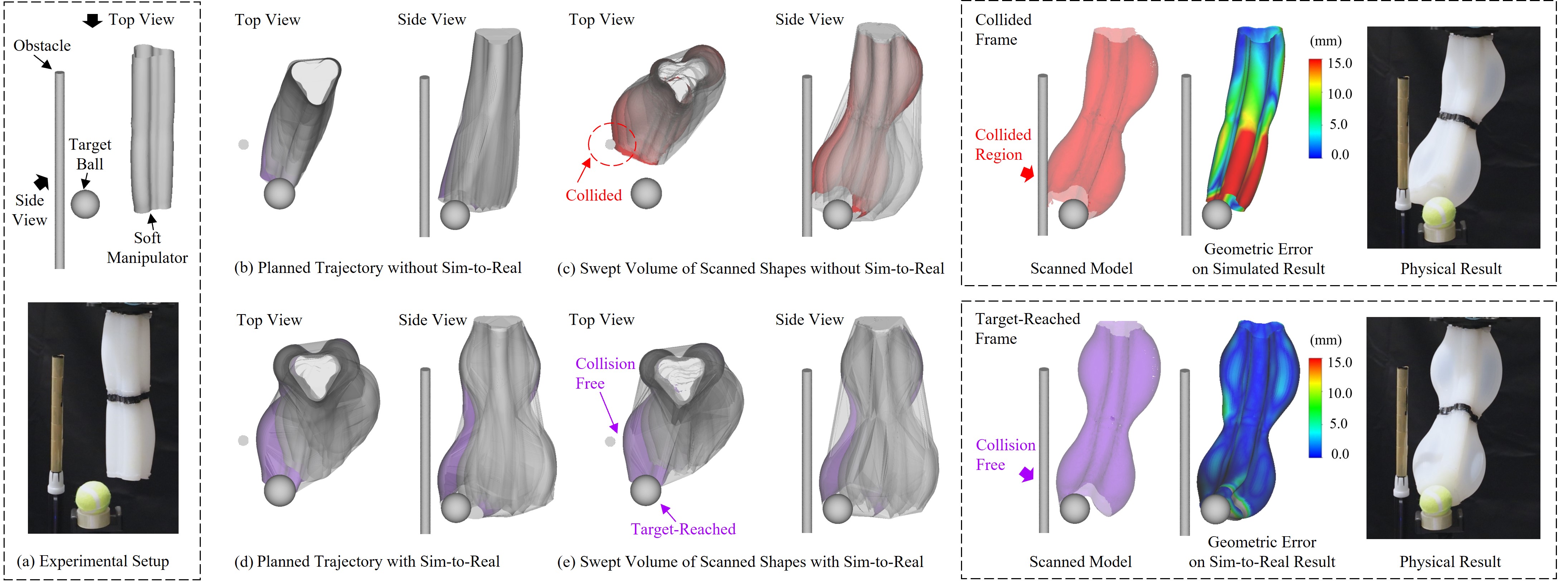}
\caption{\JENew{This example demonstrates the importance of sim-to-real in motion planning to compute a collision-free trajectory for deforming the soft manipulator to touch the target ball while presenting a stick obstacle around -- see (a) for the models in simulation and reality. When using the kinematic computation without sim-to-real, the swept volume of a planned trajectory is as shown in (b). This trajectory however leads to collision when being applied to the physical setup (see the scanned results as shown in the left of (c)) due to the large gap between simulation and reality (see the right of (c) circled by the dash line). Differently, when incorporating our sim-to-real approach into the kinematic computation, the physical execution results are obtained as scanned swept volume shown in the right of (d), which is better aligned with the planned trajectory as shown in the left of (d).}
}\label{fig:IKResultManipulator}
\end{figure*}

\subsection{Results of Inverse Kinematics}\label{subsecResIK}
\JENew{Our sim-to-real approach can improve the IK algorithm introduced in Sec.~\ref{subsecGradientBasedIK}, enabling more accurate realization of the target shape. To validate the effectiveness of this approach in IK computation, we conducted physical experiments on soft robots to demonstrate its performance.} 

\subsubsection{\JENew{Shape Approximation on Deformable Membrane}}
\JENew{The deformation of a soft membrane can be programmed into different shapes, enabling its function as a reconfigurable mold for 3D printing. The hardware setup of our 3D printing experiment is shown in Fig.\ref{fig:IKResultMembrane}(a), where the task of 3D printing is conducted by a UR10e robot arm equipped with a Pulsar$^\mathrm{TM}$ pellet extruder. The material deposition printing process is conducted by following a toolpath computed on a surface. After accurately deforming the membrane into a target shape of this surface, the robotic hardware can print \textit{Polylactic Acid} (PLA) directly on this curved surface.}

\begin{figure}[t] 
\centering
\includegraphics[width=1.0\linewidth]{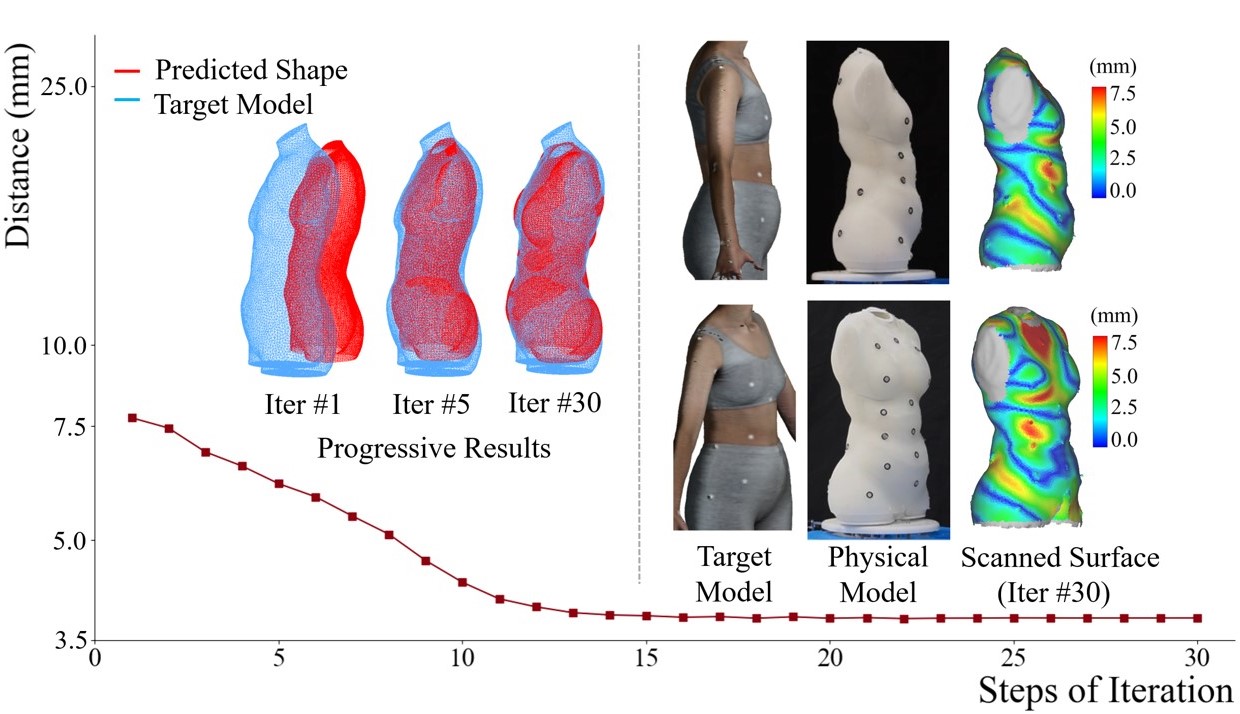}
\caption{The progressive results by applying our NN-based fast IK solver, where the curve shows the average shape approximation errors during the iterations of gradient descent. Using the actuation determined by our fast IK solver, a shape similar to the target model can be realized on the soft mannequin where the error analysis is conducted by 3D scanning with its result shown as the color map.
}\label{fig:progressIKResult}
\end{figure}

\begin{figure}[t] 
\centering
\includegraphics[width=1.0\linewidth]{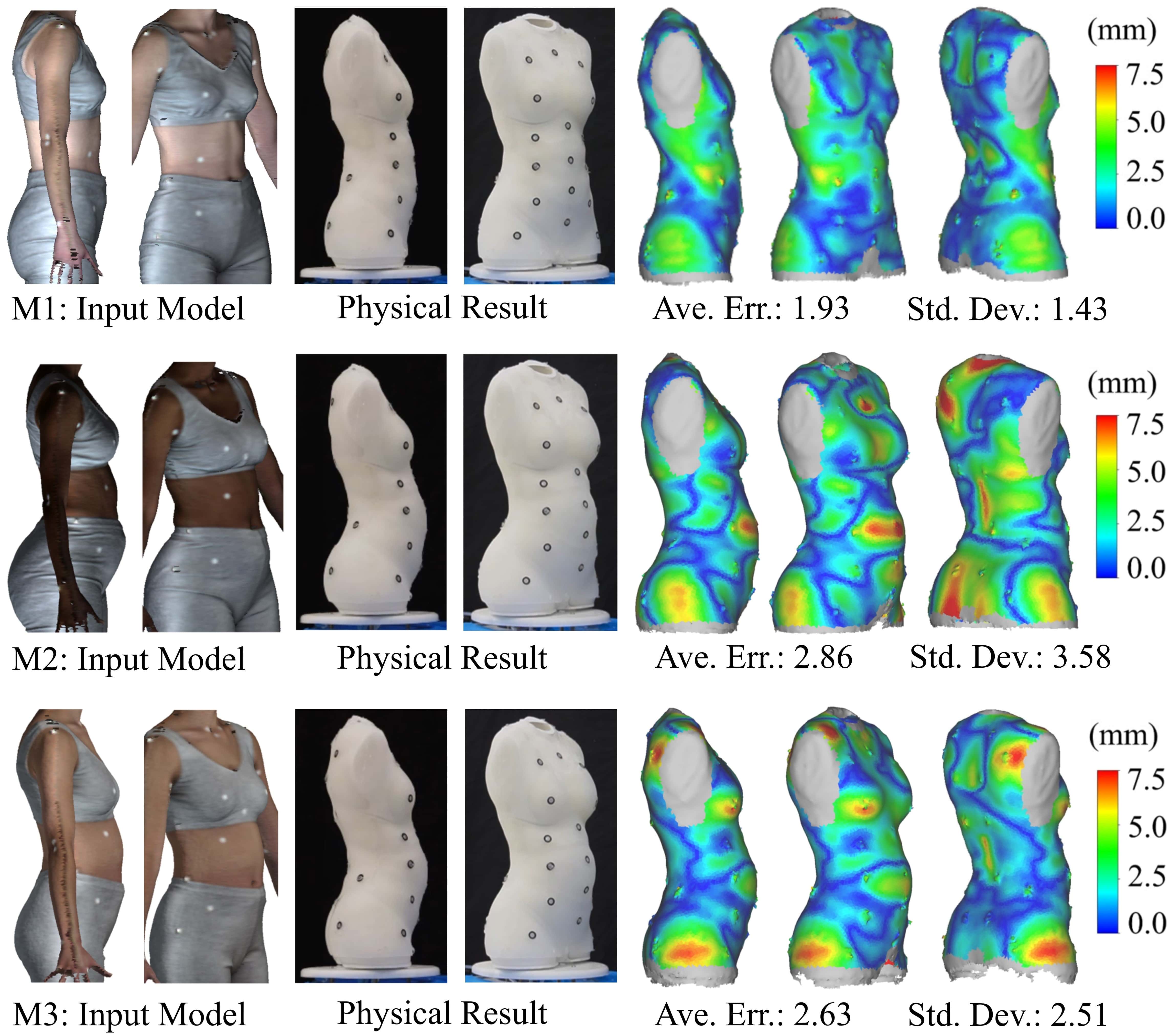}\\
\caption{The physical results of our fast IK solver for three different target models, where the distributions of shape approximation errors are visualized as the color maps. 
}\label{fig:IKResults}
\end{figure}

\JENew{This application requires accurate shape control of the mold. For example, as shown in the top row of Fig.\ref{fig:IKResultMembrane}(b) and (d) circled by red dashed lines, large gaps are formed between the printer head and the mold in the regions with large shape approximation errors (indicated by red arrows). Apparently, this leads to inaccurate material deposition -- see the top row of Fig.\ref{fig:IKResultMembrane}(c) and (e) around the regions indicated by red arrows where misalignment between two layers can be found due to the shape error of mold. The errors can be significantly reduced after incorporating our sim-to-real method in the loop of IK computation. The results with sim-to-real are given in the bottom row of Fig.\ref{fig:IKResultMembrane}(b-e), where precise material deposition can be achieved in 3D printing -- i.e., less misalignment is observed in these turning regions highlighted by blue arrows.
}

\subsubsection{\JENew{Motion Planning of Soft Manipulator}}
\JENew{The accurate shape prediction in kinematic computation is important for the motion planning of a soft robot. We conducted the physical experiment on a soft manipulator with 6 DoFs as shown in Fig.\ref{fig:IKResultManipulator}. Specifically, a sampling-based motion planning algorithm \cite{karaman2011sampling} is employed to compute a sequence of samples as actuation parameters that progressively deform the soft manipulator to touch the target ball while presenting an obstacle stick around the manipulator -- see Fig.\ref{fig:IKResultManipulator}(a) for the experimental setup. Without the sim-to-real correction, the shapes predicted by the kinematic computation are significantly different from the reality. This leads to the risk of collision on a planned collision-free trajectory (see Fig.\ref{fig:IKResultManipulator}(c) for the planned collision-free swept volume and the scanned physical execution colliding with the obstacle stick). The problem can be well solved after incorporating our sim-to-real method into the kinematic computation -- see the results in Fig.~\ref{fig:IKResultManipulator}(d).}
\begin{figure*}[t] 
\centering
\includegraphics[width=1.0\linewidth]{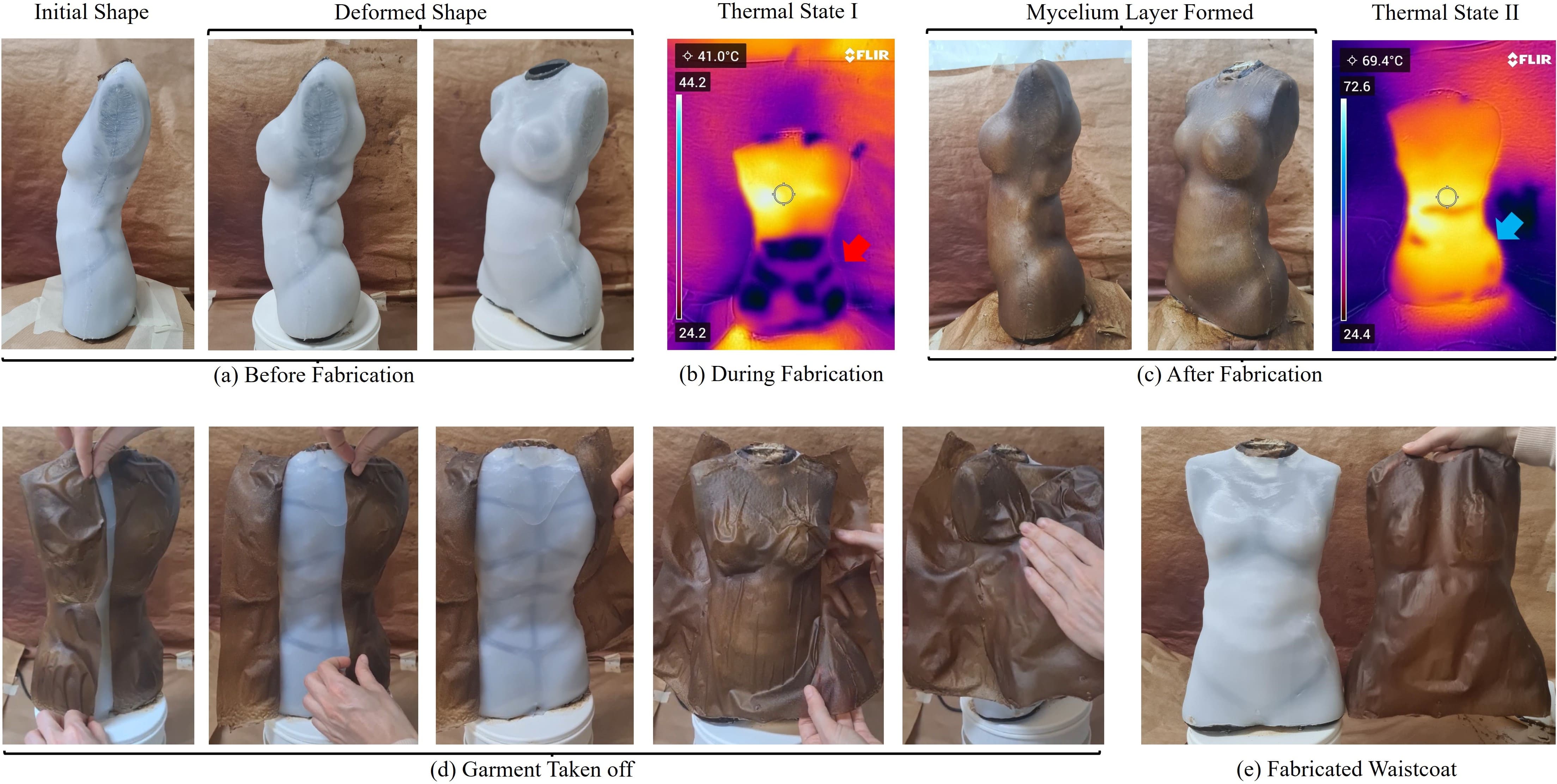}\\
\caption{\JENew{Sustainable and customized garment can be fabricated by using the deformable mannequin as a mold: (a) the mannequin is first deformed into the scanned body shape of a customer, (b) biodegradable material is placed on top of the mannequin and heated under a controlled temperature, (c) a mycelium layer is formed as a piece of garment in the mannequin's shape after heating, (d) the garment is taken off from the mannequin, and (e) the resultant  garment in a customized shape is formed.}
}\label{fig:GarmentFabrication}
\end{figure*}

\subsubsection{Shape Approximation on Deformable Mannequin}
We have tested the performance of our NN-based fast IK solver on a variety of individual shapes presented in the CAESAR dataset~\cite{CAESAR2002}. Fig.\ref{fig:progressIKResult} shows the progressive results when applying our IK solver to realize the target shape on the physical setup of pneumatic driven deformable mannequin. It can be observed that our method converges very fast -- i.e., around 15 iterations where each iteration takes less than 0.5 second. The distribution of shape approximation errors are generated by scanning the soft mannequin that has been deformed using the resultant actuation determined by our IK solver. \JENew{Results on three different models are given in Fig.\ref{fig:IKResults}}. 
Our fast IK solver has also been tested on other models randomly selected from the CAESAR dataset as shown in Appendix \ref{AppendixAdditionalIKTests}.

\section{Conclusion and Discussion}
In this paper, \JENew{we addressed the challenge of bridging the gap between simulated and physically deformed free-form surfaces. We proposed a novel sim-to-real framework that jointly learns a deformation function space and a confidence map to transfer simulated geometries to their real-world counterparts using diverse and potentially incomplete input data. Unlike prior methods, our approach does not rely on pre-established correspondences and is robust to partial observations. Integrated into a neural network-based computational pipeline, the method effectively solves the inverse kinematics problem across various pneumatically actuated deformable robots, including a deformable membrane, two soft manipulators, and a deformable mannequin. Our approach advances the state of the art in shape control of deformable free-form surfaces, paving the way for more accurate and practical deployment in soft robotic applications.}

Our approach imposes an implicit assumption on the configuration-independence of $\Phi(\cdot)$, which is a potential theoretical limitation. However, we actually did not find difficulty in practice mainly based on two reasons:
\begin{itemize}
    \item Firstly, the simulated and the real shapes usually share similar patterns so that they can be effectively addressed through sim-to-real transfer;
    \item Secondly, the training dataset has been well designed to cover most of the possible configurations to relieve the challenge.
\end{itemize}
Although the experimental tests conducted on four soft robots is promising, our method shares the limitation of all learning-based approaches -- the results heavily rely on the quality of training datasets. Moreover, our method cannot perform well on surfaces with sharp features, which is due to the fact that RBFs are inherently smooth and continuous. Therefore, we only apply it to soft robots with freeform deformable surfaces. In our future work, we will use the soft deformable membrane and mannequin as molds to fabricate products with customized shape. Specifically, as demonstrated in the proof-of-concept experiments in collaboration with the Dutch startup NEFFA~\cite{morby2016}, our soft robotic mannequin is first deformed to the body shape of a customer and then serves as a reusable 3D mold for making a garment using biodegradable materials (see Fig.\ref{fig:GarmentFabrication} for the fabrication process). Reliable shape control of deformable soft robots under different temperatures needs to be investigated in our future research. Furthermore, we plan to test the soft manipulators in pick-and-place tasks.

\section*{Acknowledgement}
The project is partially supported by the chair professorship fund of the University of Manchester and the research fund of UK Engineering and Physical Sciences Research Council (EPSRC) (Ref.\#: EP/W024985/1). \JENew{The sustainable fabrication of garments shown in Fig.\ref{fig:GarmentFabrication} is conducted with the help of Aniela Hoitink and Iman Hadzhivalcheva at NEFFA by using their material and fabrication process.}

\bibliographystyle{IEEEtran}

\bibliography{TRO.bib}

\begin{figure*}[!t] 
\centering
\includegraphics[width=.9\linewidth]{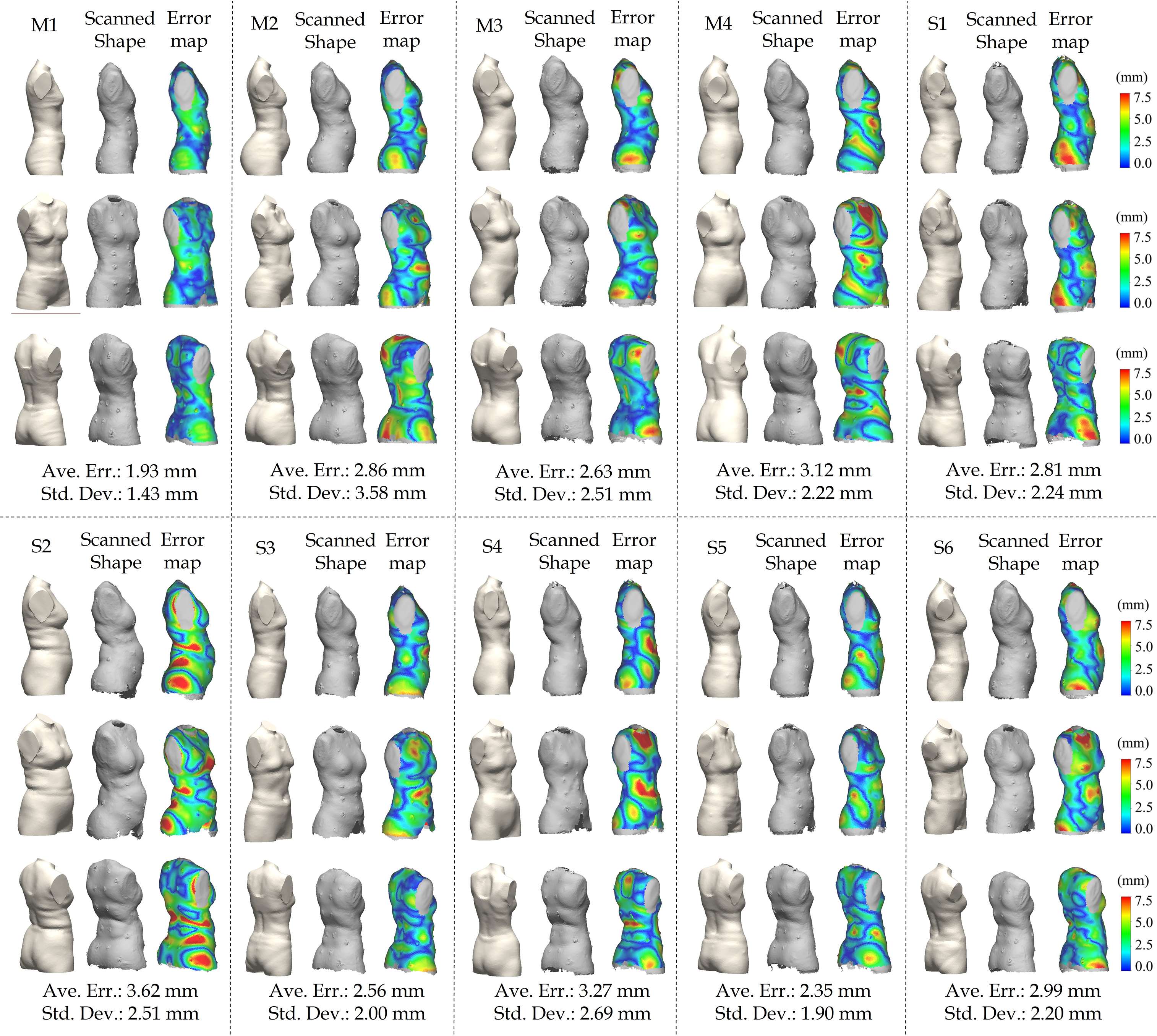}
\caption{To verify the effectiveness of our fast IK solver, we have selected 10 target models exhibiting a wide range of body dimensions. These models encompass those previously discussed in our paper (denoted as $M1 - M4$) as well as six additional models denoted as $S1 - S6$. It can be found that very small errors of shape approximation are generated between the physical results and the target shapes.
}\label{fig:IKResultsAdditional}
\end{figure*}

\appendix

\section{Differentiation of NN-based Forward Kinematics}
%

\subsection{Differentiation of NN-based Forward
Kinematics}
\label{AppendixDifferentiation}
The gradient computation of NN for forward kinematics (including sim-to-real transfer) is crucial for our fast IK solver since it provides first-order information for optimization. 

First of all, the gradient of a sample point $\mathbf{p}^*$ on a corrected model $\mathcal{S}^{*}$ with respect to the actuation parameters $\mathbf{a}$ has been presented in Eq.\eqref{eqDiffp_Diffa}, and all terms in this equation will be explained below. 


The term $\frac{\partial \mathcal{N}_{s2r}}{\partial \mathbf{Q}}$ 
can be calculated through the RBF-based space warping function described in Eq.\eqref{eqRBFSpaceWarping}.
Note that 
$\mathbf{A} = [\bm{\alpha}_1, \bm{\alpha}_2, \bm{\alpha}_3]$ and $\bm{\gamma} = [\bm{\alpha}_0, \bm{\alpha}_1, ... \bm{\beta}_N]$. All vectors are column vectors. Therefore, the gradient can be calculated as:
\begin{flalign}
\label{eqGradCalPnt2SimMarkers1}
\frac{\partial \mathcal{N}_{s2r} }{\partial \mathbf{Q}} &
= \frac{\partial \mathbf{p}^* }{\partial \{ \mathbf{q}\}}
 = \bigg[\begin{array}{c|c|c}
\frac{\partial \mathbf{p}^* }{\partial \mathbf{q}_1}
& \hdots{}
&\frac{\partial \mathbf{p}^* }{\partial \mathbf{q}_N}
\end{array}\bigg].
\end{flalign}
For each term in Eq.~(\ref{eqGradCalPnt2SimMarkers1}), we can have
\begin{flalign}
\label{eqSubTerm}
\frac{\partial \mathbf{p}^* }{\partial \mathbf{q}_i} &=\bm{\beta}_i
(\frac{\partial e^{-c\| \mathbf{p} - \mathbf{q}_i \|^2} }
{\partial \mathbf{q}_i })^T,
\end{flalign}
where the last term of Eq.~(\ref{eqSubTerm}) can be further expanded as
\begin{flalign}
\label{eqGradCalPnt2SimMarkers2}
\frac{\partial e^{-c\| \mathbf{p} - \mathbf{q}_i \|^2} }{\partial \mathbf{q}_i } = \qquad  \qquad\qquad\qquad \qquad  \qquad\qquad \quad  \nonumber\\
- \frac{\partial e^{-c\| \mathbf{p} - \mathbf{q}_i \|^2}    }{\partial \mathopen| \mathbf{p} - \mathbf{q}_{i} \mathclose|} 
\cdot((\mathbf{p} - \mathbf{q}_{i})^T(\mathbf{p} - \mathbf{q}_{i}))^{-\frac{1}{2}} (\mathbf{p} - \mathbf{q}_{i}). 
\end{flalign}
Similarly, the term $\frac{\partial \mathcal{N}_{s2r}}{\partial \mathbf{B}}$ can be calculated by
\begin{flalign}
\label{eqGradCalPnt2SimQueryPnt3}
\frac{\partial \mathcal{N}_{s2r}}{\partial \mathbf{B}} = \frac{\partial \mathbf{p}^*}{\partial \mathbf{p}} &=
\mathbf{A} + 
\sum_{i=1}^{N} \bm{\beta}_i (\frac{\partial 
e^{-c\| \mathbf{p} - \mathbf{q}_i \|^2}
}{\partial \mathbf{p}})^T.
\end{flalign}
Note that the last term of Eq. (\ref{eqGradCalPnt2SimQueryPnt3}) is similar to Eq. (\ref{eqGradCalPnt2SimMarkers2}), and we only need to change the sign as
\begin{flalign}
\label{eqGradCalPnt2SimQueryPnt2}
\frac{\partial 
e^{-c\| \mathbf{p} - \mathbf{q}_i \|^2}
}{\partial \mathbf{p}} = - \frac{\partial 
e^{-c\| \mathbf{p} - \mathbf{q}_i \|^2}
}{\partial \mathbf{q}_i }.
\end{flalign}


The gradient of a sample 
point $\mathbf{p}^*$ on the corrected model with respect to the function variable $\bm{\gamma}$ can be obtained as
\begin{flalign}
\label{eqGradCalPnt2RBFCoeff1}
\frac{\partial  \mathbf{p}^*}{\partial \bm{\gamma}} &=
\bigg[\begin{array}{c|c|c|c|c|c}
\frac{\partial  \mathbf{p}^* }{\partial \bm{\alpha}_0  } &
\hdots&
\frac{\partial  \mathbf{p}^* }{\partial \bm{\alpha}_3} &
\frac{\partial  \mathbf{p}^* }{\partial \bm{\beta}_1} &
\hdots
 &
\frac{\partial \mathbf{p}^* }{\partial \bm{\beta}_N} 
\end{array}\bigg].
\end{flalign}
Each component of $\frac{\partial  \mathbf{p}^*}{\partial \bm{\gamma}}$ as shown in Eq.(\ref{eqGradCalPnt2RBFCoeff1}) can be calculated by following equations:
\begin{align}
  \frac{\partial \mathbf{p}^* }{\partial \bm{\alpha}_0} = \mathbf{I}, &&
 \frac{\partial \mathbf{p}^* }{\partial \bm{\alpha}_1} = (\mathbf{p})_x\mathbf{I}, &&
 \frac{\partial \mathbf{p}^* }{\partial \bm{\alpha}_2} = (\mathbf{p})_y\mathbf{I},  \nonumber
\end{align}

\begin{align}
\label{eqGradCalPnt2RBFCoeff2}
 \frac{\partial \mathbf{p}^* }{\partial \bm{\alpha}_3} = (\mathbf{p})_z\mathbf{I}, &&
  \frac{\partial \mathbf{p}^* }{\partial \bm{\beta}_i} = e^{-c\| \mathbf{p} - \mathbf{q}_i \|^2} \mathbf{I}.
\end{align}
In Eq.~(\ref{eqGradCalPnt2RBFCoeff2}), $(\mathbf{p})_x$, $(\mathbf{p})_y$ and $(\mathbf{p})_z$ denote the x, y, and z components (scalar value) of the sample point $\mathbf{p}$ on the simulation surface.

The terms $\frac{\partial \mathbf{Q}} {\partial \mathcal{N}_{fk}}$ and $\frac{\partial \mathbf{B}} {\partial \mathcal{N}_{fk}}$
are the gradients of the point(s) on the simulation surface with respect to control points $\mathcal{S}^c$ (stored as a flattened column vector).
According to Eq.\eqref{eqBSplineFunc}, the gradient can then be calculated through:
\begin{flalign}
\label{eqGradientBSpline}
\frac{\partial \mathbf{B}}{\partial \mathcal{S}^c_{ij}} = 
N_{ik}(u) N_{jl}(v)\mathbf{I}.
\end{flalign}
The gradient of B-spline 
control points to B-spline control points offsets are identity matrix $\mathbf{I}$. 

The terms 
$\frac{\partial \bm{\gamma}}{\partial \mathcal{N}_{fk}}$ and $\frac{\partial \mathcal{N}_{fk}}{\partial \mathbf{a}}$ can be easily acquired through the back-propagation of $\mathcal{N}_{rbf}$ and $\mathcal{N}_{fk}$.


\subsection{Additional IK Tests with More Target Shapes}\label{AppendixAdditionalIKTests}

\begin{table}[!t]\footnotesize
\vspace{0pt}
\caption{Statistics of IK computing time for each iteration}
\label{Tab:time}
\begin{tabular}{c|c|c|c|c|c}  
\hline  \hline   
\specialrule{0em}{2pt}{1pt}         
\multicolumn{1}{c|}{\multirow{5}{*}{Target}}  & 
\multicolumn{4}{c|}{Average Time of Each Iteration (Unit: ms)} & \multicolumn{1}{c}{\multirow{5}{*}{Total Iter. \#}} 
\\ 
\specialrule{0em}{2pt}{1pt} 
\cline{2-5}
\specialrule{0em}{2pt}{1pt} 
\multicolumn{1}{c|}{}  & 
\multicolumn{2}{c|}{NN-based FK Pipeline} & 
\multicolumn{1}{c|}{\multirow{3}{*}{ICP}} &
\multicolumn{1}{c|}{\multirow{3}{*}{Total Time$^\dagger$}} &
\multicolumn{1}{c}{}
\\ 
\specialrule{0em}{2pt}{1pt} 

\cline{2-3}
\specialrule{0em}{2pt}{1pt} 
\multicolumn{1}{c|}{}  & 

\multicolumn{1}{c|}{Forward} &  
\multicolumn{1}{c|}{Gradient} &   

\multicolumn{1}{c|}{} &  
\multicolumn{1}{c|}{} & 
\multicolumn{1}{c}{} 
\\
\specialrule{0em}{2pt}{1pt} 
\hline \hline    
\specialrule{0em}{2pt}{1pt} 
\multicolumn{1}{c|}{M1}  & 

\multicolumn{1}{c|}{65.8} &  
\multicolumn{1}{c|}{137.7} &  

\multicolumn{1}{c|}{171.0} &  
\multicolumn{1}{c|}{374.5} & 
\multicolumn{1}{c}{10}  

\\

\multicolumn{1}{c|}{M2}  & 

\multicolumn{1}{c|}{75.7} &  
\multicolumn{1}{c|}{139.6} &  

\multicolumn{1}{c|}{174.3} &  
\multicolumn{1}{c|}{389.6} & 
\multicolumn{1}{c}{22}  

\\

\multicolumn{1}{c|}{M3}  & 

\multicolumn{1}{c|}{83.5} &  
\multicolumn{1}{c|}{118.8} &  

\multicolumn{1}{c|}{175.2} &  
\multicolumn{1}{c|}{377.5} & 
\multicolumn{1}{c}{19}  

\\

\multicolumn{1}{c|}{M4}  & 

\multicolumn{1}{c|}{69.9} &  
\multicolumn{1}{c|}{120.6} &  

\multicolumn{1}{c|}{213.8} &  
\multicolumn{1}{c|}{404.3} & 
\multicolumn{1}{c}{15}  

\\

\multicolumn{1}{c|}{S1}  & 

\multicolumn{1}{c|}{65.1} &  
\multicolumn{1}{c|}{131.1} &  

\multicolumn{1}{c|}{210.7} &  
\multicolumn{1}{c|}{406.9} & 
\multicolumn{1}{c}{13} \\ 

\multicolumn{1}{c|}{S2}  & 

\multicolumn{1}{c|}{74.8} &  
\multicolumn{1}{c|}{133.8} &  

\multicolumn{1}{c|}{180.3} &  
\multicolumn{1}{c|}{388.9}  & 
\multicolumn{1}{c}{25}  

\\

\multicolumn{1}{c|}{S3}  & 

\multicolumn{1}{c|}{75.1} &  
\multicolumn{1}{c|}{125.3} &  

\multicolumn{1}{c|}{193.4} &  
\multicolumn{1}{c|}{393.8} & 
\multicolumn{1}{c}{12} \\ 

\multicolumn{1}{c|}{S4}  & 

\multicolumn{1}{c|}{80.1} &  
\multicolumn{1}{c|}{125.8} &  

\multicolumn{1}{c|}{186.7} &  
\multicolumn{1}{c|}{392.6}  &
\multicolumn{1}{c}{17} \\ 

\multicolumn{1}{c|}{S5}  & 

\multicolumn{1}{c|}{68.3} &  
\multicolumn{1}{c|}{139.3} &  

\multicolumn{1}{c|}{179.2} &  
\multicolumn{1}{c|}{386.8}  &
\multicolumn{1}{c}{14} \\ 

\multicolumn{1}{c|}{S6}  & 

\multicolumn{1}{c|}{73.5} &  
\multicolumn{1}{c|}{122.4} &  

\multicolumn{1}{c|}{178.5} &  
\multicolumn{1}{c|}{374.4} & 
\multicolumn{1}{c}{19} \\ 
\specialrule{0em}{2pt}{1pt} 

\hline \hline 
\end{tabular}
\begin{flushleft}\footnotesize
$^\dagger$~This is the running time of the inference phase on C++ using libTorch.
\end{flushleft}
\end{table}

We further test the performance of our fast IK solver by six additional target models from the CAESAR dataset with large variations in body dimensions, labeled S1 to S6. We verify the effectiveness of our IK solver by applying the determined actuation parameters to physically deform the mannequin. The resultant free-form shape is then scanned and compared with the target shape. The comparisons have been shown in Fig.~\ref{fig:IKResultsAdditional}.

Furthermore, the statistics of computational time have been given in Table~\ref{Tab:time} to demonstrate the efficiency of our algorithm. It can be found that the most time-consuming step is the ICP-based pose estimation. Each iteration of the IK computation can be completed in 300-400 ms. 

\vfill

\begin{IEEEbiography}[{\includegraphics[width=1in,height=1.25in,clip,keepaspectratio]{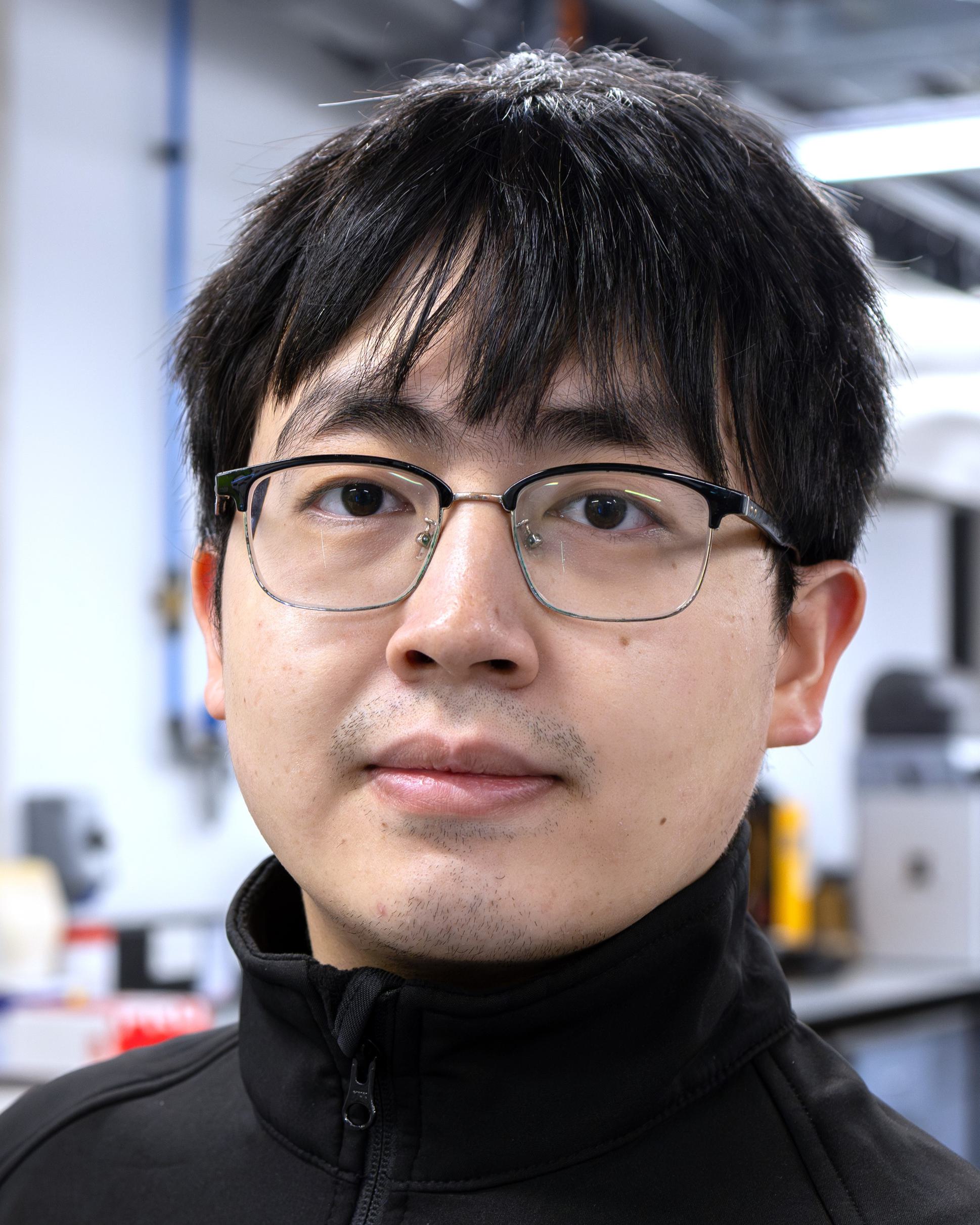}}]{Yingjun Tian} received the B.E. degree in mechanical engineering from the University of Science and Technology of China, Hefei, China, in 2019, and the Ph.D. degree in mechanical engineering from The University of Manchester, Manchester, U.K., in 2024.

He is currently a Postdoctoral Research Associate with the Digital Manufacturing Lab, The University of Manchester. His research interests include computational design, soft robotics, and robot learning.
\end{IEEEbiography}

\begin{IEEEbiography}[{\includegraphics[width=1in,height=1.25in,clip,keepaspectratio]{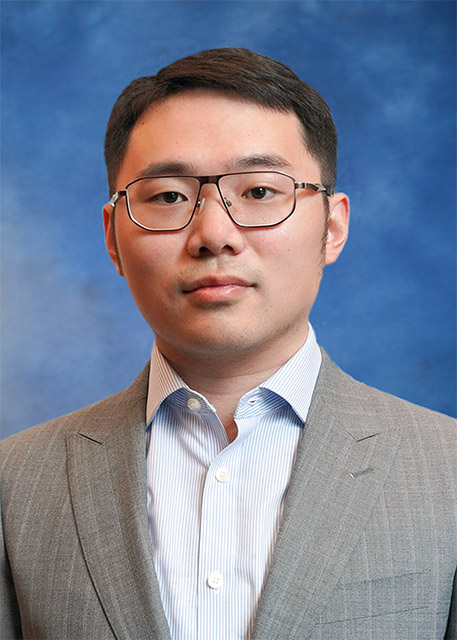}}]{Guoxin Fang} (Member, IEEE) received the B.E. degree in mechanical engineering from the Beijing Institute of Technology, Beijing, China, in 2016, and the Ph.D. degree in advanced manufacturing from Delft University of Technology, Delft, The Netherlands, in 2022. He is currently an Assistant Professor with the Department of Mechanical and Automation Engineering at The Chinese University of Hong Kong, Hong Kong. Prior to this, he was a Research Associate with the Department of Mechanical, Aerospace and Civil Engineering at The University of Manchester, U.K. His research interests include geometric computing, computational design, digital fabrication, and robotics.
\end{IEEEbiography}

\begin{IEEEbiography}[{\includegraphics[width=1in,height=1.25in,clip,keepaspectratio]{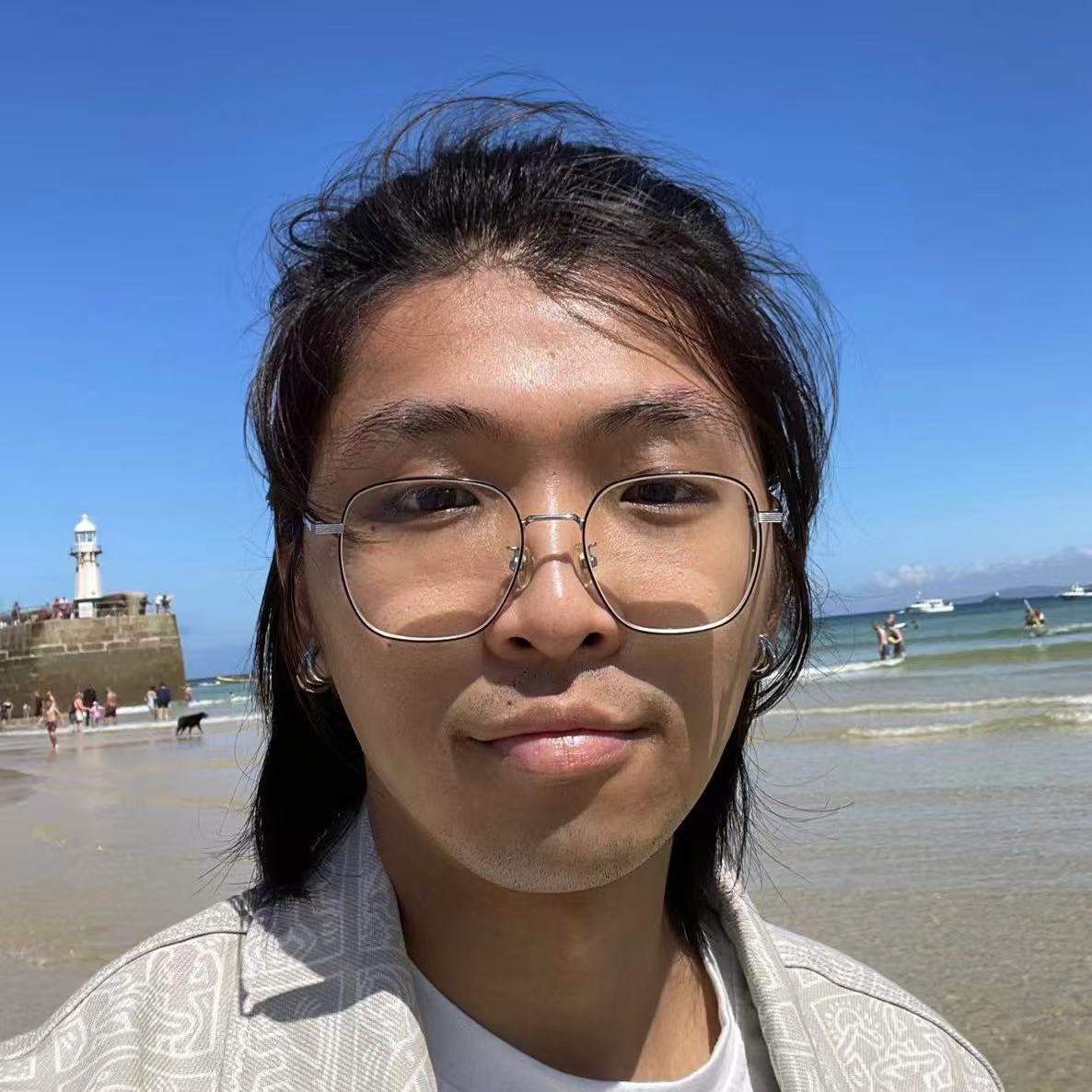}}]
{Renbo Su} received his B.Eng. degree in Material Forming and Control Engineering from East China University of Science and Technology, Shanghai, China, in 2019. He then earned his M.Sc. degree in Advanced Manufacturing Technology and Systems Management in 2020 and his Ph.D. degree in Mechanical Engineering in 2025, both from the University of Manchester, UK. He is currently a postdoctoral research fellow with the College of Intelligent Robotics and Advanced Manufacturing, Fudan University, Shanghai, China. 

His research interests include the design and manufacturing of rigid–soft hybrid robots, metamaterials, as well as the development of equipment for multi-axis additive manufacturing.
\end{IEEEbiography}

\begin{IEEEbiography}[{\includegraphics[width=1in,height=1.25in,clip,keepaspectratio]{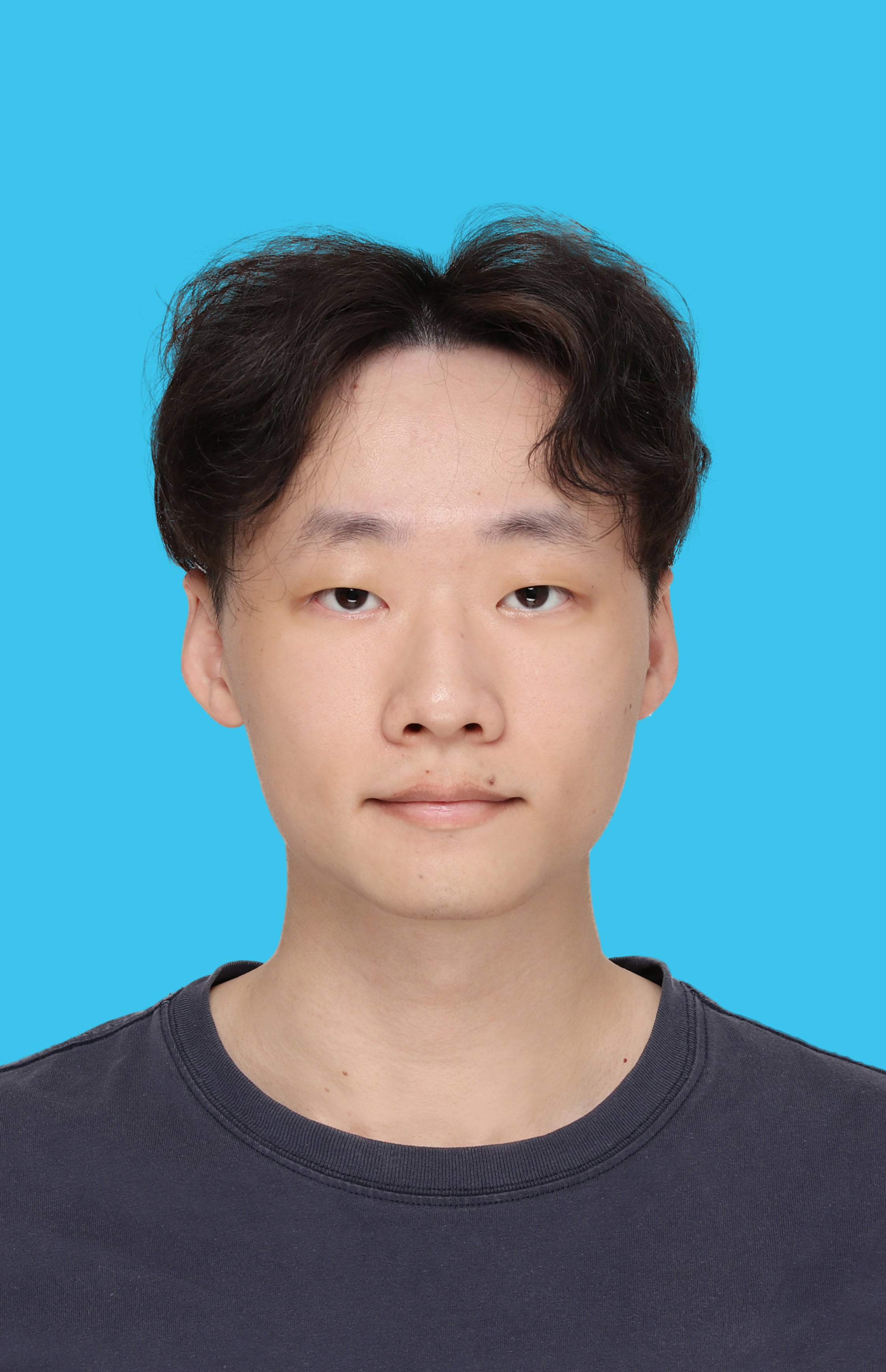}}]
{Aoran Lyu} is currently pursuing a Ph.D. degree in Production and Manufacturing Engineering in the Department of Mechanical and Aerospace Engineering at the University of Manchester. He received his B.S. degree in Mathematics from South China University of Technology in 2021 and his M.E. degree in Computer Technology from the same university in 2024. His research interests include computer graphics, physics-based simulation, and computational design.
\end{IEEEbiography}

\begin{IEEEbiography}[{\includegraphics[width=1in,height=1.25in,clip,keepaspectratio]{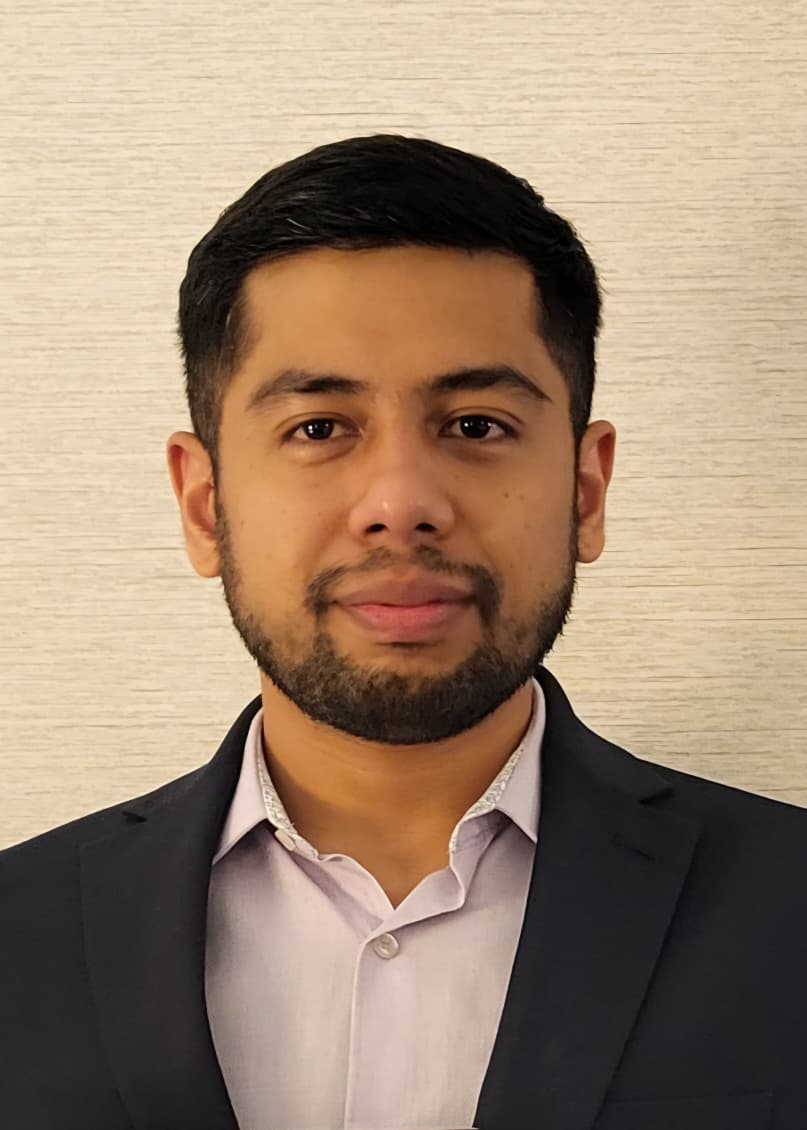}}]
{Neelotpal Dutta} received the B.Tech. (Hons.) degree in Mechanical Engineering from the Indian Institute of Technology Mandi, India, in 2020, where he was awarded the President of India Gold Medal. He is currently a Ph.D. student in the Department of Mechanical and Aerospace Engineering at the University of Manchester, U.K. His research interests include geometric computing, digital manufacturing, and computer-aided design and manufacturing, with a focus on toolpath generation and process planning for additive and subtractive manufacturing.
\end{IEEEbiography}
\vspace{-10pt}

\begin{IEEEbiography}[{\includegraphics[width=1in,height=1.25in,clip,keepaspectratio]{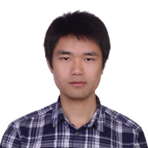}}]
{Weiming Wang} received the BS and PhD degrees from the Dalian University of Technology in 2010 and 2016, respectively.

He was a Postdoctoral Researcher supported by the Marie Skłodowska-Curie LEaDing Fellows Programme at the Delft University of Technology, Delft, The Netherlands. Currently, he is a Postdoctoral Research Associate at The University of Manchester, Manchester, U.K. His research interests are Computational Fabrication and Additive Manufacturing.
\end{IEEEbiography}

\begin{IEEEbiography}[{\includegraphics[width=1in,height=1.25in,clip,keepaspectratio]{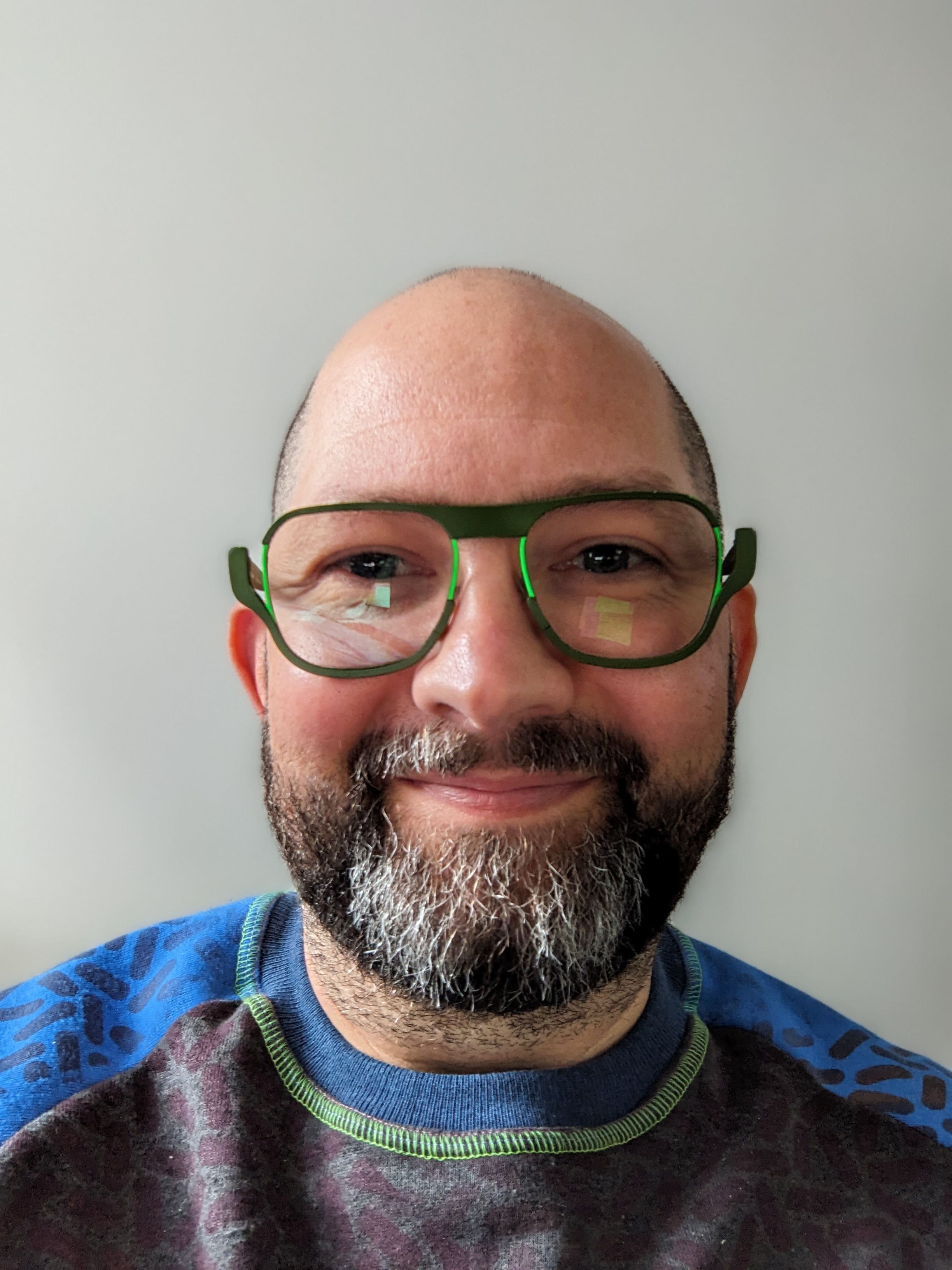}}]
{Simeon Gill} received his PhD on functional ease in 2009 at Manchester Metropolitan University (MMU) and has since focused on developing pattern-cutting theory informed by body scan data. He is currently a Reader in Fashion Technology at the University of Manchester, and his recent work uses parametric pattern techniques to automate engineered patterns that meet the fit and functional requirements of individual bodies.
\end{IEEEbiography}

\begin{IEEEbiography}[{\includegraphics[width=1in,height=1.25in,clip,keepaspectratio]{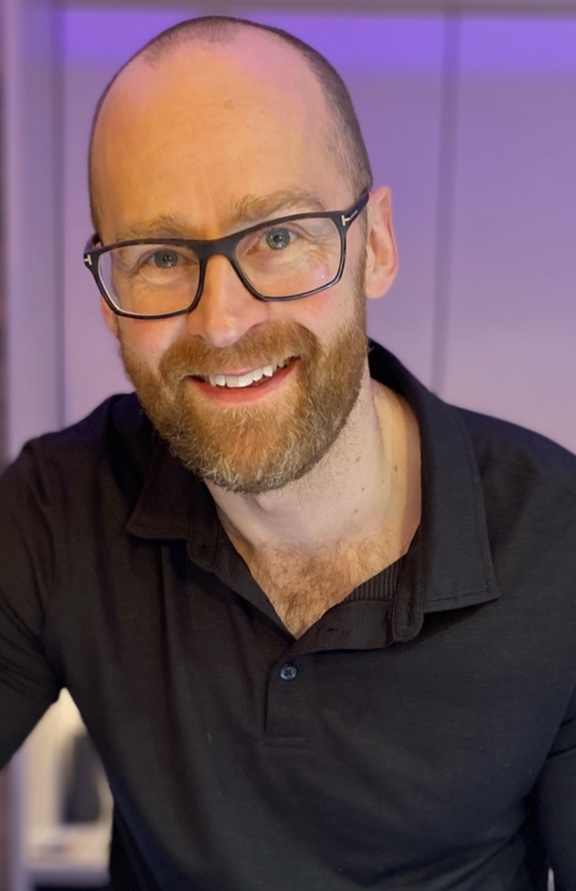}}]
{Andrew Weightman} graduated in 2006 with a Ph.D. in mechanical engineering from the University of Leeds. While at the University of Leeds, he developed rehabilitation robotic technology for improving upper limb function in adults and children with neurological impairment, which was successfully utilized in homes, schools, and clinical settings. He has been at The University of Manchester as part of the centre for robotics and artificial intelligence since 2013. He has research interests in biomimetic mobile robotics, rehabilitation robotics, robotics for nuclear decommissioning, and soft robotics.

\end{IEEEbiography}

\begin{IEEEbiography}[{\includegraphics[width=1in,clip,keepaspectratio]{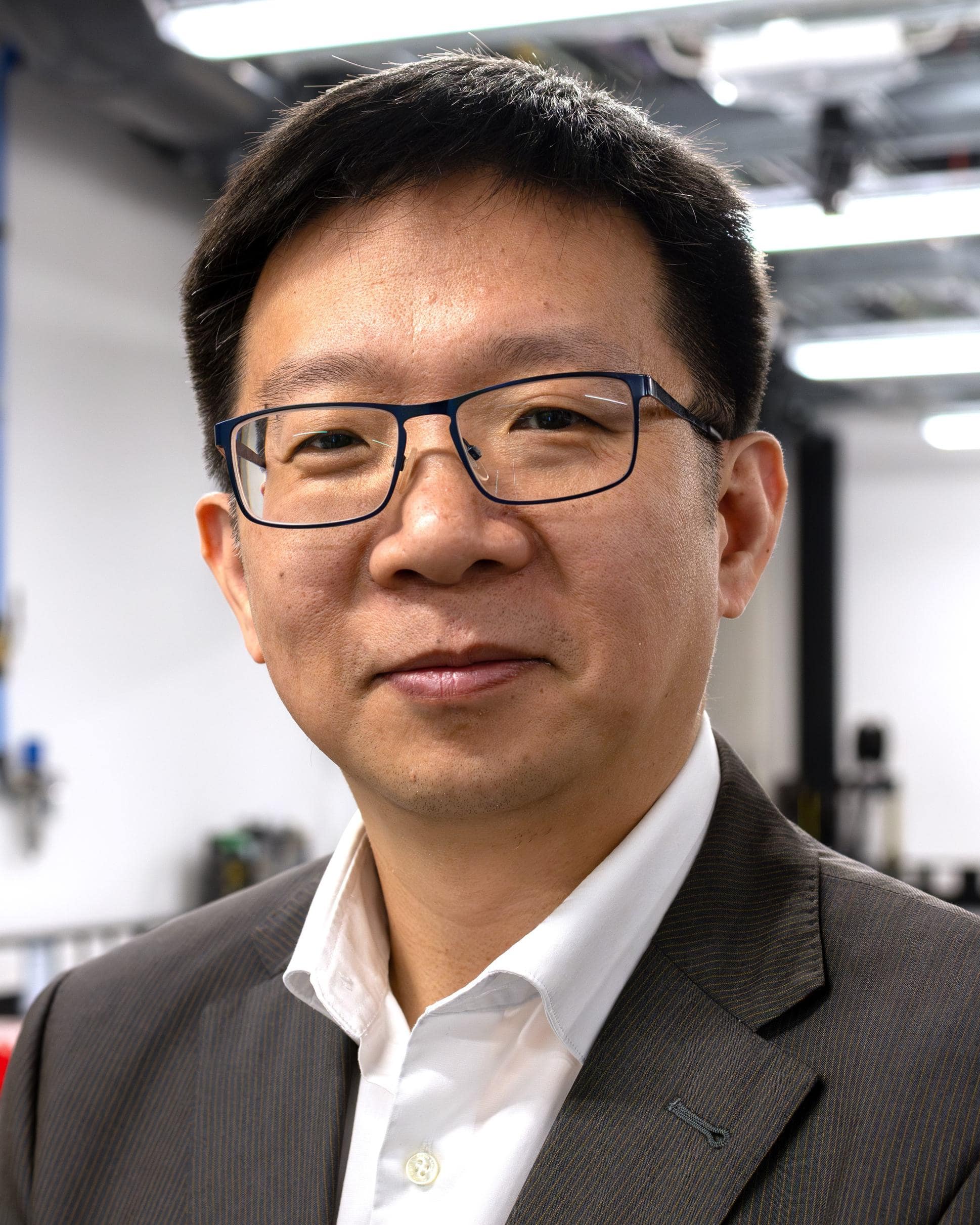}}]{Charlie~C.L.~Wang} is currently Professor of Smart Manufacturing at the University of Manchester (UoM). Before joining UoM in 2020, he worked as Professor and Chair of Advanced Manufacturing at Delft University of Technology, The Netherlands and as Professor of Mechanical and Automation Engineering at the Chinese University of Hong Kong. He received his Ph.D. degree (2002) in mechanical engineering from Hong Kong University of Science and Technology (HKUST). His research interests include Digital Manufacturing, Computational Design, Additive Manufacturing, Soft Robotics, Geometric Computing, and Computer Graphics. He is Fellow of the American Society of Mechanical Engineers (ASME) and the Solid Modelling Association (SMA).
\end{IEEEbiography}

\end{document}